\documentclass{article}

\usepackage[final]{styles/neurips_2025}

\usepackage[utf8]{inputenc} %
\usepackage[T1]{fontenc}    %
\usepackage{hyperref}       %
\usepackage{url}            %
\usepackage{booktabs}       %
\usepackage{amsfonts}       %
\usepackage{nicefrac}       %
\usepackage{microtype}      %
\usepackage{xcolor}         %
\usepackage{xspace}
\usepackage{graphicx}
\usepackage{amsmath} %
\usepackage{multicol}
\usepackage{multirow}
\usepackage{breakurl}
\usepackage[toc,page,header]{appendix}

\usepackage{minitoc}
\usepackage{arydshln}
\usepackage{colortbl}
\usepackage[linesnumbered,ruled,vlined]{algorithm2e}
\usepackage{ml-defs}

\newcommand{\modelname}{TiRex\xspace}
\newcommand{\fullpatchmask}{Contiguous Patch Masking\xspace}
\newcommand{\fullpatchmaskabrr}{CPM\xspace}
\newcommand{\giftevalzs}{GiftEval-ZS benchmark\xspace}
\newcommand{\chronoseval}{Chronos-ZS benchmark\xspace}

\newcommand{\ffdim}{d_{\text{ff}}}
\newcommand{\embdim}{d}
\newcommand{\patchoutdim}{m_{\text{out}}}
\newcommand{\patchindim}{m_{\text{in}}}
\newcommand{\fullpatchmaxprob}{p_{\text{mask}}^{\text{max}}}
\newcommand{\fullpatchmaxcons}{c_{\text{mask}}^{\text{max}}}

\newcommand{\sigB}[1]{$\underline{#1}$}
\newcommand{\sigL}[1]{$#1$}

\newcommand{\nextcaption}{}
\newcommand{\setNextCaption}[1]{\renewcommand{\nextcaption}{#1}}

\title{\modelname: Zero-Shot Forecasting Across Long and Short Horizons with Enhanced In-Context Learning}

\author{%
  \ \ \ \ \ \ Andreas Auer$^{\ 1,2}$
  \ \ \ \ \ \ \ \ \ \ \ \ \ \ 
  Patrick Podest$^{\ 2}$
  \ \ \ \ \ \ \ \ \ \ \ \ \ \ \
  Daniel Klotz$^{\ 3}$ \ \\
 \ \ \ \ \ \ \ \ \ \
  \textbf{Sebastian Böck}$^{\ 1}$ \ \ \ \ \ \ \ \ 
   \ \ \ \  \textbf{G\"{u}nter Klambauer}$^{\ 1,2}$ 
  \ \ \ \ 
  \textbf{Sepp Hochreiter}$^{\ 1,2}$ \\ \\
{$^1$}{NXAI GmbH, Linz, Austria} \\
{$^2$}{ELLIS Unit, LIT AI Lab, Institute for Machine Learning, JKU Linz, Austria}\\
{$^3$}{Interdisciplinary Transformation University Austria, Linz, Austria}
}

\begin{document}

\maketitle

\doparttoc %
\faketableofcontents %

\begin{abstract}
In-context learning, the ability of large language models to perform tasks using only examples provided in the prompt, has recently been adapted for time series forecasting. 
This paradigm enables zero-shot prediction, where past values serve as context for forecasting future values, 
making powerful forecasting tools accessible to non-experts and increasing the performance when training data are scarce.
Most existing zero-shot forecasting approaches rely on transformer architectures, which, despite their success in language, often fall short of expectations in time series forecasting, where recurrent models like LSTMs frequently have the edge. Conversely, while LSTMs are well-suited for time series modeling due to their state-tracking capabilities, they lack strong in-context learning abilities.
We introduce \modelname that closes this gap by leveraging xLSTM, an enhanced LSTM with competitive in-context learning skills. Unlike transformers, state-space models, or parallelizable RNNs such as RWKV, \modelname retains state-tracking, a critical property for long-horizon forecasting. To further facilitate its state-tracking ability, we propose a training-time masking strategy called \fullpatchmaskabrr.
\modelname sets a new state of the art in zero-shot time series forecasting on the HuggingFace benchmarks \textit{GiftEval} and \textit{Chronos-ZS}, outperforming significantly larger models including \textit{TabPFN-TS} (Prior Labs), \textit{Chronos Bolt} (Amazon), \textit{TimesFM} (Google), and \textit{Moirai} (Salesforce) across both short- and long-term forecasts.
\end{abstract}

\section{Introduction}

Recent research in time series forecasting
has adopted in-context learning through large-scale pre-trained models,
analogous to large language models \citep{wooUnifiedTrainingUniversal2024a, ansariChronosLearningLanguage2024b, dasDecoderonlyFoundationModel2024e}. 
These models enable zero-shot forecasting, 
allowing them to generalize to unseen datasets without parameter updates, akin to meta-learning \cite{Hochreiter:01}. 
This capability empowers practitioners without machine learning expertise to use advanced forecasting tools. 
More importantly, zero-shot forecasting significantly improves performance in data-scarce settings, 
where training task-specific models often fail to generalize. 
As a result, in-context learning models hold promise
for broad adoption in domains such as energy, retail, or healthcare.

Most pre-trained time series models are based on transformer architectures \citep{vaswani_attention_2017}, which are well suited for in-context learning, but despite their success in language, often fall short of expectations in time series forecasting \citep[e.g.,][]{zengAreTransformersEffective2023}. 
In contrast, LSTMs \citep{Hochreiter:91a,Hochreiter:97} have demonstrated strong results in time series forecasting 
due to their recurrence and effective state-tracking~\citep[e.g.,][]{nearing2024global}.
Therefore, LSTMs are more expressive than state-space models (SSMs), parallelizable RNNs like RWKV~\citep{peng2023rwkv}, and transformers~\citep{Merrill:23,Merrill:24,Deletang:23}.
However, they lack strong in-context learning capabilities.
To bridge this gap, we adopt xLSTM \citep{beckXLSTMExtendedLong2024a}, 
a modern LSTM variant that incorporates architectural enhancements
for scalability and improved generalization. 
In particular, xLSTM has demonstrated in-context learning performance
comparable to that of transformer-based large language models \citep{beckXLSTM7BRecurrent2025}.

\begin{figure}
    \includegraphics[width=\textwidth]{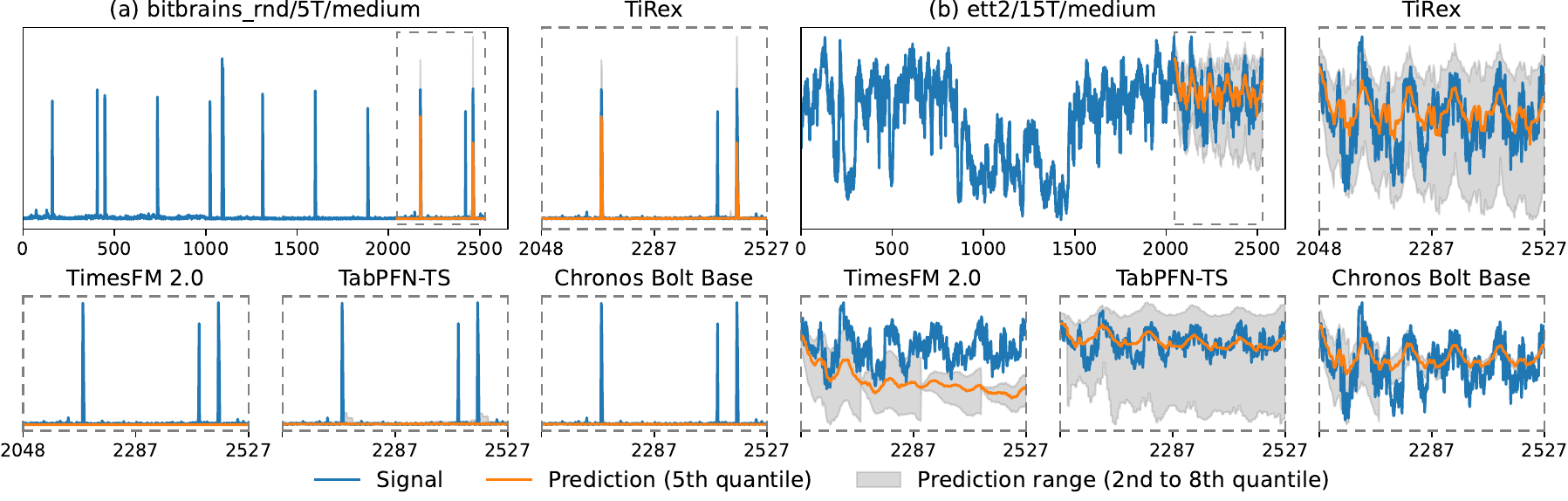}
    \caption{
    Two exemplary time series from the GiftEval benchmark.
    For both examples, we show one plot with the full context and \modelname's prediction, as well as zoomed-in forecasts of the best-performing zero-shot models. Each plot shows the ground truth signal in blue, the model's (median) prediction in orange, and the uncertainty bounds in gray. 
    (a) A time series that exhibits strong peaks. Only \modelname is capable of predicting the periodic short spikes.
    (b) A time series with strong but noisy periodical behavior. \modelname predicts a meaningful uncertainty estimate (quantile range) over the long forecast horizon, while TimesFM and Chronos Bolt struggle because of collapsing quantiles.}
    \label{fig:main-qualiative-comparison}
\end{figure}

To fully unlock xLSTM’s state-tracking abilities, 
we introduce \fullpatchmask (\fullpatchmaskabrr), a novel training-time masking strategy. 
\fullpatchmaskabrr enhances xLSTM's ability to produce coherent long-horizon predictions
by mitigating degradation common in autoregressive multi-step forecasting, as illustrated in Figure~\ref{fig:main-qualiative-comparison}.

While synthetic datasets are frequently used for pre-training forecasting models, 
the potential of data augmentation strategies remains largely untapped, 
unlike their established role in vision pre-training \citep{tian2020makes}. 
To address this, we design and utilize a suite of augmentations. 

\textbf{Our key contributions are:}
\begin{itemize}
    \item \modelname: We present \modelname, a pre-trained time series model based on xLSTM, 
    which sets a new state of the art in zero-shot forecasting. 
    It achieves superior performance across standardized benchmarks, 
    improving both short- and long-term forecasting accuracy.
    \item \fullpatchmask\ (\fullpatchmaskabrr): We propose a novel masking strategy that enhances 
    state-tracking abilities, therefore enabling pre-trained time series models 
    to produce reliable uncertainty estimates over long prediction horizons, 
    effectively addressing autoregressive error accumulation.
    \item Data Augmentation Strategies:
    We introduce three augmentation techniques for time series model pre-training and 
    demonstrate their effectiveness in enhancing the robustness and overall performance of \modelname.
\end{itemize}

After introducing the problem setup and a review of related work, the paper is structured as follows:
Section~\ref{sec:model} introduces \modelname, its architecture, inference strategy, and \fullpatchmask utilized for training.
Section~\ref{sec:datapipeline} describes the proposed training augmentations.
Section~\ref{sec:experiments} evaluates \modelname on two standardized real-world benchmarks and examines the impact of the individual components.
In Section~\ref{sec:conclusion}, we discuss limitations of our approach and conclude the paper.

\subsection{Problem Setup: Zero-Shot Forecasting}
Time series forecasting aims to predict future values of a time series 
based on its past values.
Formally, given a time series $(y_1, y_2, \ldots, y_T)$, with $y_t \in \mathbb{R}$ denoting the 
value at time $t$, the forecasting objective is to predict its future horizon 
$(y_{T+1}, \ldots, y_{T+h})$, where $h$ is the forecast horizon's length.
Throughout the paper, we adopt Python-style array notation and denote a contiguous sequence of values by  $\mathbf{y}_{1:T} := (y_t)_{t=1}^T=(y_1, y_2, \ldots, y_T)$.
Probabilistic forecasting extends this setup by modeling the uncertainty inherent in most time series data.
Instead of producing point estimates, the model learns to approximate the conditional distribution over future outcomes:
\begin{equation}
    \mathcal{P}(\mathbf{y}_{T+1:T+h} \mid \mathbf{y}_{1:T}).
\end{equation}

In a zero-shot forecasting setting, the prediction model is pre-trained on a corpus of time series datasets $C =\{D_1, D_2, \ldots, D_N\}$, where
each $D_n$, $1 \leq n \leq N$, is a time series dataset, e.g., a set of 
time series from a particular domain.
At inference the model is applied directly to time series of new, unseen dataset, i.e., $\mathbf{y} \in D^{\text{test}}\; \text{and}\; \mathbf{y} \notin \bigcup_{i=1}^N D_i$, without any fine-tuning or task-specific supervision.

\subsection{Related Work}

Statistical models such as ARIMA~\citep{boxRecentAdvancesForecasting1968} and exponential smoothing~\citep{hyndmanForecastingExponentialSmoothing2008} are classical approaches in time series forecasting.
In the last decades, however, neural network-based models have emerged as effective alternatives:
Notable examples include DeepAR \citep{salinasDeepARProbabilisticForecasting2020a}, based on a LSTM with a mixture density head; N-BEATS \citep{oreshkinNBEATSNeuralBasis2019}, the first approach that employed a deep block architecture; PatchTST\citep{nieTimeSeriesWorth2022}, a patch-based attention approach; and TFT\citep{limTemporalFusionTransformers2021}, which combines LSTM and transformer components.
These models are trained on multiple time series from a single dataset and require retraining when applied to new tasks.

Currently, pre-trained time series models, with the capability of zero-shot generalization across datasets, predominantly adopt different transformer architectures.
For instance, Chronos~\citep{ansariChronosLearningLanguage2024b}, Chronos-Bolt~\citep{ansariabdulfatirFastAccurateZeroshot2024}, and COSMIC~\citep{auer2025zero} use an encoder-decoder variant. 
Moirai~\citep{wooUnifiedTrainingUniversal2024a} adopts an encoder-only design with a masked modeling objective, and TimesFM~\citep{dasDecoderonlyFoundationModel2024e} follows a decoder-only causal modeling strategy for autoregressive generation.
TabPFN~\citep{hooTabularFoundationModel2025} and its adaptation to time series TabPFN-TS~\citep{hoo2025tablestimetabpfnv2outperforms} use a modified transformer encoder and pre-train only on synthetic data.
A notable exception is TTM~\citep{ekambaramTinyTimeMixers2024b}, since it builds on the MLP-based TSMixer architecture~\citep{chenTSMixerAllMLPArchitecture2023a}.

The dominance of transformer architectures echoes their strong in-context learning capabilities (an essential property for zero-shot forecasting) which are known from the language domain~\citep{brown2020language}.
However, despite their success in language, they often fall short of expectations in time series forecasting. For example, \citet{zengAreTransformersEffective2023} show that DLinear, a simple linear model, can outperform transformers in multiple scenarios.
Classical models, like LSTMs, are still widely used and remain competitive.
While LSTMs are well-suited for time series, they lack strong in-context learning capabilities.
Recent advancements in recurrent architectures --- such as the xLSTM~\citep{beckXLSTM7BRecurrent2025} --- closed this gap.
xLSTM shows promise in task-specific time series applications~\citep{kraus2024xlstm}, yet its potential for pre-trained, general-purpose models remains underexplored.

\section{\modelname}\label{sec:model}

\begin{figure}[th]
  \centering
  \includegraphics[width=\textwidth]{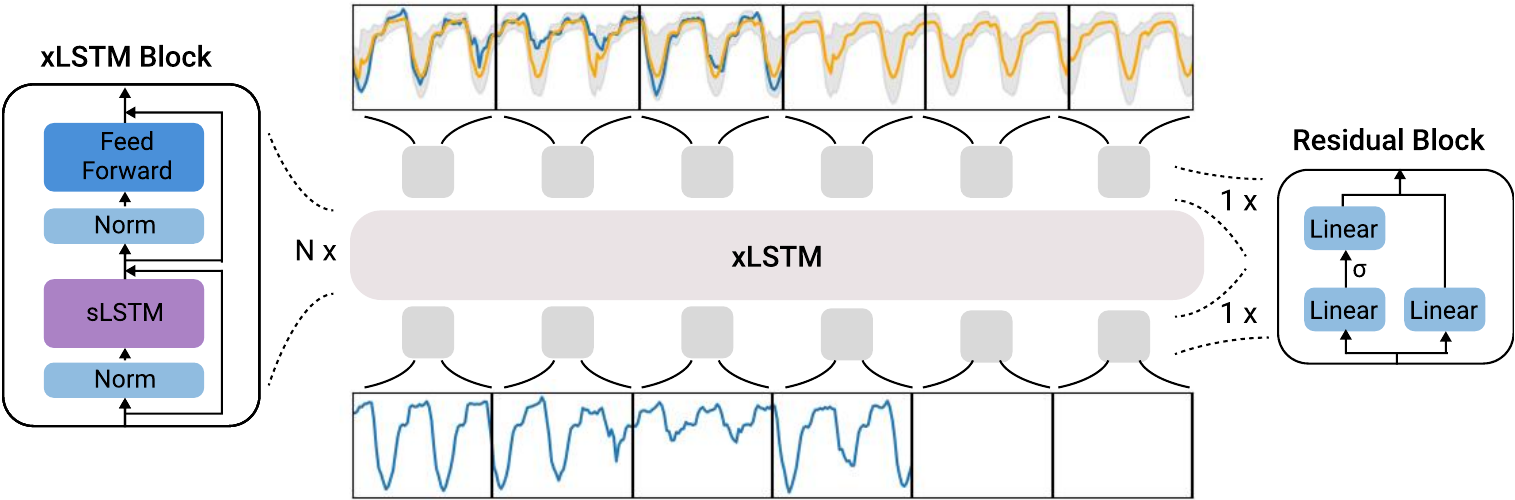} 
  \caption{Architecture overview of \modelname. The model comprises two main components: the xLSTM blocks and a residual block in the input and output layers. The illustrated forecast shows the forecasted series is in blue and the forecast of \modelname in orange. 
  During inference, only the last three output windows are of interest.}
  \label{fig:architecture}
\end{figure}

\modelname utilizes xLSTM as its backbone architecture, and adopts a decoder-only mode, which allows for efficient training.
It stacks multiple xLSTM blocks between a lightweight input and output layer.
The input layer preprocesses the time series via scaling and patching operations, producing tokens that are subsequently processed by the xLSTM blocks.
The output tokens correspond to forecasted patches of the target series and are mapped back to the forecast horizon.
For multi-patch forecasting, additional inputs are encoded as missing values.
An overview of the architecture is provided in Figure~\ref{fig:architecture} with individual components being described in detail below.

\paragraph{xLSTM Block}
\modelname adpots the block design proposed by \citet{beckXLSTM7BRecurrent2025}, but substitutes the mLSTM with a sLSTM module as the sequence mixing component.
Both module options were introduced in the original publication, but only sLSTM allows for state-tracking \citep[by trading it for reduced memory capacity][]{beckXLSTMExtendedLong2024a}.
Each block comprises a sLSTM module followed by a feed-forward network, with both components preceded by RMSNorm \citep{zhang2019root}.
Additionally, all sLSTM and feedforward layers include residual skip connections.
sLSTM supports real recurrence, to enable state-tracking, yet is still efficient in training and inference due to an optimized kernel architecture~\citep{poppelFlashRNNOAwareOptimization2024}.
\modelname stacks multiple of these blocks, depending on the model size.
After the last block, an additional RMSNorm is applied.
More details on the general xLSTM architecture are provided in Appendix~\ref{app:xlstm}.

\paragraph{Input/Output Layer and Loss}
\modelname is designed to generalize across diverse time series domains, which often exhibit significant variation in scale.
To ensure robustness, \modelname applies instance normalization to each time series \citep{kimReversibleInstanceNormalization2021}.
Specifically, $z$-score normalization is used. That is,  $\mathbf{\tilde{y}}_{0:T} = \frac{\mathbf{y}_{0:T} - \bar{y}_{0:T}}{\sigma_{y_{0:T}}}$, where, $\bar{y}$ and, $\sigma_y$ denote the mean and standard deviation of the time series sample.

\modelname segments time series into non-overlapping windows, and maps each window to the input space of the xLSTM using a two-layer residual block~\citep{heDeepResidualLearning2015, srivastavaTrainingVeryDeep2015}.
This patching mechanism is inspired by vision transformers~\citep{dosovitskiyImageWorth16x162020} and was adapted for time series by \citet{nieTimeSeriesWorth2022} and \citet{wooUnifiedTrainingUniversal2024a}.
It reduces the effective sequence length of the xLSTM blocks by a factor defined by the window size.
To account for missing values, a binary mask indicating presence or absence is concatenated to the time series values before the residual block.
Given an input window of size $\patchindim$ and xLSTM hidden dimension $\embdim$, the patching block defines a mapping $\mathbb{R}^{2\patchindim} \rightarrow \mathbb{R}^\embdim$.
The same residual block is shared across all time windows.

Mirroring the input layer, the decoder's output tokens are transformed back to the dimensions of the output-patch window using a residual block and subsequently scaled back to the original target space.
Hereby, the model outputs provides $|Q|$ quantile values for each time step of the output-patch window, rather than single-point predictions.
Hence, the output block defines a mapping $\mathbb{R}^{\embdim} \rightarrow \mathbb{R}^{\patchoutdim \times |Q|}$.
Specifically, \modelname  predicts nine equidistant quantile levels, $Q=\{0.1, 0.2, \dots, 0.9\}$.
The model's parameters are optimized by minimizing the quantile loss.
The loss is calculated for each output token, therefore, the loss does not distinguish context and forecast for a training sample, but implicitly ``forecast after each input token''.
Formally, the loss for an output window, given the true value $y_t$ at time $t$ and its corresponding quantile predictions $\hat{y}_t^q$ for quantile level $q$ is computed as:
\begin{equation}
    L = \frac{1}{|Q|\ \patchoutdim} \ \sum_{t=1}^{\patchoutdim} \ \sum_{q \in Q} \
    \begin{cases}
    q \ ( y_t - \hat{y}_t^q) & \text{if } \ \hat{y}_t^q \leq y_t \\
    (1 - q) \ (\hat{y}_t^q - y_t) & \text{else }
    \end{cases}\ .
\end{equation}
The losses of all output tokens of a training sample are averaged --- missing values in the output window are ignored for the loss calculation.

\paragraph{Multi-Patch Horizon Forecasts}
When the forecast horizon $h$ exceeds the output patch length, multiple future patches must be predicted.
We refer to this as multi-patch prediction.
Existing pre-trained models \citep{dasDecoderonlyFoundationModel2024e, ansariabdulfatirFastAccurateZeroshot2024} typically address multi-patch prediction via autoregressive generation, using point estimates (say, mean or median) of previous outputs as inputs for subsequent patches.
However, this approach reinitializes the probabilistic forecast at each step, disrupting the propagation of uncertainty.
In contrast, \modelname treats future inputs as missing values, allowing the internal memory to propagate both predictive state and uncertainty across patches.
This results in more coherent probabilistic and overall better forecasts, as our quantitative and qualitative experiments show (Section~\ref{sec:experiments} and Figure~\ref{fig:main-qualiative-comparison}).

\subsection{\fullpatchmask}\label{sec:fullpatchmask}
To facilitate the stable multi-patch prediction capability of \modelname, we propose \fullpatchmask (\fullpatchmaskabrr), illustrated in Figure~\ref{fig:datapipeline}.
\fullpatchmaskabrr randomly masks full and consecutive patches in pre-training.
Such a masked patch is represented as ``missing values'' in the model input, hence corresponds to the structure of the input when multi-patch forecasts are used in the inference.
The procedure is as follows:
For each training sample, we first uniformly sample the amount of consecutive patches $c_{\text{mask}} \sim U(1, \fullpatchmaxcons)$  and the masking probability $p_{\text{mask}}\sim U(0, \fullpatchmaxprob)$.
Afterwards, we mask the time series: For a time series of length $T$ we sample a binary mask of length $\lfloor\frac{T}{c_{\text{mask}} \patchoutdim}\rfloor$ with Bernoulli probability $p_{\text{mask}}$ and repeat each element $c_{\text{mask}} \cdot\patchoutdim$ so that the mask has a length of $T$ too.
Note that when neighboring elements are masked, the actual maximum of consecutive masked patches can be greater than $c_{\text{mask}}^{\text{max}}$.
Further, while \fullpatchmaskabrr incorporates elements from BERT-style \citep{devlin2019bert} masked-modeling, our training is still more similar to the typical causal-style masking of decoder-only approaches \citep{radford2018improving} since the target is shifted and the information flow is uni-directional.
Appendix~\ref{app:ext-ablation-results} provides a sensitivity analysis of the parameters $\fullpatchmaxprob$ and  $\fullpatchmaxcons$.

\section{Data Augmentation}\label{sec:datapipeline}

\begin{figure}
  \centering
  \includegraphics[width=\textwidth]{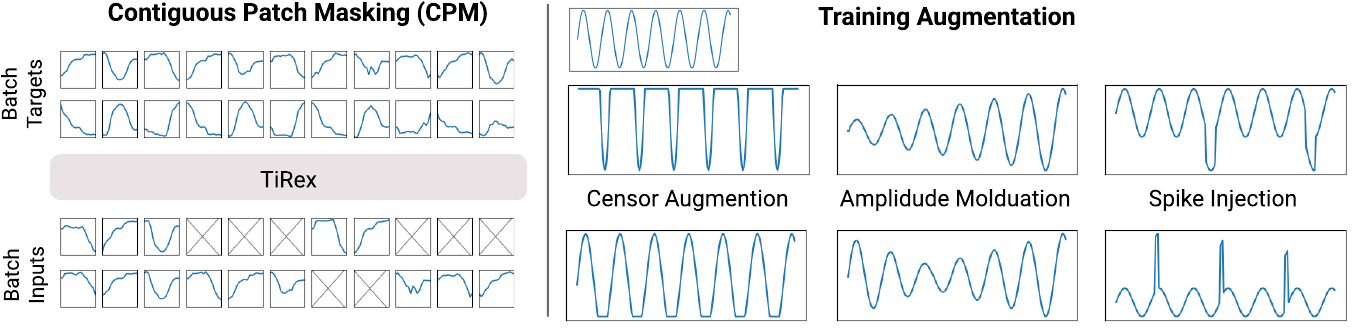} 
  \caption{Illustration of \fullpatchmask and the different training augmentations.}
  \label{fig:datapipeline}
\end{figure}

To facilitate more diverse time series patterns and enhance the model's exposure to a wider range of potentially relevant dynamics, we propose three augmentations for pre-training.
This is inspired by the successful application of augmentation techniques in pre-training of other modalities, e.g., vision \citep{tian2020makes}.
The employed augmentations, illustrated in Figure~\ref{fig:datapipeline}, are:
(1) \textit{Amplitude Modulation}, which introduces trends and change points in the scale of the time series.
Formally, a time series $\mathbf{y}$ is transformed by $y'_t = y_t \cdot a_t$, where $a_t$ follows a linear trend (potentially with change points).
(2) \textit{Censor Augmentation}, which censors values within the time series at a random threshold.
The augmented respective series is computed by $y'_t = \text{max/min}(y_t, c)$, 
The censor threshold $c$ is sampled by uniformly drawing a quantile from the empirical distribution of the signal.
(3) \textit{Spike Injection}, which adds short, periodic spike signals to the time series.
The augmented time series is computed by $y'_t = y_t + s_t$, where $s_t$ represents the added spike signals.
The shape of each spike is defined by a kernel, which can be a tophat, a radial basis function (RBF), or a linear kernel.
The periodicity and the specific parameters of the kernel (e.g., width, height for tophat; center, variance for RBF) are sampled from predefined distributions for each sample.
This randomization is designed to encourage the model to learn a general concept of transient events, rather than memorizing specific periodic patterns.
The augmentations are applied with a probability of $0.5$ for amplitude modulation and censor augmentation and $0.05$ for spike augmentation.
Appendix~\ref{app:augmentation} provides a more detailed description of the augmentation procedures. 

\section{Experiments}\label{sec:experiments}
This section outlines the experimental setup, including training procedures, evaluation benchmarks, and comparison models.
We further report the main results demonstrating the effectiveness of \modelname in general (Section~\ref{sec:zs-results}) and the specific components --- the xLSTM backbone, \fullpatchmask, and the proposed augmentations (Section~\ref{sec:ablation-results}).
Full experimental details are provided in Appendix~\ref{app:experiment-details}, while extended results are presented in Appendix~\ref{app:extended-results}.

\paragraph{\modelname Training}
Our training data comprises three components: (1) We utilize the training datasets from Chronos \citep{ansariChronosLearningLanguage2024b}, and adopt their TSMixup procedure to augment this data. 
(2) We enrich our training data with synthetic time series data generated through a procedure closely inspired by KernelSynth \citep{ansariChronosLearningLanguage2024b}.
(3) We add parts of the pre-training dataset proposed by GiftEval \citep{aksuGIFTEvalBenchmarkGeneral2024a}.
In total, our training dataset encompasses $47.5$ million time series samples.
Details regarding the specific datasets employed, as well as the implementation of the TSMixup and synthetic data generation procedures, can be found in Appendix~\ref{app:traing-data}.
During training, we augment the samples with the proposed training augmentations and apply \fullpatchmask as described in Section~\ref{sec:datapipeline}.
We train \modelname with a context length of $2048$ and a window size of $32$ for input and output patches.

\paragraph{Evaluation Benchmarks}
We evaluate \modelname on two standardized benchmarks with public leaderboards:
(1) the Chronos Zero-Shot Benchmark\footnote{\url{HuggingFace.co/spaces/autogluon/fev-leaderboard}}, comprising $27$ diverse datasets, each in one evaluation setting, primarily for short-term horizons \citep{ansariChronosLearningLanguage2024b}, and
(2) the GiftEval benchmark\footnote{\url{HuggingFace.co/spaces/Salesforce/GIFT-Eval}} \citep{aksuGIFTEvalBenchmarkGeneral2024a}, which includes $24$ datasets that are evaluated in different settings, and covers short-, medium-, and long-term horizons, as well as different frequencies --- in sum $97$ evaluation settings.
The training data of \modelname has no overlap with this data, hence \modelname operates in zero-shot.
We denote this benchmark as \textit{\chronoseval}.
To also ensure zero-shot conditions on the GiftEval benchmark, we exclude $16$ of the $97$ evaluation settings, which overlap with our training data, and denote this benchmark as \textit{\giftevalzs}; complete results for GiftEval are reported in Appendix~\ref{app:gift-eval-full}.
We note that all Chronos models do have the same overlap as \modelname; TimesFM, and Moirai have additional overlapping datasets and additionally also have overlaps with the Chronos-ZS benchmark.

The evaluation follows the respective benchmark protocols using mean absolute scaled error (MASE) for point forecast performance and the continuous ranked probability score (CRPS) for probabilistic forecast performance.
Practically, the CRPS is approximated by the mean weighted quantile loss (WQL) over nine quantiles: 0.1 to 0.9 in increments of 0.1\footnote{In the Chronos-ZS benchmark, the mean weighted quantile loss (WQL) is used directly, hence, essentially both benchmarks evaluate on the same metric}.
Aggregated performance is computed by normalizing each evaluation setting's score by that of a seasonal naive baseline, followed by the geometric mean across evaluation settings\footnote{GiftEval benchmark scores are reported based on the leaderboard computation at the time of our submission. A subsequent update to the seasonal naive baseline affects the absolute aggregated scores but does not change any discussed model ranking, result, or conclusion.}.
Additionally, the average rank across evaluation settings is reported to ensure robustness against outlier performance.

\paragraph{Compared Models}
We compare \modelname against a broad set of state-of-the-art models, including zero-shot pre-trained models and task-specific models.
The zero-shot models include Chronos and Chronos-Bolt~\citep{ansariChronosLearningLanguage2024b, ansariabdulfatirFastAccurateZeroshot2024}, TimesFM (v1.0, v2.0)~\citep{dasDecoderonlyFoundationModel2024e}, Moirai~1.1 \citep{wooUnifiedTrainingUniversal2024a}, TabPFN-TS~\citep{hooTabularFoundationModel2025}, and Tiny Time Mixer (TTM)~\citep{ekambaramTinyTimeMixers2024b}.
The task-specific models are PatchTST~\citep{nieTimeSeriesWorth2022}, TFT~\citep{limTemporalFusionTransformers2021}, DLinear~\citep{zengAreTransformersEffective2023}, DeepAR~\citep{salinasDeepARProbabilisticForecasting2020a}, and N-BEATS~\citep{oreshkinNBEATSNeuralBasis2019}.
These models are trained individually on each dataset. Hence, they do not operate in `zero-shot` and only provide an asymmetric comparison.
Reporting their result is useful to contextualize the current strengths and limitations of zero-shot approaches.
We report the results of the public leaderboards if available and replicate the outcomes of pre-trained models within our evaluation pipeline to ensure validity.

\subsection{Zero-Shot Forecasting}\label{sec:zs-results}

\begin{figure}
  \centering
  \includegraphics[width=\textwidth]{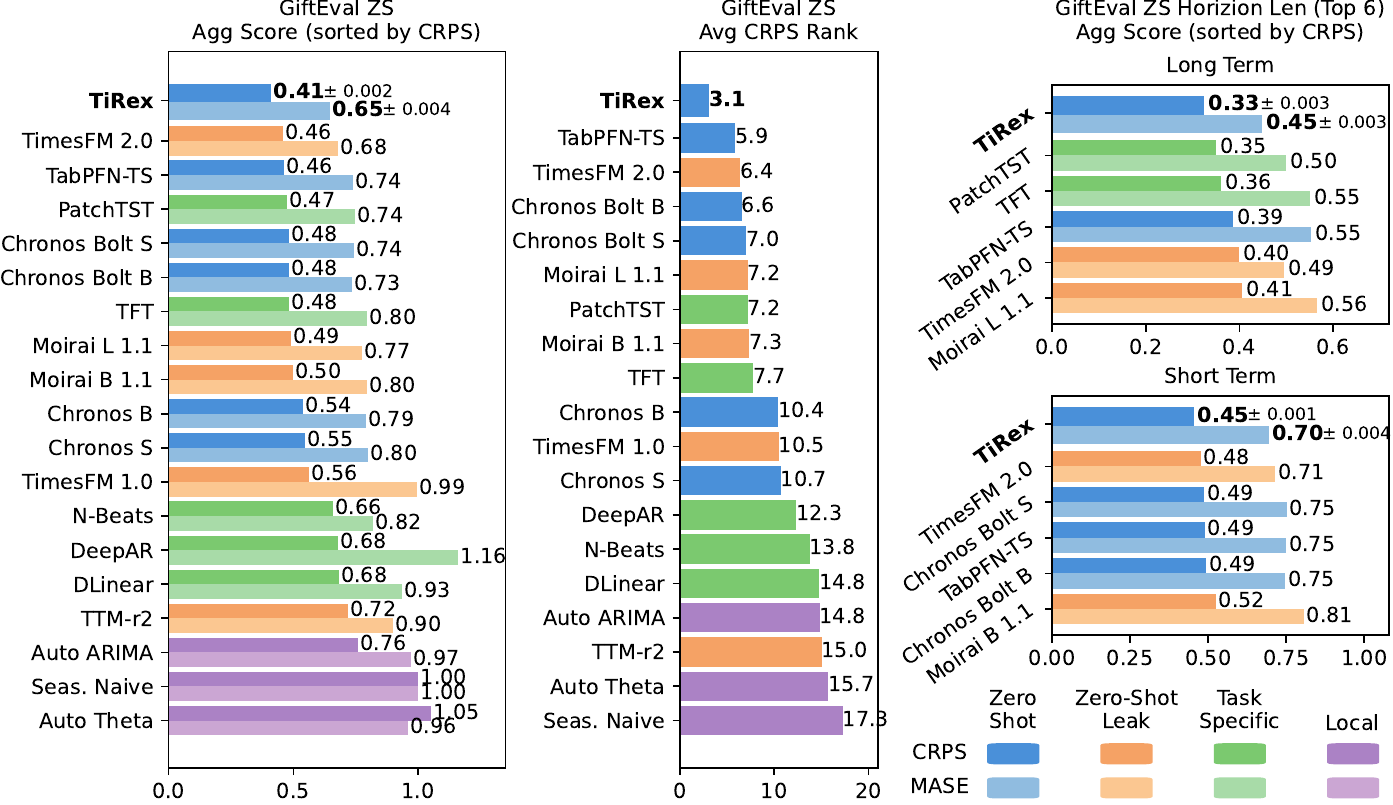} 
  \caption{Results of the \giftevalzs: Aggregated scores of the overall benchmark and the short- and long-term performances.
  Additionally, the average rank in terms of CRPS, as in the public leaderboard, is presented. 
  Lower values are better. ``Zero-shot Leak'' refers to models which are partly trained on the benchmark datasets (Overlap:: Moirai $19\%$, TimesFM $10\%$, TTM: $16\%$). We trained \modelname with 6 different seeds and report the observed standard deviation in the plot.}
  \label{fig:gift-eval-main-comp}
\end{figure}

\paragraph{GiftEval-ZS Benchmark}
In this benchmark \modelname consistently outperforms all competing methods across both short- and long-term forecasting tasks (Figure~\ref{fig:gift-eval-main-comp}). 
In terms of CRPS, \modelname achieves a score of $0.411$ (with standard deviation over training with 6 seeds of $\pm 0.002$), notably surpassing the next best zero-shot models ($0.459$, $0.463$, $0.481$).
This performance gap is also reflected in the average rank, where \modelname shows a substantial lead over the second-best model, while the three models that following in the ranking (TimesFM-2.0, TabPFN, and Chronos-Bolt-Base) achieve very similar scores among themselves. 
Importantly, while other models tend to peak either in short- (e.g., Chronos-Bolt) or long-term (e.g., TabPFN-TS) forecasting, \modelname is the only model to excel simultaneously at both. 
Moreover, \modelname attains these results with significantly fewer parameters (35M) compared to Chronos-Bolt-Base (200M) and TimesFM-2.0 (500M).
The advantage is most pronounced in long-term forecasting, where \modelname becomes the first zero-shot model to surpass the performance of PatchTST and TFT.

\begin{figure}
    \begin{minipage}[t]{0.5\textwidth}
      \includegraphics[width=\textwidth]{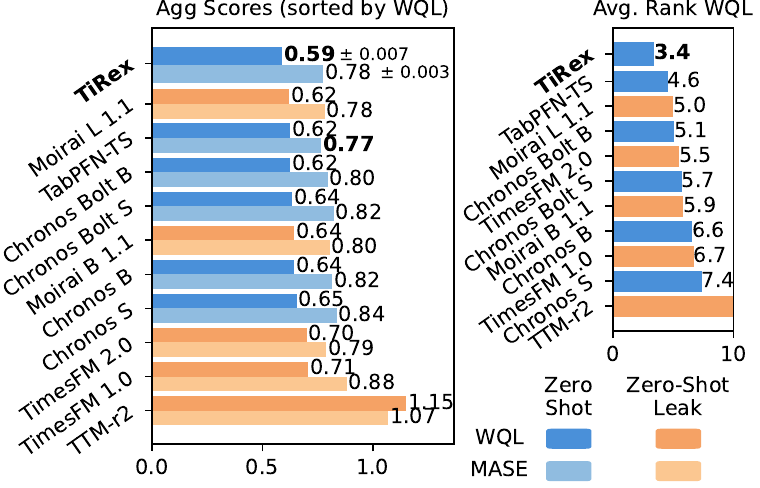} 
      \caption{Results of pre-trained model on the \chronoseval. The aggregated MASE and WQL scores, and the average rank in terms of WQL is shown. 
      Lower values are better. ``Zero-shot Leak'' refers to models which are partly trained on the benchmark datasets (Overlap: Moirai $82\%$, TimesFM $15\%$, TTM: $11\%$).}
      \label{fig:chronos-zs-main-comp}
    \end{minipage}
    \hfill
    \begin{minipage}[t]{0.47\textwidth} %
      \includegraphics[width=\textwidth]{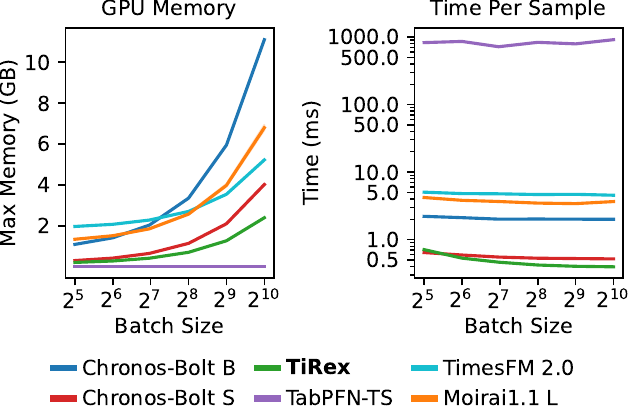} 
      \caption{Inference efficiency of the best-performing pre-trained models. Left: Relation between GPU memory and batch size. Right: Inference time per sample and batch size.}
      \label{fig:fulpatchmask}
    \end{minipage}
\end{figure}

\paragraph{Chronos-ZS Benchmark}
The main results on the \chronoseval exhibit similar performance patterns to those observed on the \giftevalzs (Figure~\ref{fig:chronos-zs-main-comp}).
\modelname again achieves the best results in terms of WQL score and rank.
In the MASE score \modelname has second best score, closely behind TabPFN-TS.
Notably, Moirai performs substantially better on the \chronoseval (compared to the \giftevalzs).
However, this improvement is likely due to the substantial overlap of $82\%$ between its pre-training data and the Chronos-ZS test set.
These results highlight the robustness of \modelname in zero-shot generalization.
The full results, including task-specific models are presented in Appendix~\ref{app:ext-zs-results}.

\paragraph{Qualitative Analysis}
We also qualitatively analyzed the predictions of \modelname and compared them against current state-of-the-art methods. 
Beyond its generally higher forecasting accuracy, the analysis shows that \modelname demonstrates robust multi-patch prediction capabilities, maintaining coherent uncertainty estimates across different forecast horizons --- and more accurately forecasts short periodic spikes, which are often missed or smoothed out by other models (Figure~\ref{fig:main-qualiative-comparison} and Appendix~\ref{app:ext-quali-results}). 
We hypothesize that these improvements mainly stem from (i) the effective multi-patch forecasting due to \fullpatchmaskabrr, (ii) the sLSTM architecture that provides state-tracking for uncertainty propagation and strong periodicity modeling, and (iii) the spike augmentation strategy enhancing the model's sensitivity to rare, sharp events during training.

\paragraph{Inference Speed \& Memory}
Apart from the forecasting performance, we analyze the GPU memory consumption and inference runtime across all models.
Specifically, we evaluate samples with a context length of $2048$ and a prediction length of $32$ over multiple batch sizes.
As expected, given \modelname substantially smaller size compared to the next best models (TimesFM-2.0 and Chronos-Bolt-Base), \modelname requires significantly less GPU memory and achieves faster inference speeds. 
Specifically, \modelname is over $11 \times$ faster than TimesFM-2.0, over $4 \times$ faster than Chronos-Bolt Base, and over $2176 \times$ faster than TabPFN-TS.
Furthermore, \modelname even outperforms Chronos-Bolt Small, a similarly sized transformer-based architecture for larger batch sizes.
The differences in maximum GPU memory consumption follow a similar order.

\subsection{Ablations}\label{sec:ablation-results}
We conduct ablation studies to analyze the impact of key components in \modelname.
Specifically, this section focuses on \fullpatchmask (\fullpatchmaskabrr), the proposed data augmentations, and the xLSTM backbone architecture.

\paragraph{\fullpatchmask and Multi-Patch-Inference}
We analyze the effectiveness of the proposed \fullpatchmask (\fullpatchmaskabrr) procedure by comparing three configurations: (1) standard decoder-only training with autoregressive inference --- this setting is similar e.g., to TimesFM; (2) training with a naive multiple-patch procedure where we place the ``rollout'' always at the end of the sequence; and (3) training with \fullpatchmaskabrr.
Configurations (2) and (3) use the same inference procedure with masked tokens for multi-step forecasting, whereas (1) relies on patch-wise autoregressive decoding. 
As shown in Table~\ref{tab:main-augment-comparision} only \fullpatchmaskabrr enables strong long-term performance without degrading short-term performance.
In contrast, na\"ive multi-patch training diminishes the short-term forecasting performance, and standard next token training combined with autoregressive inference harms the long-term forecasting performance.
These findings suggest that \fullpatchmaskabrr is essential for training and inference behaviors under multi-patch prediction.

\paragraph{Augmentation}
To assess the impact of our augmentations, we trained our model excluding each augmentation and with no augmentations at all.
The results, detailed in Table~\ref{tab:main-augment-comparision}, indicate that including the augmentations is beneficial, i.e., improves the performance of the model.
Performance consistently decreased in at least one benchmark metric when any single augmentation was removed.
The most substantial decline occurs when no augmentations were applied.
This underscores the effectiveness of each augmentation and their combined positive impact on the model's generalization capabilities.
Appendix~\ref{app:ext-ablation-results} provides a sensitivity analysis in terms of application probability.

\paragraph{Backbone}
To assess the architectural choice of xLSTM with sLSTM modules, we replace it with mLSTM \citep{beckXLSTMExtendedLong2024a} and transformer blocks \citep{touvron2023llama} while keeping the patching and training procedure unchanged.
For the transformer variant, rotary positional embeddings~\citep{su2024roformer} are added to mitigate the absence of inherent positional information, due to their permutation-invariance property. 
We also analyze mLSTM and sLSTM mix architectures as proposed in the original xLSTM paper.
We denote xLSTM[i:j] for an architecture where $i$ sLSTM blocks are combined with $j$ mLSTM blocks.
Additionally, we ablate our overall architecture by comparing it to a Chronos-Bolt architecture (Base and Small) that we train with our training procedure, using the same datasets and augmentations as for \modelname.
Table~\ref{tab:main-augment-comparision} summarizes the results.
\modelname with only sLSTM blocks yields the best performance, especially on long-term forecasts.
We hypothesize that this is because its explicit state-tracking capabilities \citep{beckXLSTMExtendedLong2024a}, which might facilitate uncertainty propagation and enable accurate modeling of periodic temporal structures over extended horizons.
Using only mLSTM performs worst.
However, switching just one of these blocks back to a sLSTM improves the results 
close to the sLSTM-only architecture of \modelname.
The comparison to a Chronos Bolt architecture, which is consistently outperformed by \modelname, highlights that our overall architecture is critical to achieve good performance on both long and short-term forecasting.
A more detailed comparison to the Chronos Bolt architecture is presented in Appendix~\ref{app:ext-ablation-results}.

\begin{table}[h!]
    \centering
    \caption{Ablation study of individual components. The top two rows report the mean and standard deviation of \modelname over six runs with different random seeds. For the ablation variants, results that degrade performance by more than 3$\times$ the standard deviation relative to \modelname are underlined. Columns correspond to evaluation settings: \giftevalzs (overall, short-term, and long-term) and \chronoseval. Lower values indicate better performance.}
    \label{tab:main-augment-comparision}
    \begin{tabular}{llrrrrrrrr}
        \toprule
        &  Benchmark & \multicolumn{2}{c}{Gift-ZS Overall} & \multicolumn{2}{c}{Gift-ZS Long} & \multicolumn{2}{c}{Gift-ZS Short} & \multicolumn{2}{c}{Chronos-ZS} \\
        &  & {CRPS} & {MASE}  & {CRPS} & {MASE}  & {CRPS} & {MASE}  & {WQL} & {MASE} \\
        \midrule
        & \modelname & 0.411 & 0.647 & 0.325 & 0.45 & 0.455 & 0.696 & 0.592 & 0.776 \\
        & \small{$\pm 6$ seeds} & \small{0.002} & \small{0.004} & \small{0.003} &\small{0.003} & \small{0.001} & \small{0.004} &\small{0.007} & \small{0.003} \\ 
         \midrule
        \multirow{2}{*}{\rotatebox[origin=c]{90}{CPM}} 
        &\begin{tabular}[r]{@{}l@{}}na\"ive\\ multi-patch \end{tabular} & \sigB{0.424} & \sigB{0.662} & \sigB{0.335} & \sigB{0.460} & \sigB{0.475} & \sigB{0.718} & \sigB{0.650} & \sigB{0.817} \\
        & w/o multi-patch & \sigB{0.445} & \sigB{0.704} & \sigB{0.370} & \sigB{0.518} & \sigB{0.471} & \sigB{0.719} & $0.589$ & $0.777$ \\
        \midrule
        \multirow{5}{*}{\rotatebox[origin=c]{90}{Augment}}
        & w/o any & \sigB{0.430} & \sigB{0.682} & \sigB{0.339} & \sigB{0.478} & \sigB{0.473} & \sigB{0.722} & \sigB{0.623} & \sigB{0.800} \\
        & w/o censor & $0.417$ & $0.652$ & \sigB{0.336} & $0.457$ & $0.458$ & $0.699$ & $0.595$ & $0.767$ \\
        & w/o spike & $0.415$ & \sigB{0.660} & $0.328$ & $0.459$ & \sigB{0.462} & \sigB{0.710} & $0.591$ & $0.773$ \\
        &\begin{tabular}[r]{@{}l@{}}w/o amplidude \\ modulation \end{tabular}
        & $0.409$ & $0.644$ & $0.323$ & $0.448$ & $0.455$ & $0.694$ & \sigB{0.618} & \sigB{0.798} \\
        \midrule
        \multirow{6}{*}{\rotatebox[origin=c]{90}{Backbone}} & Transformer & \sigB{0.422} & \sigB{0.662} & \sigB{0.342} & \sigB{0.472} & \sigB{0.461} & $0.702$ & $0.597$ & $0.768$ \\
        & mLSTM & \sigB{0.457} & \sigB{0.718} & \sigB{0.430} & \sigB{0.589} & $0.456$ & $0.699$ & $0.588$ & $0.775$ \\        
        & xLSTM[1:11] & $0.414$ & $0.652$ & $0.330$ & $0.456$ & $0.455$ & $0.698$ & \sigB{0.631} & \sigB{0.807} \\
        & xLSTM[1:5] & $0.412$ & $0.651$ & $0.330$ & \sigB{0.460} & \sigL{0.450} & $0.693$ & $0.611$ & \sigB{0.791} \\[0.5mm]
        \cdashline{2-10}\noalign{\vskip 1mm}
        & Chronos Bolt S & \sigB{0.456} & \sigB{0.676} & \sigB{0.413} & \sigB{0.498} & \sigB{0.463} & $0.705$ & $0.609$ & \sigB{0.791} \\        
        & Chronos Bolt B & \sigB{0.454} & \sigB{0.670} & \sigB{0.418} & \sigB{0.493} & $0.458$ & $0.701$ & \sigB{0.627} & \sigB{0.807} \\       
        \bottomrule
    \end{tabular}
\end{table}

\section{Conclusion} \label{sec:conclusion}

This work introduces \modelname, a pre-trained time series forecasting model based on xLSTM.
To fully unlock the state-tracking capabilities of xLSTM, we further propose \fullpatchmask, a training-time masking strategy tailored for in-context learning.
\fullpatchmask is a crucial component in our modeling pipeline, since it enables strong long-term forecasting performance without sacrificing short-term capabilities.
\modelname establishes a new state-of-the-art in zero-shot forecasting, outperforming prior methods on both short- and long-term horizons across the Chronos-ZS and GiftEval benchmarks.
Our ablation studies highlight the individual contributions of each component to overall performance.

\paragraph{Limitations \& Future Work}

Like most pre-trained forecasting models, \modelname focuses on univariate time series.
Although modeling multivariate series as independent univariate signals often performs well --- as reflected in the GiftEval results and, for example, shown in \citet{nieTimeSeriesWorth2022} --- future work could incorporate multivariate data, for example in the form of extended contexts or modified input layers.
Due to computational constraints, we did not extensively tune hyperparameters and only conducted a sensitivity analysis on key parameters.
Future work should explore more comprehensive tuning for additional performance gains and investigate leveraging the model’s learned representations for other downstream tasks, such as classification~\citep{auer2025pretrained} or anomaly detection.

\begin{ack}
We thank Maximilian Beck and Korbinian Pöppel for the discussions and advice with regard to xLSTM.
We thank Elias Bürger, Bernhard Voggenberger, Marco Obermeier, and Levente Zolyomi for their efforts in providing an updated TiRex 1.1.
We thank Martin Loretz for making TiRex run efficiently on CPU.
The ELLIS Unit Linz, the LIT AI Lab, and the Institute for Machine Learning are supported by the Federal State Upper Austria.
\end{ack}

\clearpage

\bibliography{bib/references_assets, bib/references_ts, bib/references_mlinstitute, bib/references_xlstm, bib/references_misc, bib/sepp}
\bibliographystyle{ml-institute} %

\newpage

\appendix

{
\hypersetup{linkcolor=black}
\addcontentsline{toc}{section}{Appendix} %
\part{Appendix} %
\parttoc %
}

\section{xLSTM}\label{app:xlstm}

\begin{figure}[ht]
    \centering
    \includegraphics[width=0.8\textwidth]{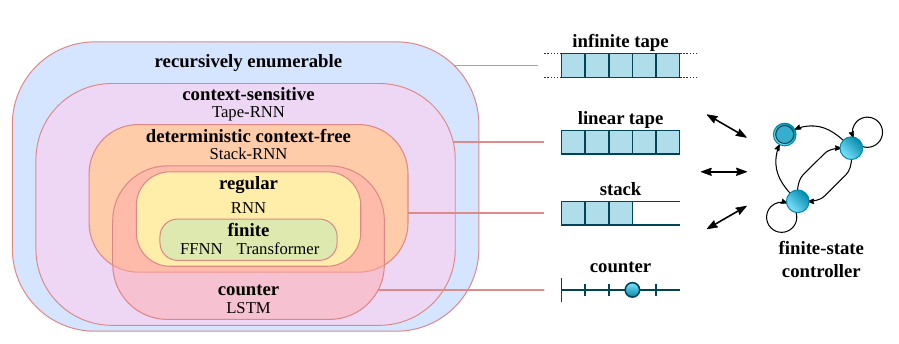}
    \vspace{5mm}
\begin{tabular}{@{}lll@{}}
    \toprule
    \textbf{Grammar type (low $\to$ high)} & \textbf{Automaton} & \textbf{Memory} \\
    \midrule
    Regular (R)  & Finite-state automaton (FSA)   & Automaton state \\
    Context-free (CF)        & Push-down automaton (PDA)   & $+$ infinite stack  \\
    Context-sensitive (CS)       & Linear bounded automaton (LBA)   & $+$ bounded tape  \\
    Recursively enumerable (RE)       & Turing machine (TM)    & $+$ infinite tape \\
    \bottomrule
    \end{tabular}
    \caption{ Formal language classes and their correspondence with neural network architectures and $k$-counter machines that have a counting mechanism \citep[from:][]{Deletang:23}.
    }
    \label{fig:chomsky_new}
\end{figure}

\modelname utilizes xLSTM~\citep{beckXLSTMExtendedLong2024a} as its backbone architecture. 
xLSTM extends the classical LSTM~\citep{Hochreiter:91a,Hochreiter:97} by incorporating modern design principles to improve its scalability, parallelization, and in-context modeling capabilities.

xLSTM introduces two cell types:
the matrix LSTM (mLSTM) and the scalar LSTM (sLSTM).
The mLSTM is designed to increase memory capacity through a matrix-based memory representation, and enables efficient parallel computation.
In contrast, the sLSTM preserves a true recurrent pathway as in the original LSTM, enabling strong state-tracking capabilities \citep{beckXLSTMExtendedLong2024a}.
The recurrent pathway makes LSTM more expressive than State Space Models (SSMs), parallelizable RNNs like RWKV, and transformers~\citep{Merrill:23,Merrill:24,Deletang:23}.
Figure~\ref{fig:chomsky_new} illustrates the respective expressivity hierarchies.
We hypothesize that this advantage in expressivity allows \modelname to better model complex temporal dynamics, leading to improved forecasting performance, especially over long horizons.
\citet{Deletang:23} demonstrate on synthetic language tasks that this state-tracking capability of LSTMs yields empirical advantages. \citet{beckXLSTMExtendedLong2024a} show that sLSTM retains these advantages.
sLSTM employs a multi-head strategy along the recurrent pathway to improve efficiency.

\modelname exclusively employs sLSTM cells.
Given a sequence of $T$ embedded input patches $\BX_{1:T} = (\Bx_1, \Bx_2, \dots, \Bx_T ) \in \reals^{d \times T}$, the forward computation of an sLSTM cells for a given time step is defined as follows:

\begin{align}
\Bc_t \ &= \  \bff_t \odot \Bc_{t-1} \ + \ \bfi_t \odot \Bz_t \ , &  & &\text{cell state} \\
\Bn_t \ &= \  \bff_t \odot \Bn_{t-1} \ + \ \bfi_t  \ , &  & &\text{normalizer state} \\
\Bh_t  \ &= \ \bfo_t \odot \tilde{\Bh}_t \ , 
  & \tilde{\Bh}_t \ &= \ \Bc_t  \odot \Bn_t ^{-1}
  &\text{hidden state} \\
\Bz_t \ &= \ \varphi \left( \tilde{\Bz}_t \right) \ , 
  &\tilde{\Bz}_t \ &=  \ \BW_{\Bz} \ \Bx_t \ + \
  \BR_{\Bz}  \ \Bh_{t-1} \ + \  \Bb_{\Bz} \ \
  &\text{cell input} \\
\bfi_t\ &= \  \exp \left( \tilde{\bfi}_t  \right) \ , 
  &\tilde{\bfi}_t \ &= \ \BW_{\bfi} \ \Bx_t \ + \
  \BR_{\bfi}  \ \Bh_{t-1} \ + \  \Bb_{\bfi} \ \
  &\text{input gate} \\
\bff_t \ &= \ \exp \left( \tilde{\bff}_t \right) \ ,
  &\tilde{\bff}_t \ &= \ \BW_{\bff} \ \Bx_t  \ + \
  \BR_{\bff}  \ \Bh_{t-1} \ + \  \Bb_{\bff} \ \
  &\text{forget gate} \\
\bfo_t \ &= \ \sigma \left( \tilde{\bfo}_t \right) \ , 
  &\tilde{\bfo}_t  \ &= \ \BW_{\bfo} \ \Bx_t \ + \
  \BR_{\bfo}  \ \Bh_{t-1} \ + \  \Bb_{\bfo} \ \
  &\text{output gate},
\end{align}

where
$\Bh_{t} \in \reals^{d}$ denotes the hidden state, 
$\Bc_t \in \reals^{d}$ denotes the cell states and, $\Bn_{t} \in \reals^d$ denotes a normalizer state.
Further, $\bfi_t ,\bfo_t, \bff_t \in \reals^d$ are the input, 
output and forget gate, respectively,
$\BW_{\Bz}, \BW_{\bfi},\BW_{\bff},\BW_{\bfo} \in \reals^{d \times D}$, 
$\BR_{\Bz}, \BR_{\bfi},\BR_{\bff},\BR_{\bfo} \in \reals^{d \times d}$,
and $\Bb_{\Bz}, \Bb_{\bfi}, \Bb_{\bff}, \Bb_{\bfo} \in \reals^d$
are trainable weight matrices and biases.  
The matrices $\BR_{\Bz}$, $\BR_{\bfi}$, $\BR_{\bff}$, $\BR_{\bfo}$ are block-diagonal, where each block represents one head.
This way, the parameters reduce to $d^2/(N_h)$, where $N_h$ is the number of heads, limiting the cell interactions to individual heads.
The input-, output-, and forget-gates are activated by exponential ($\exp$) or sigmoid functions ($\sigma$);
The cell inputs use a hyperbolic tangent function ($\varphi$).

To enable deep modeling, xLSTM organizes its recurrent layers into blocks that combine sLSTM and/or mLSTM layers with additional architectural components.
Specifically, \modelname uses the block structure from~\citet{beckXLSTM7BRecurrent2025} that consists of

\begin{enumerate}
    \item an sLSTM module
    \item a feed forward network
    \item residual connections around each subcomponent,
    \item and pre-normalization layers (RMSNorm)
\end{enumerate}

This block architecture is illustrated in Figure~\ref{fig:architecture} and allows for the training of deep networks.
To achieve scalability, xLSTM employs custom CUDA kernels that enable high-throughput training and inference on modern hardware \citep{poppelFlashRNNOAwareOptimization2024}.

\newpage

\section{Data Augmentation}\label{app:augmentation}

This section details the time series augmentations proposed and used for pre-training \modelname.

\paragraph{Amplitude Modulation}
This augmentation introduces scale trends and change points into the time series by multiplying the signal with a piecewise linear trend.
The modulation trend is generated by sampling change points and interpolating amplitudes between them. See Algorithm~\ref{alg:amp-mod}.

\paragraph{Censor Augmentation}
This augmentation censors (clips) the input signal either from below or above, depending on a randomly sampled direction. 
The clipping threshold is determined by drawing a quantile uniformly from the empirical distribution of the signal. See Algorithm~\ref{alg:censor-aug}.

\paragraph{Spike Injection}
This augmentation injects a structured additive signal in the form of sparse, periodic spikes.
First, a periodic pattern is sampled, which is then tiled across the time axis with a sampled periodicity.
Each spike label in the pattern is mapped to a kernel (selected from a fixed set) with randomized parameters that control its shape and magnitude.
The final augmentation is the sum of all such kernel evaluations added to the original signal. See Algorithm~\ref{alg:spike-aug}.
The goal of this augmentation is to improve generalization to sharp, transient events by exposing the model to a diversity of spike structures.
The spike kernel types and their parameter ranges are defined in Table~\ref{tab:spike-kernels}, while the available temporal patterns and their sampling probabilities are described in Table~\ref{tab:spike-patterns}.

\begin{algorithm}
    \caption{Amplitude Modulation}
    \label{alg:amp-mod}
    \KwIn{Time series $\mathbf{y} \in \mathbb{R}^T$}
    \KwOut{Augmented time series $\mathbf{y_{\text{aug}}}$}
    $k \sim \mathrm{Uniform}(0, 5)$ \tcp*[r]{Number of changepoints} 
    Sample $k$ changepoints $\{c_1, \dots, c_k\} \subset \{1, \dots, T-1\}$ \;
    $\mathbf{c} \gets [0, \text{sorted}(\{c_1, \dots, c_k\}), T]$ \;
    Sample amplitudes $\mathbf{a} \sim \mathcal{N}(1, 1)^{k+2}$ \;
    Interpolate trend $\mathbf{t} \in \mathbb{R}^T$ from $(\mathbf{c}, \mathbf{a})$ \;
    $\mathbf{y_{\text{aug}}} \gets \mathbf{y} \odot \mathbf{t}$ \;
    \Return $\mathbf{y_{\text{aug}}}$
\end{algorithm}

\begin{algorithm}
    \caption{Censor Augmentation}
    \label{alg:censor-aug}
    \KwIn{Time series $\mathbf{y} \in \mathbb{R}^T$}
    \KwOut{Augmented time series $\mathbf{y_{\text{aug}}}$}
    Sample quantile level $q \sim \mathrm{Uniform}(0, 1)$ \;
    Compute threshold $c \gets \mathrm{Quantile}(\mathbf{y}, q)$ \;
    Sample censor direction $b \sim \mathrm{Bernoulli}(0.5)$ \tcp*[r]{Bottom or top censor}
    \eIf{$b = 1$}{
        $\mathbf{y_{\text{aug}}} \gets \max(y_t, c)$ for all $t$ \tcp*[r]{Bottom censoring}
    }{
        $\mathbf{y_{\text{aug}}} \gets \min(y_t, c)$ for all $t$ \tcp*[r]{Top censoring}
    }
    \Return $\mathbf{y_{\text{aug}}}$
\end{algorithm}

\begin{algorithm}
    \caption{Spike Injection}
    \label{alg:spike-aug}
    \KwIn{Time series $\mathbf{y} \in \mathbb{R}^T$ spike patterns $\mathcal{P}$, spike kernel set $\mathcal{K}$}
    \KwOut{Augmented time series $\mathbf{y_{\text{aug}}}$}
    Sample periodicity $\pi \sim \mathcal{U}(10, \min(512, T))$ \;
    Sample periodic pattern $z \in \mathcal{P}$ and shift it randomly \;
    Sample $s \sim \mathcal{U}(T - \pi, T)$ \;
    Generate spike positions $\mathbf{m} \in \mathbb{Z}^T$ by repeating $z$ with spacing $\pi$ and shifting to align last spike at $s$ \;
    Sample kernel type $\kappa \in \mathcal{K}$ \;
    For each unique spike label in $\mathbf{m}$, sample kernel parameters and generate kernel centered at each occurrence \;
    Sum kernels to obtain additive spike signal $\mathbf{s} \in \mathbb{R}^T$ \;
    $\mathbf{y_{\text{aug}}} \gets \mathbf{y} + \mathbf{s}$ \;
    \Return $\mathbf{y_{\text{aug}}}$
\end{algorithm}

\begin{table}[h!]
    \centering
    \caption{Kernel types and parameterizations (as functions of the periodicity $\pi$) employed for the spike injection augmentations}
    \label{tab:spike-kernels}
    \begin{tabular}{lll}
    \toprule
    \textbf{Kernel} & \textbf{Width Param}  & \textbf{Amplidue Param} \\
    \midrule
    Tophat & $w \sim [0.05\pi, 0.2\pi]$ & $h \sim [0.5, 3]$ \\
    RBF & $\sigma_{\text{RBF}} \sim [0.05\pi, 0.2\pi]$ & $h \sim [0.5, 3]$ \\
    Linear & $w \sim [0.05\pi, 0.2\pi]$ & $h \sim [0.5, 3]$  \\
    \bottomrule
    \end{tabular}
\end{table}

\begin{table}[h!]
    \centering
    \caption{Spike pattern, representative patterns, and sampling probabilities employed for the spike injection augmentations.}
    \label{tab:spike-patterns}
    \begin{tabular}{lll}
    \toprule
    \textbf{Category} & \textbf{Utilized Pattern} & \textbf{Sample Probability} \\
    \midrule
    Simple         & $[0]$, $[0,1]$                       & 0.75 \\
    3-periodic     & $[0,1,2]$, $[0,0,1]$                 & 0.10 \\
    4-periodic     & $[0,0,1,1]$, $[0,1,0,2]$             & 0.10 \\
    Weekly-like    & $[0,0,0,0,0,1,1]$, $[0,0,0,0,0,1,2]$ & 0.05 \\
    \bottomrule
    \end{tabular}
\end{table}

\clearpage

\section{Experiment Details}\label{app:experiment-details}

\subsection{Model and Training Hyperparameter}

\modelname utilizes a xLSTM-based architecture with hyperparameters summarized in Table~\ref{tab:model-hparams}.
The model has 35 million model parameters.

\paragraph{Pre-Training}
\modelname is pre-trained for $500{,}000$ steps with a batch size of $256$ using the AdamW optimizer with a learning rate of $0.001$ and weight decay of $0.01$. We also employ a cosine learning rate scheduler with linear warm-up (the warm-up ratio is set to 5 \% and the minimum learning rate is $0.0001$).
The context of \modelname length is $2048$.

\begin{table}[ht]
\centering
\caption{\modelname model architecture hyperparameters.}
\begin{tabular}{@{}ll@{}}
\toprule
\textbf{Parameter} & \textbf{Value} \\
\midrule
Input patch size ($\patchindim$)         & 32 \\
Output patch size ($\patchoutdim$)       & 32 \\
Embedding dimension ($\embdim$)          & 512 \\
Feed-forward dimension ($\ffdim$)        & 2048 \\
Number of heads                          & 4 \\
Number of blocks                         & 12 \\
\bottomrule
\end{tabular}
\label{tab:model-hparams}
\end{table}

\subsection{Pre-Training Data Corpus}\label{app:traing-data}
We construct a diverse training corpus by combining real and synthetic time series to support robust generalization across heterogeneous forecasting tasks.
Our training dataset has three components:
\begin{enumerate}
    \item \textbf{Chronos Training Data} ($30$ million time series):
    We incorporate the training datasets from Chronos and adopt their proposed time series mixup augmentation strategy~\citep{ansariChronosLearningLanguage2024b}.
    In contrast to Chronos, we generate significantly more and longer time series, expanding the diversity and temporal span of the data.
    Table~\ref{tab:chronos-trainig-data} lists the respective datasets, which are provided on HuggingFace: \url{https://HuggingFace.co/datasets/autogluon/chronos_datasets}.
    Details on the TsMixup procedure are provided in the paragraph below.
    \item \textbf{Synthetic Gaussian Process Data} ($15$ million time series):
    Inspired by KernelSynth~\citep{ansariChronosLearningLanguage2024b}, we generate synthetic time series using Gaussian Processes (GPs).
    Details on the synthetic data generation procedure are provided in the paragraph below.
    \item \textbf{GiftEval Pre-training Data} ($\approx 2.5$ million time series):
    We integrate a subset of the pre-training corpus from GiftEval and use it for pre-training. It does not overlap with the GiftEval benchmark evaluation data.
    Table~\ref{tab:gifteval-pretraining-data} lists the respective datasets, which are provided on HuggingFace: \url{https://HuggingFace.co/datasets/Salesforce/GiftEvalpre-train}.
\end{enumerate}

\paragraph{Data Mix Probabilities}
For the Chronos training data and the synthetic Gaussian process data, each time series is sampled with equal probability.
The GiftEval Pre-training data is sampled with a probability of approximately $8 \%$, hence slightly oversampled compared to the share of series due to technical implementation details.

\paragraph{TsMixup}
To augment data diversity, we apply TsMixup~\citep{ansariChronosLearningLanguage2024b}, a convex combination of $k$ time series of length $l$ to the training data from Chronos.
Each series is $z$-score normalized (Chronos used mean normalization) prior to combination to ensure comparable magnitude.
The number $k$ is sampled uniformly from $\{1, \dots, K_{\max}\}$, the length $l$ is sampled uniformly from $[L_{\min}, L_{\max}]$, and the mixing weights $\lambda_i$ are drawn from a Dirichlet distribution:
\begin{equation}
    \mathbf{x}^{\text{mix}}_{1:l} = \sum_{i=1}^{k} \lambda_i \cdot \mathbf{\tilde{x}}^i_{1:l},
\end{equation}
where $\mathbf{\tilde{x}}_i \in \mathbb{R}^{l}$ is a normalized time series segment, and $\boldsymbol{\lambda} \sim \text{Dir}(\alpha)$. 
As $k=1$ is sampled with non-zero probability, the augmented dataset includes original sequences, thereby preserving base data fidelity while enhancing variability.
In our procedure we utilize $K_{\max}=4$, $L_{\min}=128$, $L_{\min}=4096$, and $\alpha=1.5$; we generate $30$ million time series.

\paragraph{Synthetic GP Data}

We generate synthetic time series using Gaussian Processes (GPs), building upon the core ideas of KernelSynth~\citep{ansariChronosLearningLanguage2024b}. 
Each synthetic time series $\mathbf{x} \in \mathbb{R}^{L_{\text{syn}}}$ is sampled from a GP:
\begin{equation}
    \mathbf{x} \sim \mathcal{GP}(0, \tilde{\kappa}(t, t')),
\end{equation}
where $\tilde{\kappa}(t, t')$ is a composite kernel constructed by randomly sampling and combining kernels from a kernel bank $\mathcal{K}$.
We sample $j \sim U\{1, J\}$ base kernels (with replacement) from $\mathcal{K}$ and combine them using random binary operations from $\{+, \times\}$ to obtain $\tilde{\kappa}$.
Kernel parameters (e.g., length scale, periodicity) are sampled from predefined priors.
In our procedure, we utilize $J=4$ and $L_{\text{syn}}=4096$; Our kernel bank $\mathcal{K}$ includes periodic, Radial Basis Function (RBF), Rational Quadratic (RQ), and Piecewise Polynomial kernels.
We generate $15$ million time series with this procedure.

In contrast to KernelSynth, we introduce the following adaptations:
\begin{enumerate}
    \item We sample periodicities from both fixed sets (as KernelSynth) and additionally from continuous distributions to increase temporal diversity.
    \item We employ a more scalable GP sampler with GPU support and approximations for longer series~\citep{gardner2018gpytorch}.
    This enables the generation of more and longer sequences.
    \item We use a modified kernel bank.
\end{enumerate}

\begin{table}[ht]
\centering
\caption{Training Datasets published by \cite{ansariChronosLearningLanguage2024b} --- (\url{https://huggingface.co/datasets/autogluon/chronos_datasets}) --- that were utilized to train \modelname.}
\small
\label{tab:chronos-trainig-data}
\begin{tabular}{l r r}
\toprule
Name & \#Series & Avg. Length \\
\midrule
Mexico City Bikes & 494 & 78313 \\
Brazilian Cities Temperature & 12 & 757 \\
Solar (5 Min.) & 5166 & 105120 \\
Solar (Hourly) & 5166 & 105120 \\
Spanish Energy and Weather & 66 & 35064 \\
Taxi (Hourly) & 2428 & 739 \\
USHCN & 6090 & 38653 \\
Weatherbench (Hourly) & 225280 & 350639 \\
Weatherbench (Daily) & 225280 & 14609 \\
Weatherbench (Weekly) & 225280 & 2087 \\
Wiki Daily (100k) & 100000 & 2741 \\
Wind Farms (Hourly) & 100000 & 8514 \\
Wind Farms (Daily) & 100000 & 354 \\
Electricity (15 Min.) & 370 & 113341 \\
Electricity (Hourly) & 321 & 26304 \\
Electricity (Weekly) & 321 & 156 \\
KDD Cup 2018 & 270 & 10897 \\
London Smart Meters & 5560 & 29951 \\
M4 (Daily) & 4227 & 2371 \\
M4 (Hourly) & 414 & 901 \\
M4 (Monthly) & 48000 & 234 \\
M4 (Weekly) & 359 & 1035 \\
Pedestrian Counts & 66 & 47459 \\
Rideshare & 2340 & 541 \\
Taxi (30 Min.) & 2428 & 1478 \\
Temperature-Rain & 32072 & 725 \\
Uber TLC (Hourly) & 262 & 4344 \\
Uber TLC (Daily) & 262 & 181 \\
\bottomrule
\end{tabular}
\end{table}

\begin{table}[ht]
\centering
\caption{Subset of pre-training datasets published by \cite{aksuGIFTEvalBenchmarkGeneral2024a} ---  (\url{https://huggingface.co/datasets/Salesforce/GiftEvalPretrain}) --- that were utilized to train \modelname.}
\small
\label{tab:gifteval-pretraining-data}
\begin{tabular}{l r r}
\toprule
Name & \#Series & Avg. Length \\
\midrule
azure vm traces 2017 & 159472 & 5553 \\
borg cluster data 2011 & 143386 & 3749 \\
bdg-2 panther & 105 & 8760 \\
bdg-2 fox & 135 & 17219 \\
bdg-2 rat & 280 & 16887 \\
bdg-2 bear & 91 & 16289 \\
lcl & 713 & 13385 \\
smart & 5 & 19142 \\
ideal & 217 & 5785 \\
sceaux & 1 & 34223 \\
borealis & 15 & 5551 \\
buildings 900k & 1795256 & 8761 \\
largest 2017 & 8196 & 105120 \\
largest 2018 & 8428 & 105120 \\
largest 2019 & 8600 & 105120 \\
largest 2020 & 8561 & 105408 \\
largest 2021 & 8548 & 105120 \\
PEMS03 & 358 & 26208 \\
PEMS04 & 307 & 16992 \\
PEMS07 & 883 & 28224 \\
PEMS08 & 170 & 17856 \\
PEMS BAY & 325 & 52128 \\
LOS LOOP & 207 & 34272 \\
BEIJING SUBWAY 30MIN & 276 & 1572 \\
SHMETRO & 288 & 8809 \\
HZMETRO & 80 & 2377 \\
Q-TRAFFIC & 45148 & 5856 \\
subseasonal & 862 & 16470 \\
subseasonal precip & 862 & 11323 \\
wind power & 1 & 7397147 \\
solar power & 1 & 7397222 \\
kaggle web traffic weekly & 145063 & 114 \\
kdd2022 & 134 & 35280 \\
godaddy & 3135 & 41 \\
favorita sales & 111840 & 1244 \\
china air quality & 437 & 13133 \\
beijing air quality & 12 & 35064 \\
residential load power & 271 & 538725 \\
residential pv power & 233 & 537935 \\
cdc fluview ilinet & 75 & 852 \\
cdc fluview who & 74 & 564 \\
\bottomrule
\end{tabular}
\end{table}

\clearpage

\subsection{Benchmarks and Metrics}

We evaluated our models on two standardized benchmarks, GiftEval~\citep{aksuGIFTEvalBenchmarkGeneral2024a} and the Chronos Zero-Shot benchmark \citep{ansariChronosLearningLanguage2024b}. Both are hosted on public HuggingFace leaderboards. 
These benchmarks offer transparent, reproducible evaluations across a wide range of datasets, domains, and forecast horizons, enabling a comprehensive assessment of generalization capabilities.
The benchmarks not only specify the datasets and forecast horizons but also metric computations, and provide results of state-of-the-art models, ensuring valid baseline comparisons.

\paragraph{GiftEval.}
GiftEval~\citep{aksuGIFTEvalBenchmarkGeneral2024a} comprises 23 datasets totaling over 144,000 time series, spanning seven domains, ten sampling frequencies, and a wide range of forecast horizons from short- to long-term.
In sum the benchmark evaluates $97$ different evaluation settings.
The benchmark includes evaluations of 17 models, covering classical statistical methods, deep learning approaches, and recent foundation models, including all pre-trained models relevant to our study.

In total, $16$ out of the $97$ evaluation settings in GiftEval overlap with those used for pre-training in our work (i.e., the Chronos pre-train collection).
To ensure comparability while avoiding data leakage, we restrict our main evaluation to non-overlapping datasets.
We denote this benchmark as \textbf{\giftevalzs}.
This is possible because of the dataset-level granularity of the leaderboard submissions, which allows custom aggregations of results while preserving fidelity to the original benchmark.
Full benchmark results, including the overlapping datasets, are reported in Appendix~\ref{app:gift-eval-full}.
The individual datasets and evaluation settings of the benchmark are listed in Table~\ref{tab:chronos-zs-benchmark}.
For additional details, please refer to \citet{aksuGIFTEvalBenchmarkGeneral2024a}.

GiftEval uses the Mean Absolute Scaled Error (MASE) for point forecasts and the Continuous Ranked Probability Score (CRPS) for probabilistic forecasts as performance metrics.
Equation~\ref{eq:mase} defines the MASE: $\hat{y}_t$ is the point forecast, $y_t$ is the observed value at time $t$.
MASE scales the error by a na\"ive seasonal forecast, given the seasonal period $s$.
Equation~\ref{eq:crps} respectively defines the CRPS: $F(u)$ is the predictive cumulative distribution and $\mathbf{1}\{ y_t \leq u \}$ is the indicator function for the observed value. 
To evaluate performance over a full forecast horizon, the CRPS is averaged across time steps.
In practice, CRPS is approximated by computing the average weighted quantiles loss over a fixed set of quantile levels $Q=\{0.1, 0.2, \dots, 0.9\}$, with $\hat{y}_t^q$ as the quantile prediction of quantile $q$ at time step $t$ (see Equation~\ref{eq:crps-approx}).

\begin{equation} \label{eq:mase}
\text{MASE} = 
\frac{\frac{1}{h} \sum_{t=T+1}^{T+h} |\hat{y}_t - y_t|}
     {\frac{1}{h} \sum_{t=s+1}^{T} |y_t - y_{t - s}|}
\end{equation}

\begin{equation} \label{eq:crps}
\text{CRPS} = \frac{1}{h} \sum_{t=T+1}^{T+h} \int_{-\infty}^{\infty} \left( F(u) - \mathbf{1}\{ y_t \leq u \} \right)^2 \, du
\end{equation}

\begin{align} \label{eq:crps-approx}
\text{CRPS} &\approx \frac{1}{|Q| \cdot h} \sum_{q \in Q}    \frac{2 \sum_{t=T+1}^{T+h}\text{QL}(q, \hat{y}_t^{q}, y_t)}{\sum_{t=T+1}^{T+h}|y_t|} \\
\text{QL}(q, \hat{y}^q, y_t) &=  \begin{cases}
    q ( y_t - \hat{y}_t^q) & \text{if } \hat{y}_t^q \leq y_t \\
    (1 - q) (\hat{y}_t^q - y_t) & \text{else }.
    \end{cases}
\end{align}

Before aggregation, the metric values of both metrics are normalized per dataset using a seasonal na\"ive baseline to mitigate scale effects.
The aggregated metric scores are computed using the geometric mean of these normalized scores.
Additionally, the average rank of the CRPS across evaluation settings is reported to increase robustness against outlier results.
\textbf{GiftEval benchmark scores are reported based on the leaderboard computation at the time of the submission. A subsequent update to the seasonal naive baseline affects the absolute aggregated scores but does not change any discussed model ranking, result, or conclusion.}

\paragraph{\chronoseval.}
The \chronoseval \citep{ansariChronosLearningLanguage2024b} consists of 27 datasets, with a focus on short-term forecasting.
\modelname's pre-training data has no overlap with \chronoseval, hence we can use it to extend the assessment of its zero-shot capabilities. 
The evaluation metrics are identical in structure to GiftEval: MASE for point forecasts and Weighted Quantile Loss (WQL) for probabilistic forecasts, with WQL evaluated over the same set of quantiles, making it computationally equivalent to the CRPS approximation of GiftEval.
Aggregation procedures, including baseline normalization and geometric mean computation, are also consistent across both benchmarks.
The datasets and settings of the benchmark are presented in Table~\ref{tab:chronos-zs-benchmark}; for additional details, please refer to \citet{ansariChronosLearningLanguage2024b}.

\begin{table}
\centering
\caption{GiftEval \citep{aksuGIFTEvalBenchmarkGeneral2024a} benchmark datasets and evaluation settings. Evaluation settings that are part of \giftevalzs are marked in the respective column. Forecast Horizion is abbreviated as ``Hor'' and the number of evaluated windows is abbreviated as ``Win''.}
\label{tab:gifteval-data}
\small
\begin{tabular}{l l r r | rr rr rr}
\toprule
\multirow{2}{*}{Name} & \rotatebox{90}{GiftEval-ZS} & \rotatebox{90}{Freq} & \rotatebox{90}{\#Series} & \multicolumn{2}{c}{Short} & \multicolumn{2}{c}{Medium} & \multicolumn{2}{c}{Long} \\
& & & & Hor & Win & Hor & Win & Hor & Win \\
\midrule
bitbrains\_fast\_storage & x &5T & 1250 & 48 & 18 & 480 & 2 & 720 & 2 \\
bitbrains\_fast\_storage & x &H & 1250 & 48 & 2 & - & - & - & - \\
bitbrains\_rnd & x &5T & 500 & 48 & 18 & 480 & 2 & 720 & 2 \\
bitbrains\_rnd & x &H & 500 & 48 & 2 & - & - & - & - \\
bizitobs\_application & x &10S & 1 & 60 & 15 & 600 & 2 & 900 & 1 \\
bizitobs\_l2c & x &5T & 1 & 48 & 20 & 480 & 7 & 720 & 5 \\
bizitobs\_l2c & x &H & 1 & 48 & 6 & 480 & 1 & 720 & 1 \\
bizitobs\_service & x &10S & 21 & 60 & 15 & 600 & 2 & 900 & 1 \\
car\_parts & x &M & 2674 & 12 & 1 & - & - & - & - \\
covid\_deaths & x &D & 266 & 30 & 1 & - & - & - & - \\
electricity &   &15T & 370 & 48 & 20 & 480 & 20 & 720 & 20 \\
electricity & x &D & 370 & 30 & 5 & - & - & - & - \\
electricity &   &H & 370 & 48 & 20 & 480 & 8 & 720 & 5 \\
electricity &   &W & 370 & 8 & 3 & - & - & - & - \\
ett1 & x &15T & 1 & 48 & 20 & 480 & 15 & 720 & 10 \\
ett1 & x &D & 1 & 30 & 3 & - & - & - & - \\
ett1 & x &H & 1 & 48 & 20 & 480 & 4 & 720 & 3 \\
ett1 & x &W & 1 & 8 & 2 & - & - & - & - \\
ett2 & x &15T & 1 & 48 & 20 & 480 & 15 & 720 & 10 \\
ett2 & x &D & 1 & 30 & 3 & - & - & - & - \\
ett2 & x &H & 1 & 48 & 20 & 480 & 4 & 720 & 3 \\
ett2 & x &W & 1 & 8 & 2 & - & - & - & - \\
hierarchical\_sales & x &D & 206 & 30 & 4 & - & - & - & - \\
hierarchical\_sales & x &W & 118 & 8 & 4 & - & - & - & - \\
hospital & x &M & 767 & 12 & 1 & - & - & - & - \\
jena\_weather & x &10T & 1 & 48 & 20 & 480 & 11 & 720 & 8 \\
jena\_weather & x &D & 1 & 30 & 2 & - & - & - & - \\
jena\_weather & x &H & 1 & 48 & 19 & 480 & 2 & 720 & 2 \\
kdd\_cup\_2018 &   &D & 270 & 30 & 2 & - & - & - & - \\
kdd\_cup\_2018 &   &H & 270 & 48 & 20 & 480 & 2 & 720 & 2 \\
loop\_seattle & x &5T & 323 & 48 & 20 & 480 & 20 & 720 & 15 \\
loop\_seattle & x &D & 323 & 30 & 2 & - & - & - & - \\
loop\_seattle & x &H & 323 & 48 & 19 & 480 & 2 & 720 & 2 \\
m4\_daily &   &D & 4227 & 14 & 1 & - & - & - & - \\
m4\_hourly &   &H & 207 & 48 & 2 & - & - & - & - \\
m4\_monthly &   &M & 2400 & 18 & 20 & - & - & - & - \\
m4\_quarterly & x &Q & 24000 & 8 & 1 & - & - & - & - \\
m4\_weekly &   &W & 359 & 13 & 1 & - & - & - & - \\
m4\_yearly & x &A & 22974 & 6 & 1 & - & - & - & - \\
m\_dense & x &D & 30 & 30 & 3 & - & - & - & - \\
m\_dense & x &H & 30 & 48 & 20 & 480 & 4 & 720 & 3 \\
restaurant & x &D & 403 & 30 & 2 & - & - & - & - \\
saugeen & x &D & 1 & 30 & 20 & - & - & - & - \\
saugeen & x &M & 1 & 12 & 7 & - & - & - & - \\
saugeen & x &W & 1 & 8 & 20 & - & - & - & - \\
solar & x &10T & 137 & 48 & 20 & 480 & 11 & 720 & 8 \\
solar & x &D & 137 & 30 & 2 & - & - & - & - \\
solar & x &H & 137 & 48 & 19 & 480 & 2 & 720 & 2 \\
solar & x &W & 137 & 8 & 1 & - & - & - & - \\
sz\_taxi & x &15T & 156 & 48 & 7 & 480 & 1 & 720 & 1 \\
sz\_taxi & x &H & 156 & 48 & 2 & - & - & - & - \\
temperature\_rain &   &D & 32072 & 30 & 3 & - & - & - & - \\
us\_births & x &D & 1 & 30 & 20 & - & - & - & - \\
us\_births & x &M & 1 & 12 & 2 & - & - & - & - \\
us\_births & x &W & 1 & 8 & 14 & - & - & - & - \\
\bottomrule

\end{tabular}
\end{table}

\begin{table}
\caption{\chronoseval benchmark datasets \citep{ansariChronosLearningLanguage2024b} and evalution setting.}
\small
\label{tab:chronos-zs-benchmark}
\centering
\begin{tabular}{l r r}
\toprule
Name & Horizon & Periodicty \\
\midrule
traffic & 24 & 24 \\
australian electricity & 48 & 48 \\
ercot & 24 & 24 \\
ETTm & 24 & 96 \\
ETTh & 24 & 24 \\
exchange rate & 30 & 5 \\
nn5 & 56 & 1 \\
nn5 weekly & 8 & 1 \\
weather & 30 & 1 \\
covid deaths & 30 & 1 \\
fred md & 12 & 12 \\
m4 quarterly & 8 & 4 \\
m4 yearly & 6 & 1 \\
dominick & 8 & 1 \\
m5 & 28 & 1 \\
tourism monthly & 24 & 12 \\
tourism quarterly & 8 & 4 \\
tourism yearly & 4 & 1 \\
car parts & 12 & 12 \\
hospital & 12 & 12 \\
cif 2016 & 12 & 12 \\
m1 yearly & 6 & 1 \\
m1 quarterly & 8 & 4 \\
m1 monthly & 18 & 12 \\
m3 monthly & 18 & 12 \\
m3 yearly & 6 & 1 \\
m3 quarterly & 8 & 4 \\
\bottomrule
\end{tabular}\end{table}

\subsection{Computation \& Hardware}\label{app:hardware}

We conducted all experiments on Nvidia A40 and H100 GPUS --- A40's provide enough GPU memory to conduct all training runs.
The inference experiments (GPU memory and Inference Speed) were conducted on a Nivida A40 GPU.
CPU requirements are flexible; we utilized a 64-core Xeon(R) Platinum 8358.

\clearpage

\section{Extended Results}\label{app:extended-results}
This section presents additional experimental results complementing Section~\ref{sec:experiments}.
The structure is the same as in the main paper, preceded by results for the full GiftEval benchmark: 
First, extended results on the zero-shot evaluation of \giftevalzs and \chronoseval are presented, followed by inference efficiency results of all pre-trained models, extended ablation studies results, and additional qualitative examples.
Additionally, we provide fine-tuning results.

\subsection{Full GiftEval leaderboard}\label{app:gift-eval-full}
\begin{figure}
  \centering
  \includegraphics[width=\textwidth]{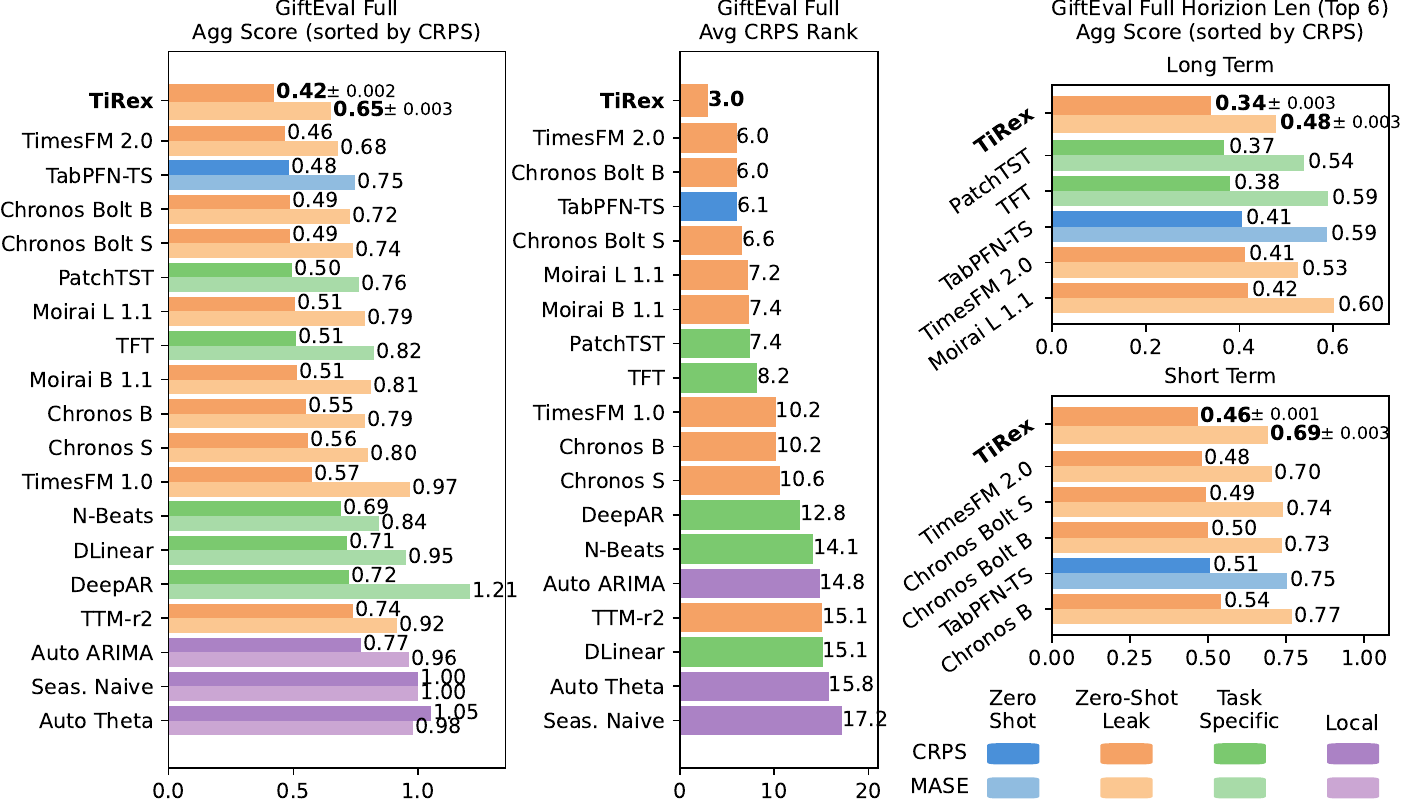} 
  \caption{Results of the \textbf{full GiftEval benchmark}: Aggregated scores of the overall benchmark and the short- and long-term performances.
  Additionally, the average rank in terms of CRPS, as in the public leaderboard, is presented. 
  Lower values are better. ``Zero-shot Leak'' refers to models which are partly trained on the benchmark datasets. We trained \modelname with 6 different seeds and report the observed standard deviation in the plot.}
  \label{fig:gift-eval-full-comparision}
\end{figure}

Figure~\ref{fig:gift-eval-full-comparision} presents the evaluation results on the full GiftEval benchmark, including settings excluded from \giftevalzs\ due to training data overlap with \modelname.
These results align with those reported on the HuggingFace GiftEval leaderboard.
The results are consistent with the trends observed in the main \giftevalzs\ evaluation.
\modelname outperforms all baseline models by a substantial margin, with the largest performance gap observed in the long-term forecasting tasks.

\subsection{Zero-Shot Forecasting}\label{app:ext-zs-results}

\begin{figure}[ht]
  \centering
  \includegraphics[width=\textwidth]{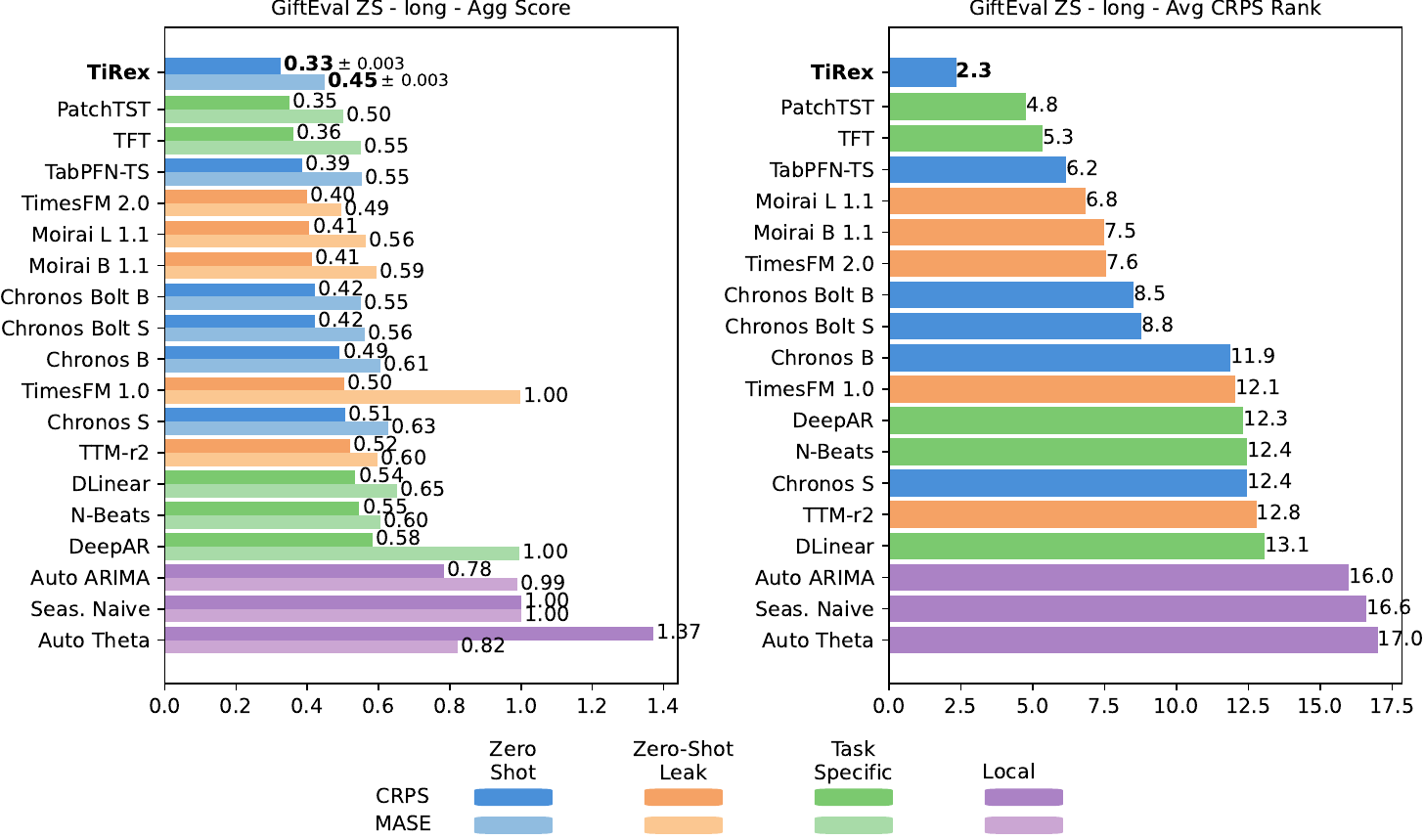} 
  \caption{Results of the \giftevalzs: The aggregated scores and the average CRPS rank of the benchmarks' \textbf{long-term} sub-results are shown.
  Lower values are better. ``Zero-shot Leak'' refers to models that are partly trained on the benchmark datasets. We trained \modelname with 6 different seeds and report the observed standard deviation of the aggregated scores.}
  \label{fig:gift-eval-app-long}
\end{figure}

\begin{figure}[ht]
  \centering
  \includegraphics[width=\textwidth]{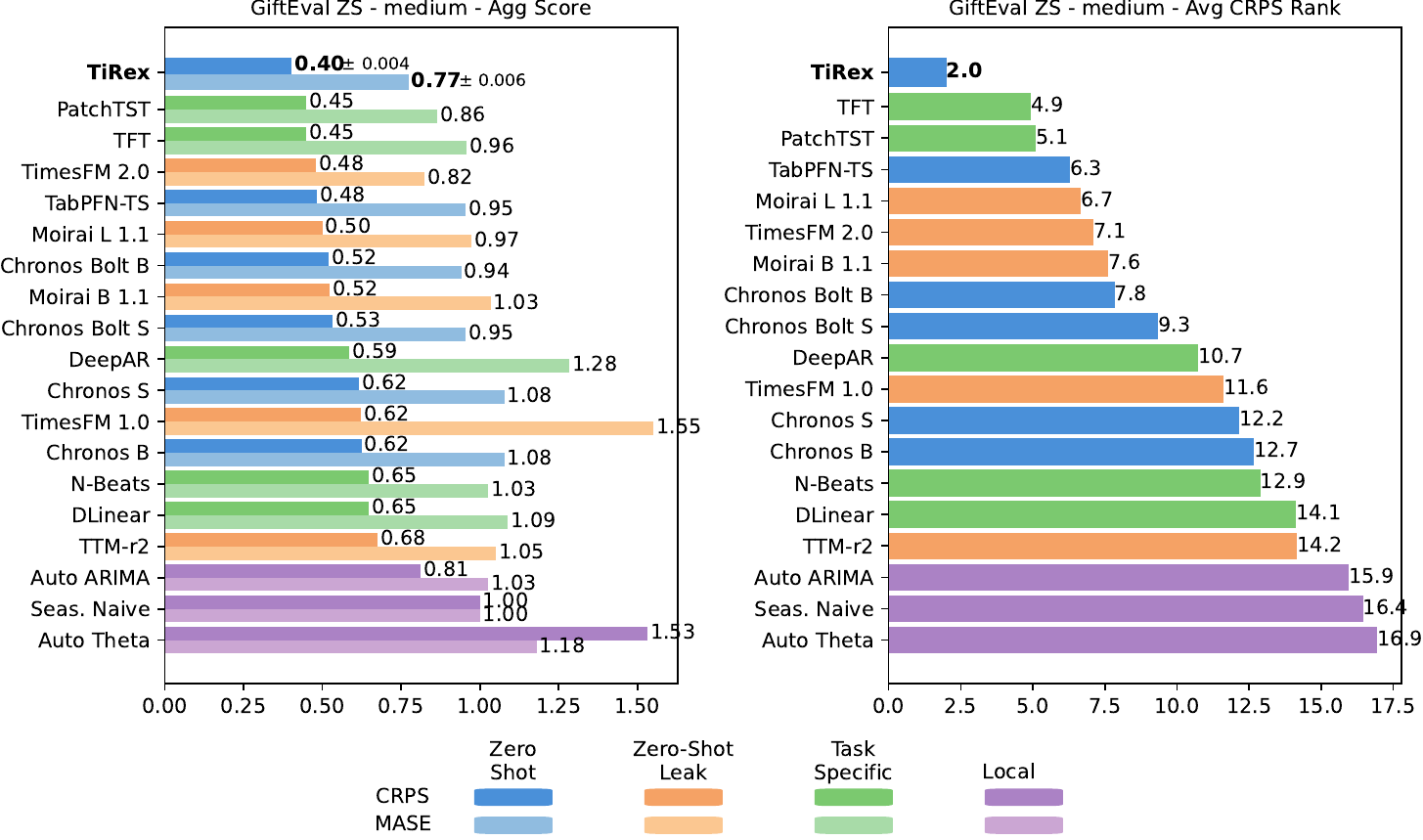} 
  \caption{Results of the \giftevalzs: The aggregated scores and the average CRPS rank of the benchmarks' \textbf{medium-term} sub-results are shown.
  Lower values are better. ``Zero-shot Leak'' refers to models that are partly trained on the benchmark datasets. We trained \modelname with 6 different seeds and report the observed standard deviation of the aggregated scores.}
  \label{fig:gift-eval-app-medium}
\end{figure}

\begin{figure}[ht]
  \centering
  \includegraphics[width=\textwidth]{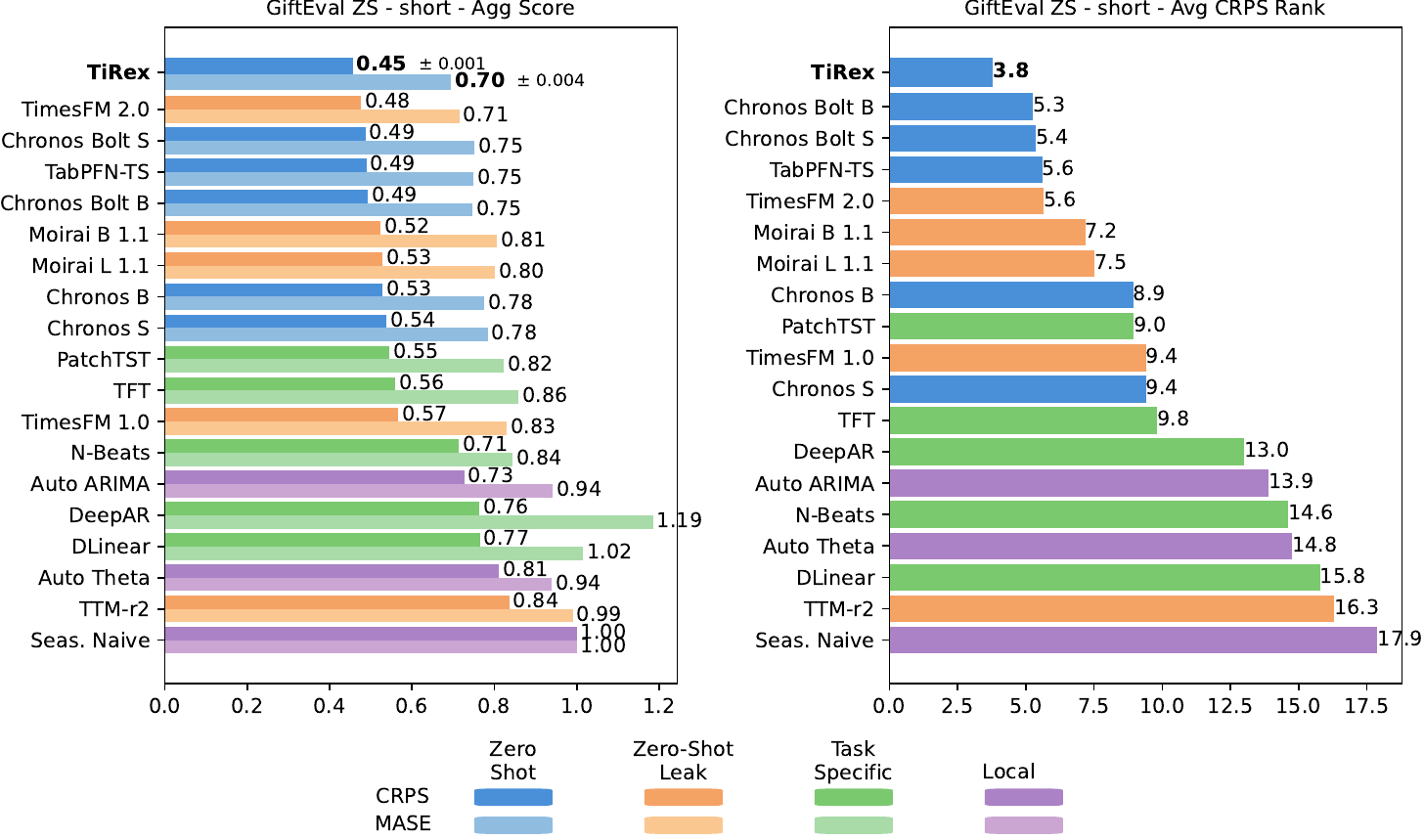} 
  \caption{Results of the \giftevalzs: The aggregated scores and the average CRPS rank of the benchmarks' \textbf{short-term} sub-results are shown.
  Lower values are better. ``Zero-shot Leak'' refers to models that are partly trained on the benchmark datasets. We trained \modelname with 6 different seeds and report the observed standard deviation of the aggregated scores.}
  \label{fig:gift-eval-app-short}
\end{figure}

\paragraph{\giftevalzs}
Figures~\ref{fig:gift-eval-app-long}–\ref{fig:gift-eval-app-short} present the short-, medium-, and long-term evaluation sub-results for all models.
Both aggregated scores and average CRPS rank metrics are reported.
As discussed in the main text, \modelname consistently achieves the best performance across all settings.
The results of the individual evaluation settings are reported in the Tables~\ref{tab:gifteval-zs-zs-model-MASE-p1}-\ref{tab:gifteval-zs-taskspec-WQL-p2}.

\begin{figure}[ht]
  \centering
  \includegraphics[width=\textwidth]{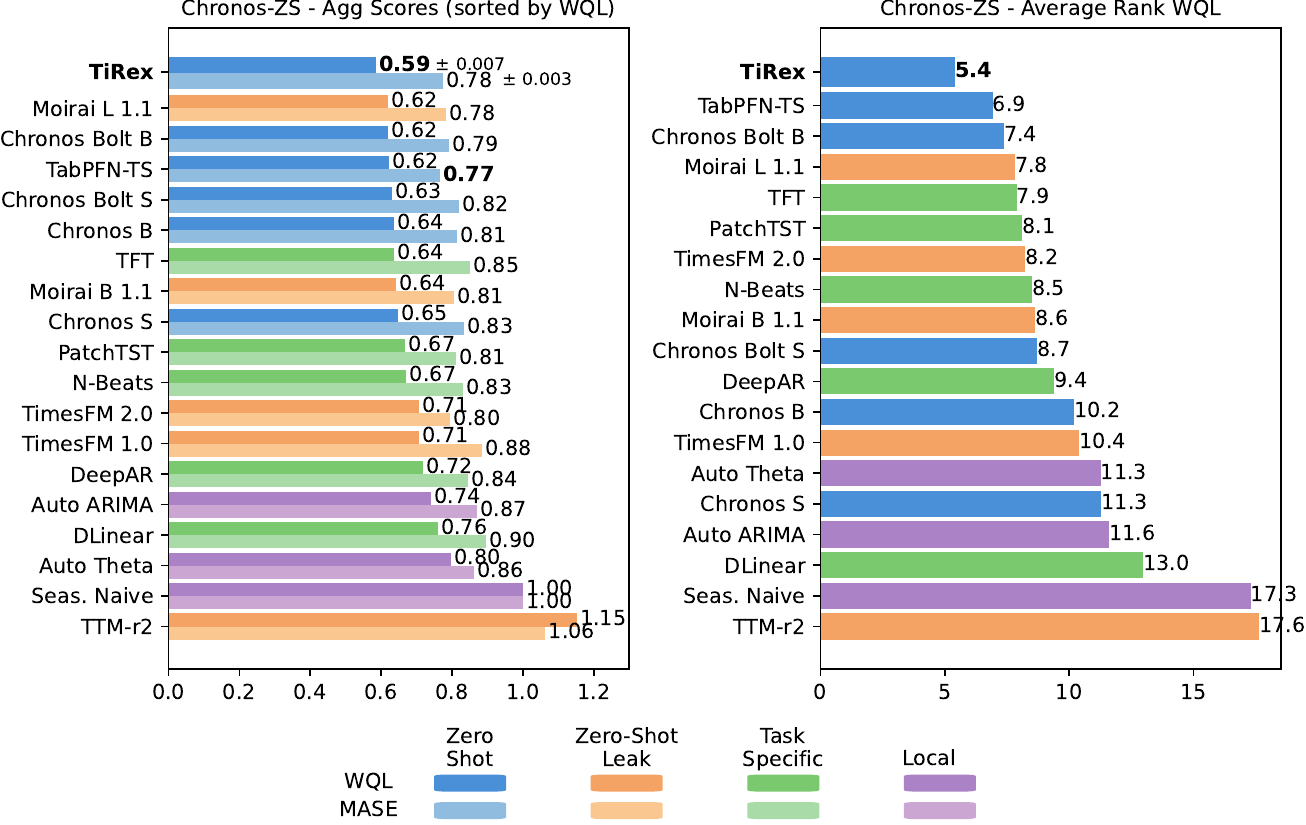} 
  \caption{Results of the \chronoseval: The aggregated scores and the average WQL rank.
  Lower values are better. ``Zero-shot Leak'' refers to models that are partly trained on the benchmark datasets (Overlap: Moirai $82\%$, TimesFM $15\%$, TTM: $11\%$). We trained \modelname with 6 different seeds and report the observed standard deviation of the aggregated scores.}
  \label{fig:chronos-zs-app}
\end{figure}

\paragraph{\chronoseval}
Figure~\ref{fig:chronos-zs-app} extends the Chronos benchmark results by including task-specific and local models not covered by the official leaderboard, which reports only pre-trained models. 
These additional results are taken from the benchmark’s original publication~\citep{ansariChronosLearningLanguage2024b}.
Consistent with the main paper, \modelname performs best and even outperforms models with substantial training data overlap and those explicitly trained for individual datasets.
The results of the individual evaluation settings are reported in the Tables~\ref{tab:chronos-zs-zs-model-MASE}-\ref{tab:chronos-zs-zs-model-WQL}.

\begin{figure}[ht]
  \centering
  \includegraphics[width=\textwidth]{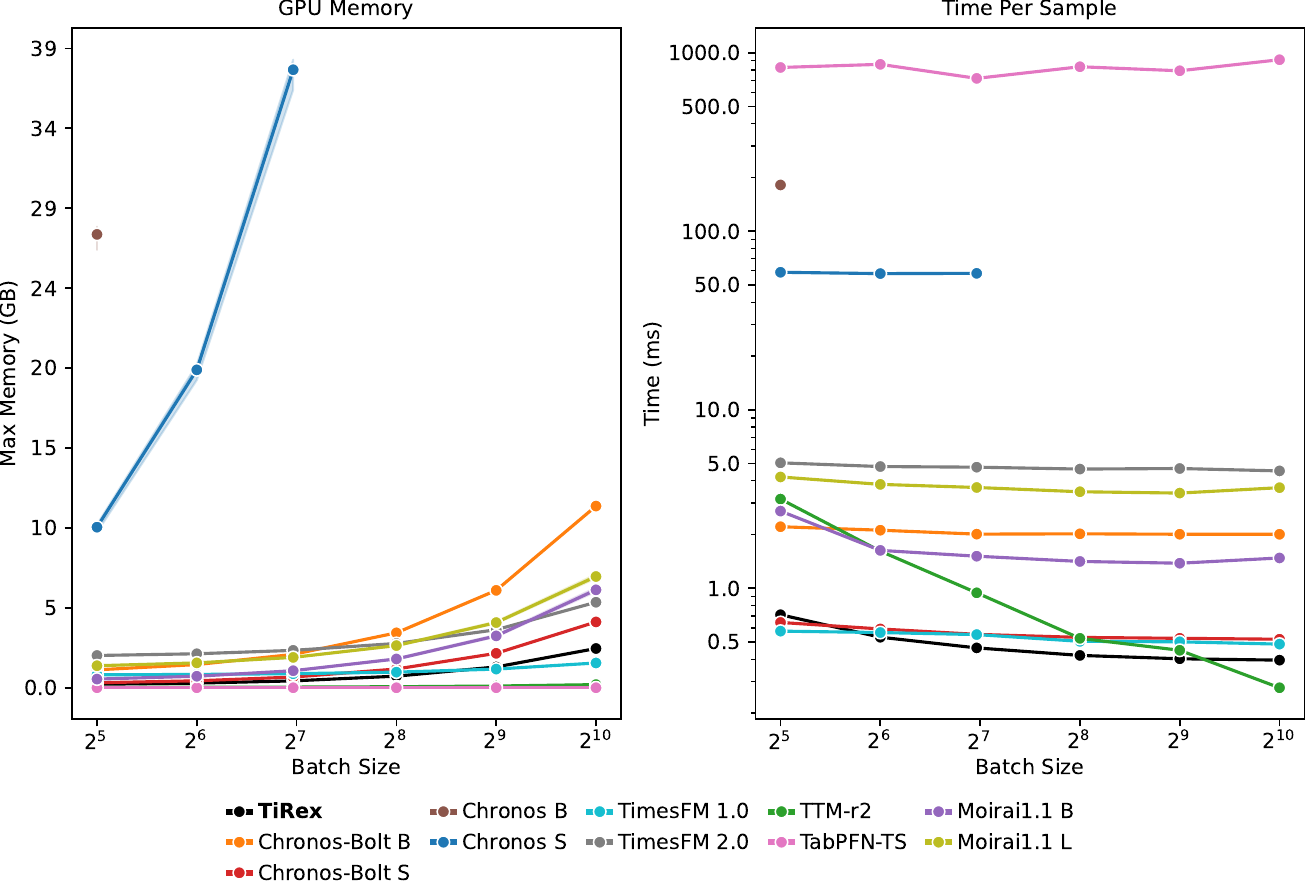} 
  \caption{Inference efficiency of different pre-trained forecasting models. Left: GPU Memory depending on the batch size. The maximum available GPU memory was 48 GB in the experiment (Nvidia A40). Right: Inference Time per sample depending on the batch size.}
  \label{fig:inference-efficiency-extended}
\end{figure}

\paragraph{Inference Efficiency}
Figure~\ref{fig:inference-efficiency-extended} presents an extended inference efficiency comparison, including additional pre-trained baselines beyond those shown in the main paper.

\subsection{Multivariate data}
While \modelname models each time series variate independently, this approach remains remarkably effective on multivariate forecasting tasks. 
This observation is consistent with previous work demonstrating that strong univariate models can serve as powerful baselines on multivariate benchmarks \citep[e.g.,][]{nieTimeSeriesWorth2022}.
Our main results on the full \giftevalzs, which contains 8 multivariate datasets, already support this, as \modelname outperforms several multivariate models.

\begin{figure}[ht]
  \centering
  \includegraphics[width=\textwidth]{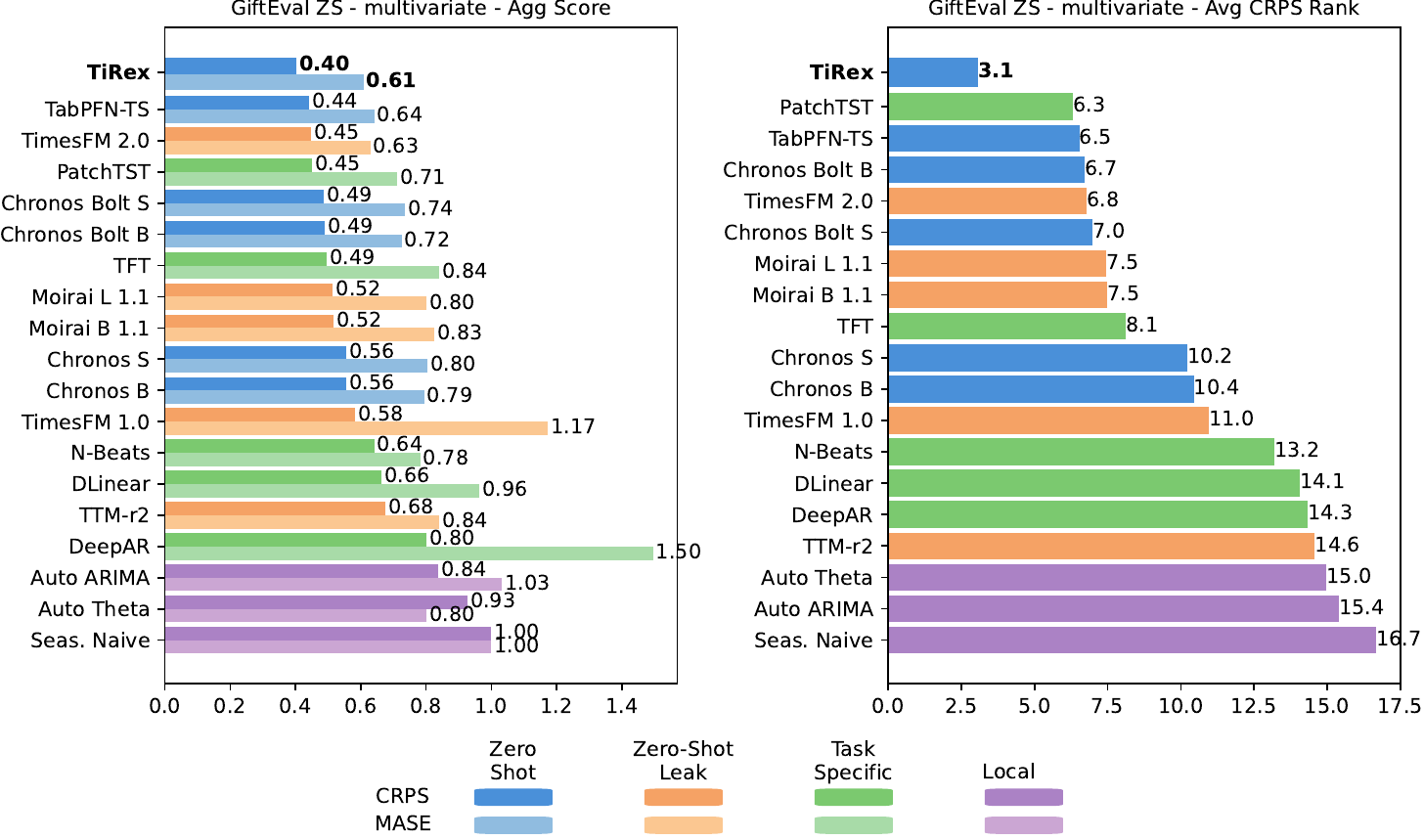} 
  \caption{Results for the \textbf{multivariate data} subset of the \giftevalzs: The aggregated scores and the average CRPS rank of the benchmarks' \textbf{short-term} sub-results are shown.
  Lower values are better. ``Zero-shot Leak'' refers to models that are partly trained on the benchmark datasets. We trained \modelname with 6 different seeds and report the observed standard deviation of the aggregated scores.}
  \label{fig:gifteval-multivariate}
\end{figure}

To make this point more explicit, Table~\ref{fig:gifteval-multivariate} isolates the performance on only the multivariate subset of the \giftevalzs benchmark.
Even in this setting, \modelname achieves the top rank among models designed specifically for multivariate forecasting (e.g., Moirai, TTM).

\vspace{3cm}

\clearpage

\subsection{Ablations}\label{app:ext-ablation-results}

\begin{figure}[ht]
  \centering
  \includegraphics[width=\textwidth]{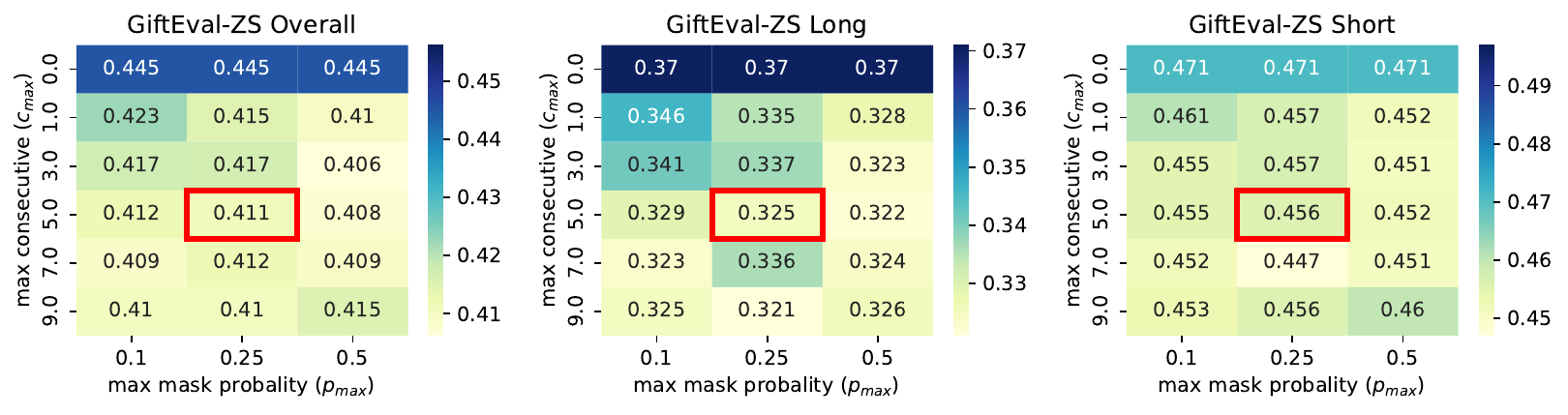} 
  \caption{CRPS results on the \giftevalzs of \modelname variants trained with a different set of hyperparameters for \fullpatchmask ($\fullpatchmaxcons=0$ indicates that \fullpatchmask is not used). The parameters used for training \modelname are enclosed in a red frame.}
  \label{fig:sensitivity-fullpatchmask}
\end{figure}

\paragraph{\fullpatchmask (\fullpatchmaskabrr)}
\fullpatchmaskabrr applies 2 hyperparameters: $\fullpatchmaxprob$, which defines the maximum masking probability sampled per sample, and $\fullpatchmaxcons,$ which defines the maximum for the number of consecutive patches sampled per sample.
As pre-training computational demand hinders an extended hyperparameter search, we heuristically selected $\fullpatchmaxprob=0.25$, which corresponds to a typical dropout probability of time series models, and $\fullpatchmaxcons=5$ to ensure multi-patch forecasts spanning multiple tokens.
To analyze the sensitivity of the model performance to these parameters, we additionally trained \modelname variants spanning the combinations of $\fullpatchmaxprob=\{0.1, 0.25, 0.5\}$ and $\fullpatchmaxcons=\{0, 1, 3, 5, 7, 9\}$.
Figure~\ref{fig:sensitivity-fullpatchmask} illustrates the results:
The results are not very sensitive to the parameters as long as \fullpatchmaskabrr is utilized ($\fullpatchmaxcons > 0$).
Additionally, good long-term forecasts require sufficient training samples with multi-patch forecasts spanning multiple tokens, i.e.,$\fullpatchmaxcons > 3$ or $\fullpatchmaxprob \geq 0.5$ (the latter more likely leads to neighboring masked patches that effectively mask out more than $c_{\text{mask}}$ patches).

\paragraph{Augmentations}
We do not apply each augmentation to every sample.
Due to the computational cost of pre-training, an extensive hyperparameter search was not feasible.
Instead, we heuristically selected application probabilities under the hypothesis that augmentations are beneficial, but excessive use may lead to diminished returns.
We chose an application probability of $0.5$ for both Censor and Amplitude Modulation.
For Spike Injection, we used a lower probability of $0.05$, as its computational cost is higher, and frequent application creates a speed bottleneck in training.

\begin{figure}[ht]
  \centering
  \includegraphics[width=\textwidth]{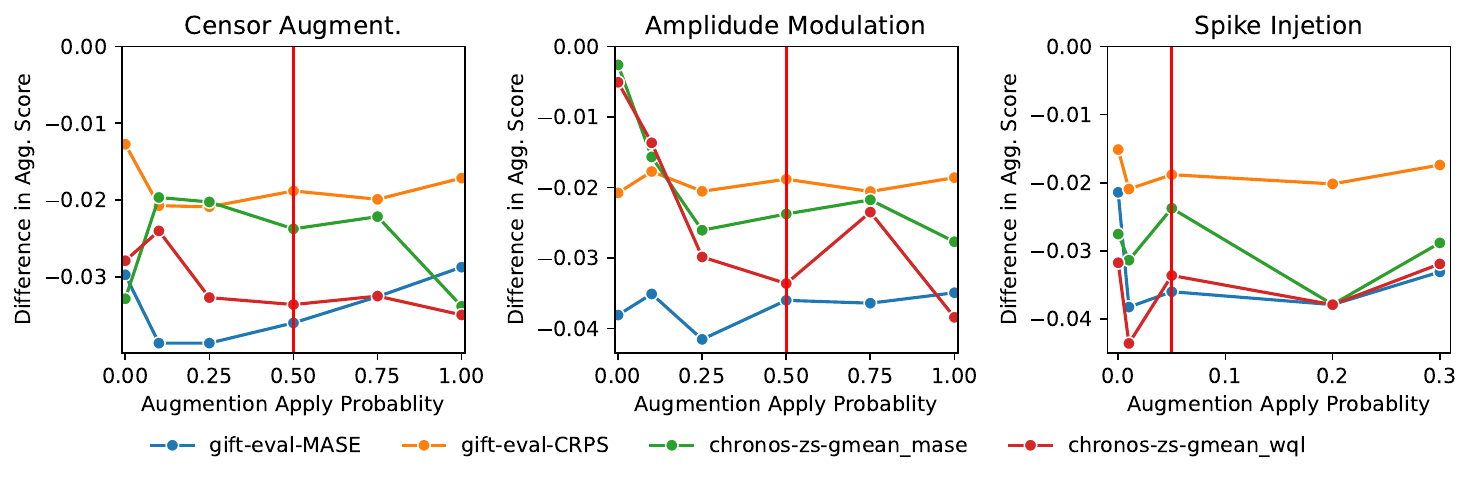} 
  \caption{CRPS performance variation on the \giftevalzs for \modelname variants trained with different augmentation application probabilities, relative to a baseline \modelname without augmentations.
  Red vertical lines indicate the parameters used in the actual model configuration.}
  \label{fig:sensitivity-applyprob}
\end{figure}

To analyze the sensitivity of \modelname's performance to these application probabilities, we trained additional variants with altered values.
Specifically, we tested probabilities in $\{0, 0.1, 0.25, 0.75, 1\}$ for Censor and Amplitude Modulation, and probabilities in $\{0, 0.01, 0.2, 0.3\}$ for Spike Injection. Figure~\ref{fig:sensitivity-applyprob} summarizes the results.
The analysis indicates that model performance is relatively robust to variations in application probability as long as the augmentations are utilized at all.
However, targeted tuning may yield marginal improvements.

\begin{figure}[th]
  \centering
  \includegraphics[width=\textwidth]{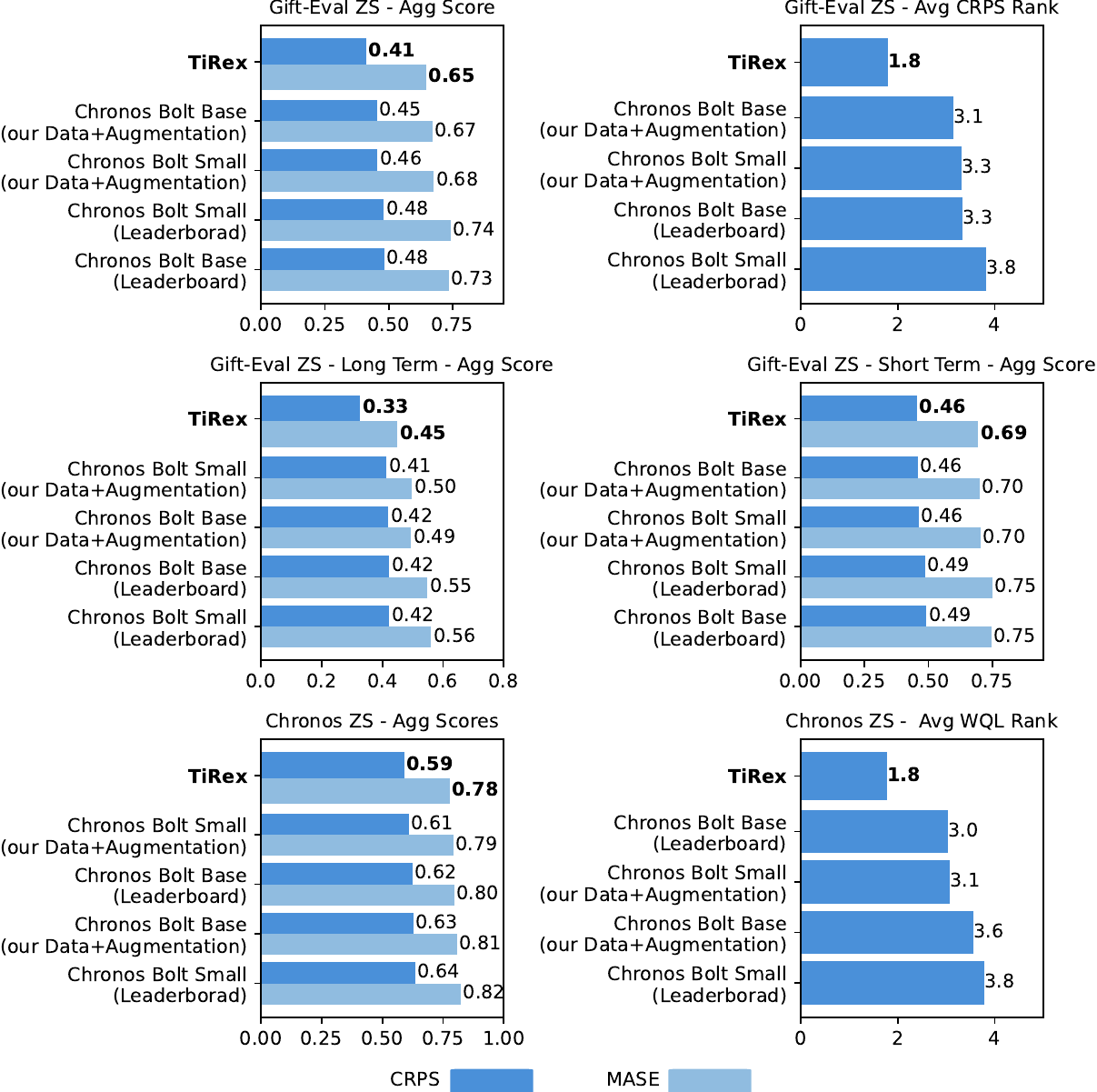} 
  \caption{Results of training a Chronos-Bolt architecture with the same data and data augmentations as \modelname\  --- compared to the published Chronos-Bolt model and \modelname. The results are from the \giftevalzs and the \chronoseval.}
  \label{fig:chronos-architecture-ablation}
\end{figure}

\paragraph{Chronos-Bolt architecture trained with our data pipeline}
In order to isolate the impact of our dataset and augmentation pipeline from other methodological contributions, we trained a Chronos-Bolt architecture, as representative of a state-of-the-art pre-trained model.
Chronos-Bolt was selected due to its publicly available code and configuration, though the exact training data and procedure remain undisclosed.
We used the same training hyperparameters (e.g., learning rate, warmup schedule, $\dots$) as for \modelname.
Figure~\ref{fig:chronos-architecture-ablation} shows that integrating our data pipeline leads to improved performance compared to the published Chronos-Bolt results on the \giftevalzs, while we see a mixed effect on the \chronoseval.
Nonetheless, \modelname consistently outperforms the optimized Chronos-Bolt variant, which highlights the contributions of both \fullpatchmaskabrr and the xLSTM backbone to \modelname{’s} effectiveness.
This difference is most pronounced for long-term forecasts.

\clearpage

\subsection{Additional qualitative examples}\label{app:ext-quali-results}
\begin{figure}[ht]
  \centering
  \includegraphics[width=\textwidth]{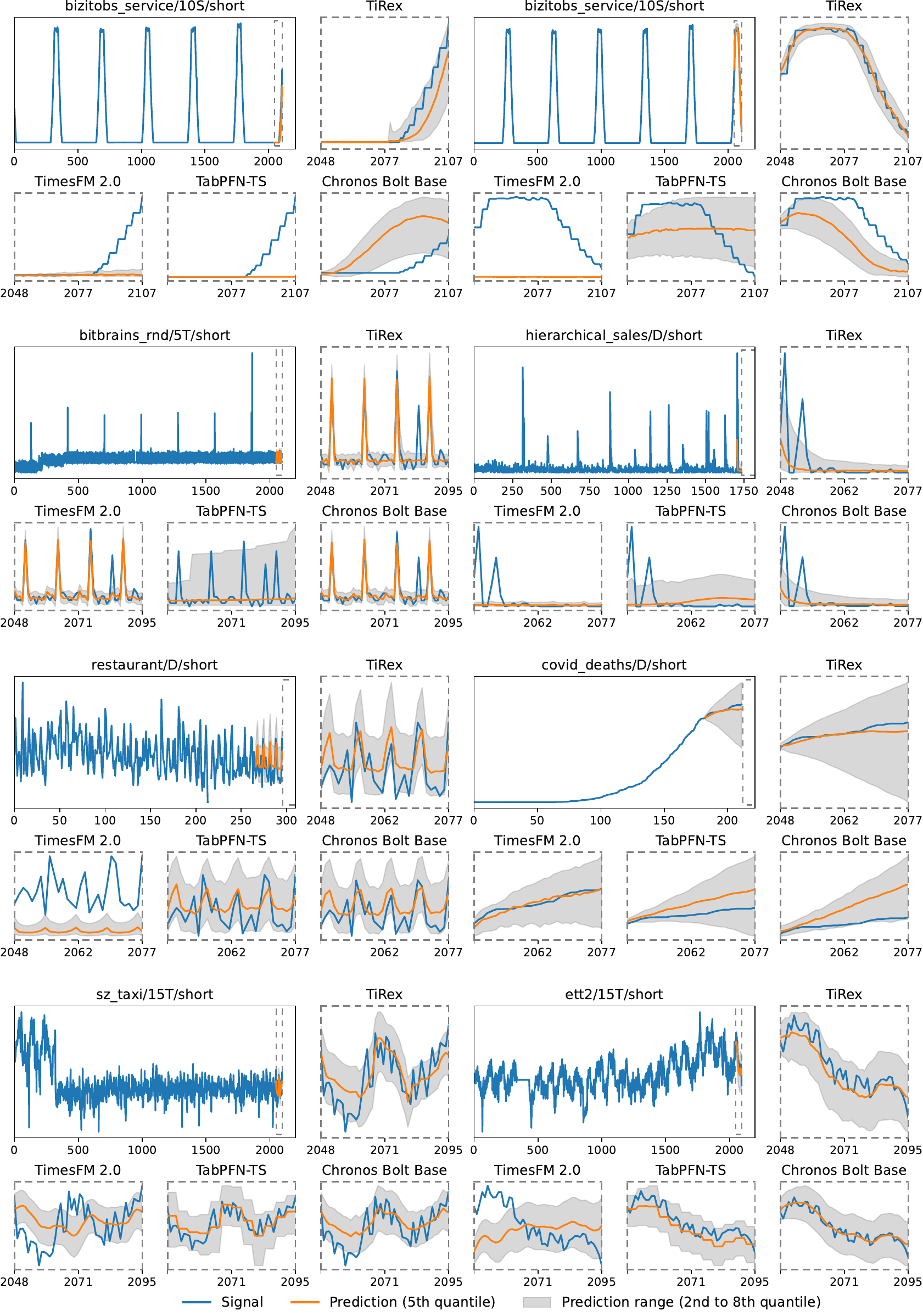} 
  \caption{Examples of \textbf{short-term forecasts} from the GiftEval benchmark. For each example, we show one plot with the full context and the \modelname prediction, as well as zoomed-in forecasts of the best-performing zero-shot models.}
  \label{fig:app-qualiative-short}
\end{figure}

In addition to Figure~\ref{fig:main-qualiative-comparison} from the main paper, we provide further qualitative examples of forecasts on the GiftEval benchmark.
Specifically, Figure~\ref{fig:app-qualiative-long} depicts the model behaviors for medium- and long-term forecasts and Figure~\ref{fig:app-qualiative-short} does the same for short-term forecasts.

\subsection{Finetuned-Forecasting}\label{app:ext-finetune-results}

To further explore the capabilities of our already strong pre-trained model, we 
finetune \modelname with the training split of the GiftEval benchmark \citep[as defined by][]{aksuGIFTEvalBenchmarkGeneral2024a}.
Fine-tuning is performed jointly across all training datasets.
To avoid overfitting, we mix the training data with our pre-training data, using a $20/80$ ratio. 
For sampling, we use a uniform distribution to choose the dataset to draw the next training sample from.
We freeze the input and output layers of \modelname and run over $40\mathrm{k}$ steps with an initial learning rate of $1\times 10^{-3}$ and a linear learning rate decay that reaches $0$ at the end of the run.
In contrast to the pre-training regime, we do not apply any data augmentation techniques (see Section~\ref{sec:datapipeline}) but still employ \fullpatchmaskabrr (Section~\ref{sec:fullpatchmask}).

\begin{figure}[ht]
    \centering
  \includegraphics[width=\textwidth]{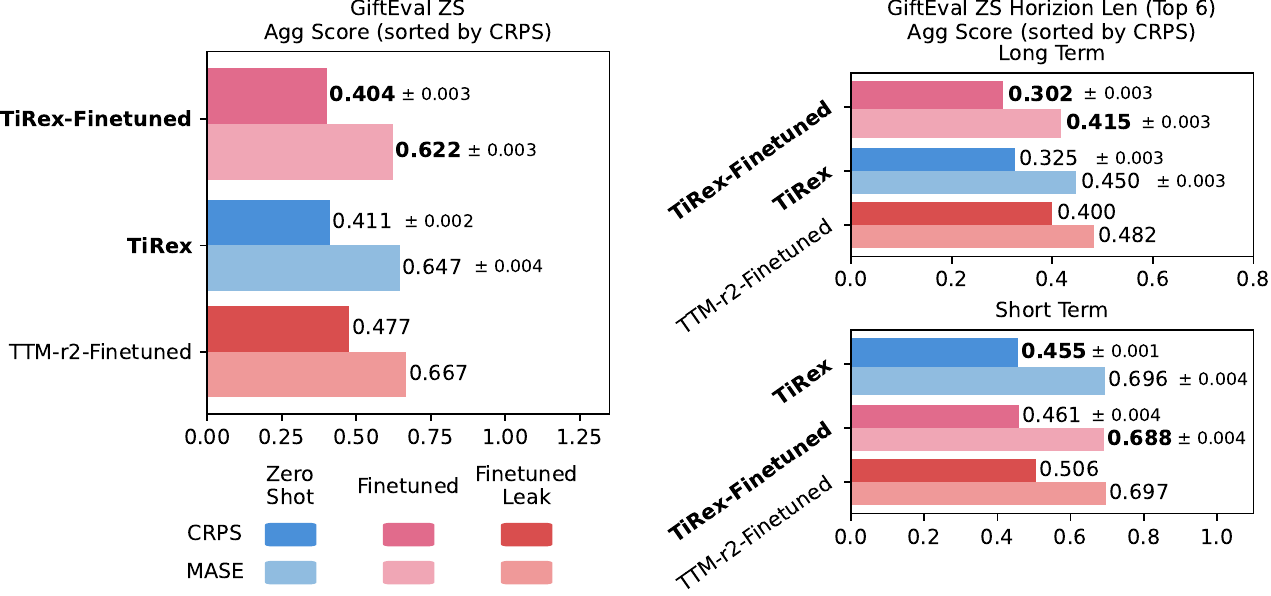} 
    \caption{\giftevalzs evaluation results comparing the finetuned models from the GiftEval leaderboard to our pre-trained and finetuned models. We trained \modelname with 6 different seeds, finetuned each of these models with 4 different seeds (24 seeds in total), and report the mean and the observed standard deviation of the aggregated scores.}
    \label{fig:finetunedresultsqualitative}
\end{figure}

In the finetuning setting, we observe an incremental improvement over the pre-trained model, especially in the MASE metric (Figure~\ref{fig:finetunedresultsqualitative}).
Specifically, we compare the pre-trained to its finetuned version as well as to a fine-tuned TTM~\citep{ekambaramTinyTimeMixers2024b}, the only zero-shot model that provides fine-tune results on the GiftEval leaderboard.

\section{TiRex 1.1 - Full GiftEval Zero-Shot}\label{app:tirex_retrained}
Subsequent to the initial release, we developed an updated model variant, denoted as TiRex 1.1, to ensure a \textbf{completely zero-shot evaluation on the full GiftEval benchmark}.
For this version, we revised the pre-training data corpus with the following modifications to eliminate any potential data leakage:
(1) All datasets present in the GiftEval benchmark were removed from our training corpus, across all their respective sampling frequencies.
(2) Datasets from the Chronos-ZS benchmark that do not overlap with GiftEval were included to enhance data diversity.
(3) To address a potential but difficult-to-verify overlap in the 'solar' dataset --- where time series, specifically from Alabama, might exist in Chronos training data and GiftEval despite differing frequencies ---we proactively removed that specific subset, thereby removing any ambiguity in its zero-shot status.

Beyond the training data adjustments, we also incorporated \textit{long-period normalization}, a pre-processing enhancement aimed at better handling long-range periodicities. This technique addresses the challenge of fitting long period patterns into \modelname's context window by identifying the dominant frequency of the time series and resampling it such that one period fits into context.

\begin{figure}
  \centering
  \includegraphics[width=\textwidth]{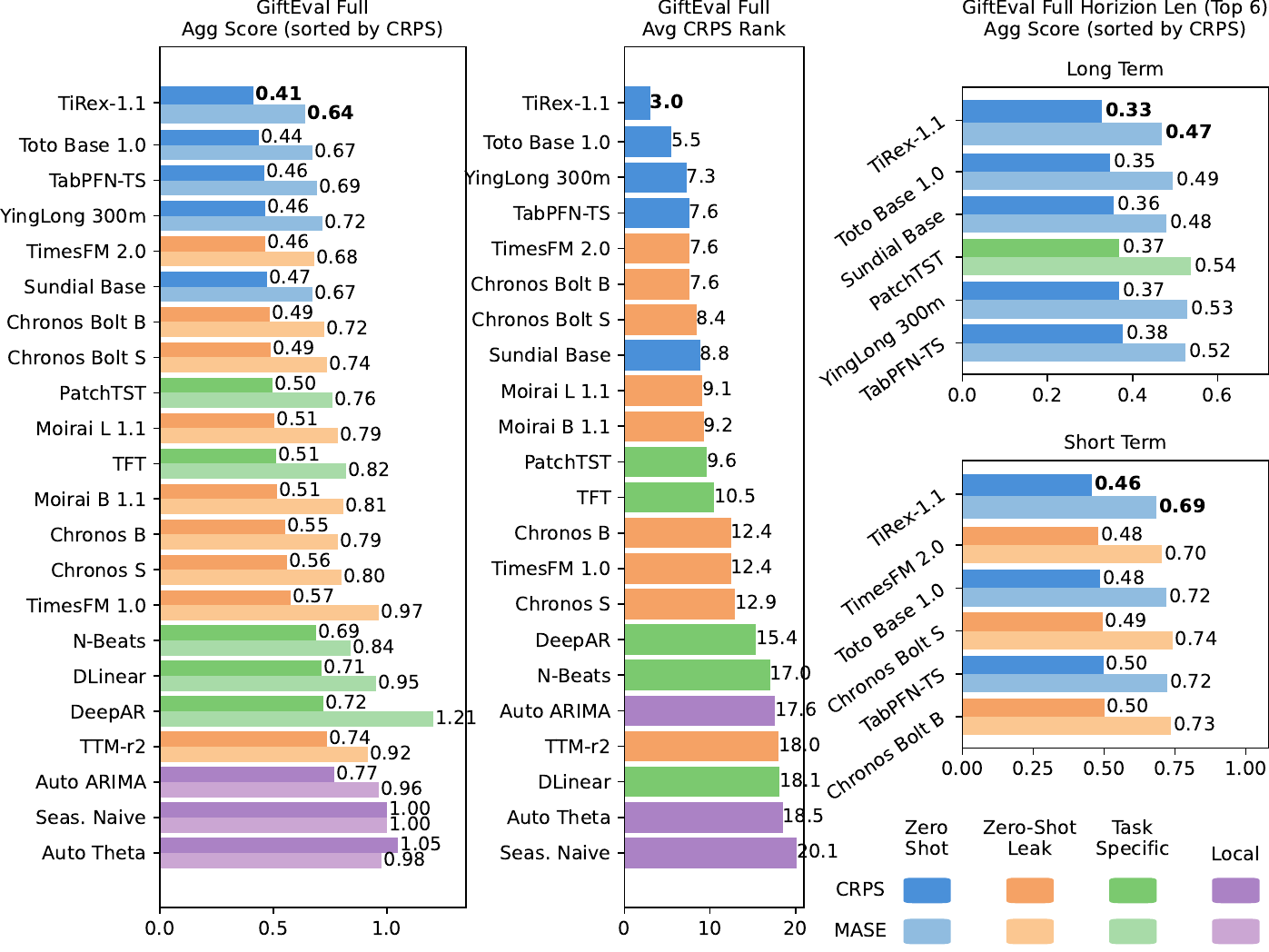} 
  \caption{Results of the \textbf{full GiftEval benchmark} with \textbf{TiRex 1.1}: Aggregated scores of the overall benchmark and the short- and long-term performances.
  Additionally, the average rank in terms of CRPS, as in the public leaderboard, is presented. 
  Lower values are better. ``Zero-shot Leak'' refers to models that are partly trained on the benchmark datasets.}
  \label{fig:gifteval-full-retrained}
\end{figure}

Figure~\ref{fig:gifteval-full-retrained} shows the performance of TiRex 1.1 on the full GiftEval benchmark\footnote{GiftEval benchmark scores are reported based on the leaderboard computation at the time of our submission. A subsequent update to the seasonal naive baseline affects the absolute aggregated scores but does not change any discussed model ranking, result, or conclusion.}, alongside new results from concurrent works published after our initial submission, including ToTo~\citep{cohen2025timedifferentobservabilityperspective}, Sundial~\citep{liu2025sundial}, and Yinglong~\citep{wang2025output}.
In this updated and strictly zero-shot comparison, TiRex 1.1 maintains its state-of-the-art performance, achieving the top rank across all reported metrics.

\section{Societal Impact}\label{app:societal-impact}
As a pre-trained zero-shot model, \modelname could democratize access to modern forecasting techniques by removing the need for task-specific training or machine learning expertise, enabling broader adoption across non-expert communities.
It also has the potential to enhance forecasting in data-sparse domains.
Nonetheless, care must be taken to ensure responsible deployment, particularly in high-stakes settings where model errors could have significant real-world consequences.

\section{Code}
The code repository for the model is hosted on GitHub: \url{https://github.com/NX-AI/tirex}

\begin{figure}[ht]
  \centering
  \includegraphics[width=\textwidth]{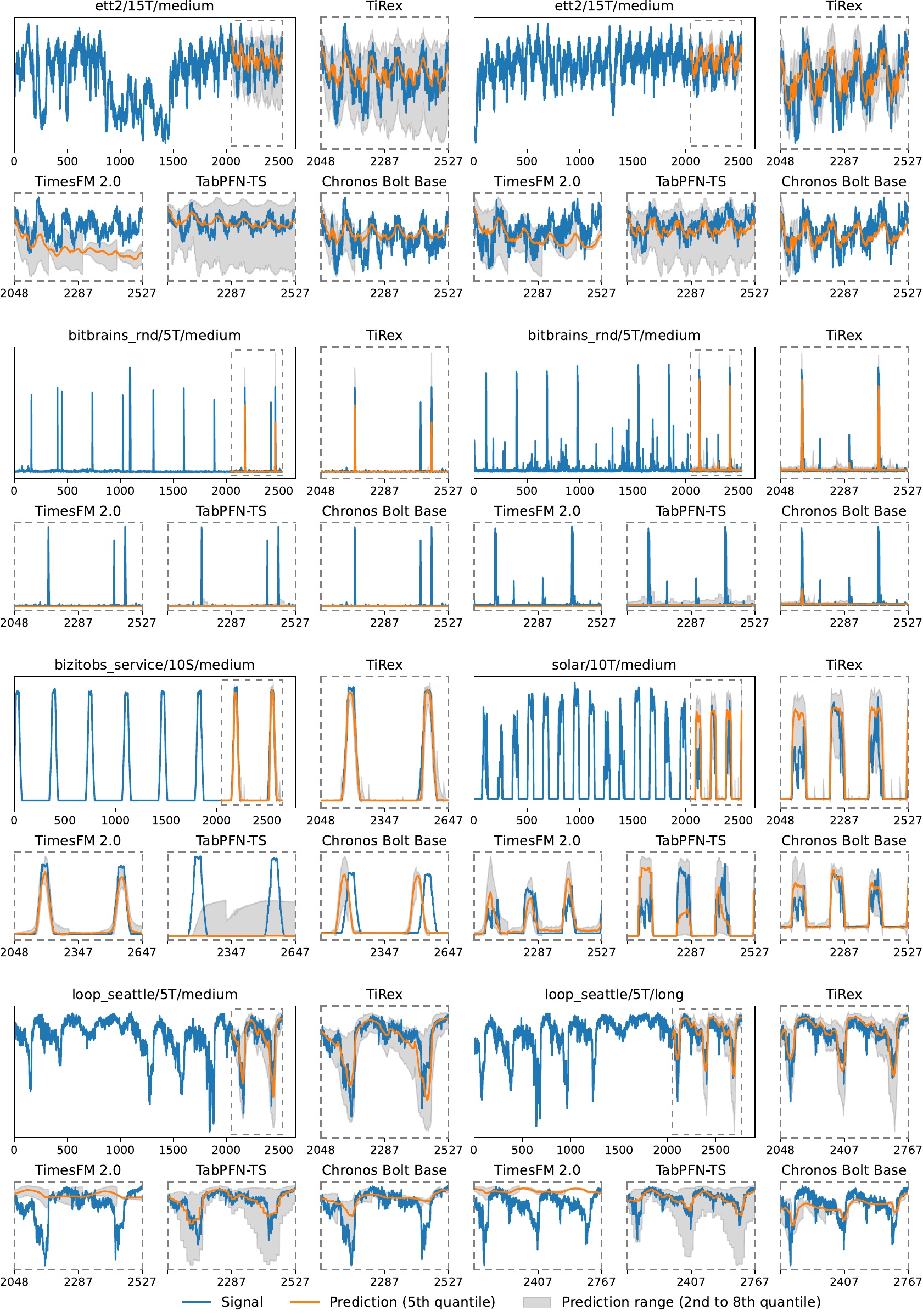} 
  \caption{Examples of \textbf{medium- and long-term forecasts} from the GiftEval benchmark. For each example, we show one plot with the full context and the \modelname prediction, as well as zoomed-in forecasts of the best-performing zero-shot models.}
  \label{fig:app-qualiative-long}
\end{figure}

\clearpage

\setNextCaption{MASE scores of different zero-shot models on the \underline{GiftEval} benchmark evaluation settings (Part 1/2). The models achieving the \textbf{best} and \underline{second-best} scores are highlighted. Results for datasets that are part of the training data for the respective models are shaded in grey, and these results are excluded from the calculation of the best score. We trained \modelname with 6 different seeds and report the observed standard deviation in the plot.}

\begin{table}[h]
\centering
\caption{\nextcaption}
{\scriptsize
\begin{tabular}{llllllllllll}
\toprule
  & \rotatebox{90}{  \modelname } & \rotatebox{90}{ Chronos Bolt B } & \rotatebox{90}{ Chronos Bolt S } & \rotatebox{90}{ TimesFM 2.0 } & \rotatebox{90}{ TimesFM 1.0 } & \rotatebox{90}{ TabPFN-TS } & \rotatebox{90}{ Moirai L 1.1 } & \rotatebox{90}{ Moirai B 1.1 } & \rotatebox{90}{ TTM-r2 } & \rotatebox{90}{ Chronos B } & \rotatebox{90}{ Chronos S }\\
\midrule
bitbrains\_fast\_storage/5T/long & \textbf{0.916} $\pm$ 0.006 & \underline{0.948} & 0.962 & 0.980 & 32.0 & 1.04 & 0.955 & 0.967 & 1.16 & 1.01 & 1.08 \\
bitbrains\_fast\_storage/5T/medium & \textbf{1.00} $\pm$ 0.011 & 1.06 & 1.06 & 1.08 & 20.3 & 1.19 & \underline{1.02} & 1.05 & 1.23 & 1.11 & 1.14 \\
bitbrains\_fast\_storage/5T/short & \textbf{0.692} $\pm$ 0.007 & 0.752 & 0.778 & \underline{0.731} & 0.874 & 0.888 & 0.827 & 0.792 & 0.966 & 0.850 & 0.833 \\
bitbrains\_fast\_storage/H/short & \textbf{1.06} $\pm$ 0.013 & \underline{1.07} & 1.08 & 1.09 & 1.18 & 1.15 & 1.09 & 1.18 & 1.28 & 1.11 & 1.14 \\
bitbrains\_rnd/5T/long & \textbf{3.35} $\pm$ 0.013 & \underline{3.40} & 3.41 & 3.60 & 91.0 & 3.51 & 3.42 & 3.45 & 3.78 & 3.77 & 3.83 \\
bitbrains\_rnd/5T/medium & \textbf{4.40} $\pm$ 0.007 & \underline{4.45} & 4.47 & 4.59 & 49.8 & 4.59 & 4.46 & 4.53 & 4.81 & 4.60 & 4.63 \\
bitbrains\_rnd/5T/short & \textbf{1.66} $\pm$ 0.004 & \underline{1.71} & 1.72 & 1.77 & 1.98 & 1.87 & 1.75 & 1.82 & 2.05 & 1.79 & 1.81 \\
bitbrains\_rnd/H/short & \underline{5.84} $\pm$ 0.011 & 5.90 & 5.88 & 5.99 & 6.09 & 5.96 & 5.93 & 6.07 & 6.16 & \textbf{5.79} & 5.89 \\
bizitobs\_application/10S/long & \textbf{3.67} $\pm$ 0.069 & 10.5 & 9.65 & \underline{4.07} & 16.3 & 7.73 & 7.84 & 13.5 & 9.59 & 9.25 & 9.67 \\
bizitobs\_application/10S/medium & \textbf{2.85} $\pm$ 0.057 & 9.72 & 9.15 & \underline{3.08} & 11.2 & 6.93 & 7.39 & 12.8 & 9.02 & 9.87 & 10.2 \\
bizitobs\_application/10S/short & \textbf{1.40} $\pm$ 0.112 & 5.53 & 5.41 & \underline{1.56} & 4.36 & 3.19 & 4.51 & 5.32 & 4.21 & 3.01 & 3.34 \\
bizitobs\_l2c/5T/long & \underline{1.19} $\pm$ 0.040 & 1.24 & 1.36 & 1.24 & 1.28 & 1.22 & \textbf{1.12} & \textbf{1.12} & 1.32 & 1.20 & 1.25 \\
bizitobs\_l2c/5T/medium & \textbf{0.849} $\pm$ 0.028 & 0.878 & 0.920 & 1.02 & 0.887 & 0.870 & 0.987 & \underline{0.853} & 0.992 & 0.942 & 0.880 \\
bizitobs\_l2c/5T/short & 0.290 $\pm$ 0.005 & \underline{0.278} & \textbf{0.272} & 0.312 & 0.310 & 0.338 & 0.285 & 0.291 & 0.324 & 0.301 & 0.301 \\
bizitobs\_l2c/H/long & \underline{0.590} $\pm$ 0.022 & \textbf{0.556} & 0.612 & 1.30 & 1.27 & 0.919 & 1.27 & 1.09 & 1.24 & 1.25 & 1.31 \\
bizitobs\_l2c/H/medium & \underline{0.525} $\pm$ 0.020 & \textbf{0.495} & 0.570 & 1.18 & 1.49 & 0.748 & 1.25 & 1.32 & 1.24 & 1.34 & 1.32 \\
bizitobs\_l2c/H/short & 0.528 $\pm$ 0.021 & \textbf{0.432} & \underline{0.485} & 0.782 & 1.05 & 0.634 & 1.15 & 0.999 & 0.983 & 0.990 & 0.905 \\
bizitobs\_service/10S/long & \textbf{1.57} $\pm$ 0.057 & 5.30 & 4.85 & \underline{2.15} & 7.77 & 3.90 & 4.33 & 6.08 & 5.34 & 4.22 & 3.89 \\
bizitobs\_service/10S/medium & \textbf{1.32} $\pm$ 0.058 & 4.98 & 4.64 & \underline{1.53} & 6.46 & 3.78 & 3.87 & 5.99 & 5.12 & 4.58 & 4.06 \\
bizitobs\_service/10S/short & \textbf{0.884} $\pm$ 0.052 & 3.32 & 2.90 & \underline{1.04} & 2.90 & 2.00 & 2.31 & 3.43 & 2.73 & 1.88 & 1.91 \\
car\_parts/M/short & \textbf{0.838} $\pm$ 0.004 & 0.855 & 0.858 & 0.922 & 0.893 & \underline{0.843} & \cellcolor{gray!30}{0.903} & \cellcolor{gray!30}{0.835} & 1.57 & 0.908 & 0.885 \\
covid\_deaths/D/short & 39.5 $\pm$ 0.803 & 38.9 & \textbf{36.5} & 47.4 & 55.6 & \underline{37.8} & \cellcolor{gray!30}{36.5} & \cellcolor{gray!30}{34.6} & 53.5 & 42.7 & 42.2 \\
electricity/15T/long & \cellcolor{gray!30}{0.891} $\pm$ 0.008 & \cellcolor{gray!30}{0.933} & \cellcolor{gray!30}{0.953} & \textbf{0.904} & 1.50 & \underline{1.02} & 1.31 & 1.32 & 1.35 & \cellcolor{gray!30}{1.01} & \cellcolor{gray!30}{1.06} \\
electricity/15T/medium & \cellcolor{gray!30}{0.841} $\pm$ 0.006 & \cellcolor{gray!30}{0.862} & \cellcolor{gray!30}{0.896} & \textbf{0.845} & 1.49 & \underline{0.977} & 1.29 & 1.33 & 1.32 & \cellcolor{gray!30}{0.990} & \cellcolor{gray!30}{1.03} \\
electricity/15T/short & \cellcolor{gray!30}{0.945} $\pm$ 0.008 & \cellcolor{gray!30}{0.935} & \cellcolor{gray!30}{0.936} & \textbf{0.907} & 1.48 & \underline{1.23} & 1.71 & 1.54 & 1.43 & \cellcolor{gray!30}{1.05} & \cellcolor{gray!30}{1.13} \\
electricity/D/short & \textbf{1.43} $\pm$ 0.010 & \underline{1.45} & 1.48 & 1.49 & 1.75 & 1.49 & 1.51 & 1.50 & 1.66 & 1.56 & 1.60 \\
electricity/H/long & \cellcolor{gray!30}{1.21} $\pm$ 0.018 & \cellcolor{gray!30}{1.24} & \cellcolor{gray!30}{1.26} & \textbf{1.05} & \underline{1.21} & 1.34 & 1.36 & 1.26 & 1.38 & \cellcolor{gray!30}{1.20} & \cellcolor{gray!30}{1.23} \\
electricity/H/medium & \cellcolor{gray!30}{1.08} $\pm$ 0.010 & \cellcolor{gray!30}{1.08} & \cellcolor{gray!30}{1.10} & \textbf{0.929} & \underline{1.07} & 1.18 & 1.20 & 1.19 & 1.25 & \cellcolor{gray!30}{1.06} & \cellcolor{gray!30}{1.09} \\
electricity/H/short & \cellcolor{gray!30}{0.869} $\pm$ 0.009 & \cellcolor{gray!30}{0.873} & \cellcolor{gray!30}{0.914} & \textbf{0.763} & \underline{0.878} & 1.04 & 1.08 & 1.09 & 1.16 & \cellcolor{gray!30}{0.902} & \cellcolor{gray!30}{0.951} \\
electricity/W/short & \cellcolor{gray!30}{1.46} $\pm$ 0.010 & \cellcolor{gray!30}{1.48} & \cellcolor{gray!30}{1.50} & \textbf{1.45} & 1.86 & \textbf{1.45} & \underline{1.79} & 1.92 & 2.52 & \cellcolor{gray!30}{1.49} & \cellcolor{gray!30}{1.54} \\
ett1/15T/long & \textbf{1.05} $\pm$ 0.009 & 1.14 & 1.19 & \underline{1.11} & 1.32 & \underline{1.11} & 1.40 & 1.12 & 1.20 & 1.35 & 1.50 \\
ett1/15T/medium & \textbf{1.04} $\pm$ 0.009 & 1.06 & 1.11 & 1.08 & 1.27 & \underline{1.05} & 1.30 & 1.24 & 1.14 & 1.32 & 1.36 \\
ett1/15T/short & 0.706 $\pm$ 0.007 & \textbf{0.680} & \underline{0.704} & 0.719 & 0.875 & 0.787 & 0.925 & 0.825 & 0.812 & 0.801 & 0.872 \\
ett1/D/short & 1.71 $\pm$ 0.016 & \underline{1.67} & 1.70 & \textbf{1.65} & 1.70 & 1.77 & 1.75 & 1.74 & 1.96 & 1.90 & 1.80 \\
ett1/H/long & \textbf{1.34} $\pm$ 0.030 & \underline{1.35} & 1.44 & 1.51 & 1.41 & 1.46 & 1.45 & 1.38 & 1.36 & 1.43 & 1.42 \\
ett1/H/medium & \textbf{1.25} $\pm$ 0.017 & 1.37 & 1.37 & \underline{1.31} & 1.44 & 1.36 & 1.34 & 1.35 & 1.32 & 1.37 & \underline{1.31} \\
ett1/H/short & \textbf{0.827} $\pm$ 0.007 & \underline{0.828} & 0.834 & 0.866 & 0.938 & 0.891 & 0.855 & 0.885 & 0.882 & 0.840 & 0.898 \\
ett1/W/short & 1.72 $\pm$ 0.044 & 1.70 & 1.70 & 1.65 & 1.73 & 1.58 & \textbf{1.51} & \underline{1.54} & \underline{1.54} & 1.66 & 1.65 \\
ett2/15T/long & \textbf{0.932} $\pm$ 0.012 & \underline{0.940} & 0.991 & 0.941 & 1.03 & 0.958 & 1.14 & 1.30 & 0.986 & 1.14 & 1.11 \\
ett2/15T/medium & \textbf{0.910} $\pm$ 0.010 & \underline{0.922} & 0.987 & 0.938 & 1.01 & 0.939 & 1.06 & 1.10 & 0.987 & 1.06 & 1.02 \\
ett2/15T/short & \underline{0.749} $\pm$ 0.010 & 0.766 & 0.788 & \textbf{0.747} & 0.898 & 0.845 & 1.00 & 0.959 & 0.832 & 0.857 & 0.885 \\
ett2/D/short & 1.28 $\pm$ 0.019 & 1.32 & \textbf{1.22} & 1.56 & 1.64 & 1.54 & 1.44 & 1.31 & 1.56 & \underline{1.26} & 1.43 \\
ett2/H/long & 1.16 $\pm$ 0.037 & \textbf{1.04} & \underline{1.07} & 1.13 & 1.09 & 1.37 & 1.28 & 1.12 & 1.13 & 1.12 & \textbf{1.04} \\
ett2/H/medium & \underline{1.05} $\pm$ 0.021 & \textbf{1.03} & \underline{1.05} & \underline{1.05} & 1.12 & 1.28 & 1.18 & \textbf{1.03} & 1.10 & 1.15 & 1.14 \\
ett2/H/short & \underline{0.742} $\pm$ 0.006 & \textbf{0.733} & 0.744 & 0.755 & 0.821 & 0.787 & 0.783 & 0.807 & 0.790 & 0.781 & 0.790 \\
ett2/W/short & 0.797 $\pm$ 0.040 & \textbf{0.739} & 0.791 & 1.12 & 1.13 & 0.959 & 1.31 & 0.851 & 1.36 & \underline{0.749} & 0.807 \\
hierarchical\_sales/D/short & \underline{0.744} $\pm$ 0.002 & \textbf{0.743} & 0.749 & 0.752 & 0.745 & 0.766 & \cellcolor{gray!30}{0.745} & \cellcolor{gray!30}{0.746} & 0.834 & 0.774 & 0.801 \\
hierarchical\_sales/W/short & \underline{0.721} $\pm$ 0.001 & 0.733 & 0.733 & \textbf{0.703} & 0.725 & 0.723 & 0.749 & 0.747 & 1.09 & 0.764 & 0.756 \\
hospital/M/short & 0.767 $\pm$ 0.003 & 0.791 & 0.801 & \underline{0.755} & 0.783 & \textbf{0.753} & \cellcolor{gray!30}{0.768} & \cellcolor{gray!30}{0.775} & 1.05 & 0.816 & 0.813 \\
\bottomrule
\end{tabular}

}
\label{tab:gifteval-zs-zs-model-MASE-p1}
\end{table}

\setNextCaption{MASE scores of different zero-shot models on the \underline{GiftEval} benchmark evaluation settings (Part 2/2). The models achieving the \textbf{best} and \underline{second-best} scores are highlighted. Results for datasets that are part of the training data for the respective models are shaded in grey, and these results are excluded from the calculation of the best score. We trained \modelname with 6 different seeds and report the observed standard deviation in the plot.}

\begin{table}[h]
\centering
\caption{\nextcaption}
{\scriptsize
\begin{tabular}{llllllllllll}
\toprule
  & \rotatebox{90}{  \modelname } & \rotatebox{90}{ Chronos Bolt B } & \rotatebox{90}{ Chronos Bolt S } & \rotatebox{90}{ TimesFM 2.0 } & \rotatebox{90}{ TimesFM 1.0 } & \rotatebox{90}{ TabPFN-TS } & \rotatebox{90}{ Moirai L 1.1 } & \rotatebox{90}{ Moirai B 1.1 } & \rotatebox{90}{ TTM-r2 } & \rotatebox{90}{ Chronos B } & \rotatebox{90}{ Chronos S }\\
\midrule
jena\_weather/10T/long & \textbf{0.641} $\pm$ 0.015 & 0.657 & 0.703 & \cellcolor{gray!30}{0.231} & \cellcolor{gray!30}{1.36} & 0.657 & 0.792 & 0.762 & \underline{0.649} & 0.802 & 0.956 \\
jena\_weather/10T/medium & \textbf{0.610} $\pm$ 0.005 & \textbf{0.610} & 0.646 & \cellcolor{gray!30}{0.191} & \cellcolor{gray!30}{1.10} & \underline{0.625} & 0.694 & 0.712 & 0.656 & 0.722 & 0.744 \\
jena\_weather/10T/short & \textbf{0.297} $\pm$ 0.006 & \underline{0.306} & 0.320 & \cellcolor{gray!30}{0.091} & \cellcolor{gray!30}{0.318} & 0.325 & 0.338 & 0.350 & 0.346 & 0.366 & 0.359 \\
jena\_weather/D/short & \textbf{1.02} $\pm$ 0.014 & 1.05 & \underline{1.03} & 1.24 & 1.20 & 1.23 & 1.14 & 1.15 & 2.53 & 1.12 & 1.14 \\
jena\_weather/H/long & \underline{0.987} $\pm$ 0.031 & 1.03 & 1.06 & 1.06 & 1.38 & 1.10 & \textbf{0.881} & 1.06 & 1.07 & 1.11 & 1.16 \\
jena\_weather/H/medium & 0.828 $\pm$ 0.023 & \underline{0.747} & \textbf{0.721} & 0.864 & 1.05 & 0.954 & 0.891 & 0.817 & 0.815 & 0.883 & 0.842 \\
jena\_weather/H/short & \textbf{0.516} $\pm$ 0.003 & 0.536 & 0.540 & \underline{0.525} & 0.589 & 0.595 & 0.585 & 0.554 & 0.575 & 0.567 & 0.583 \\
kdd\_cup\_2018/D/short & \cellcolor{gray!30}{1.21} $\pm$ 0.009 & \cellcolor{gray!30}{1.20} & \cellcolor{gray!30}{1.19} & \underline{1.21} & \underline{1.21} & \textbf{1.17} & \cellcolor{gray!30}{1.20} & \cellcolor{gray!30}{1.20} & \textbf{1.17} & \cellcolor{gray!30}{1.37} & \cellcolor{gray!30}{1.40} \\
kdd\_cup\_2018/H/long & \cellcolor{gray!30}{0.759} $\pm$ 0.027 & \cellcolor{gray!30}{0.684} & \cellcolor{gray!30}{0.925} & \underline{1.03} & 1.09 & 1.06 & \cellcolor{gray!30}{0.867} & \cellcolor{gray!30}{0.960} & \textbf{1.00} & \cellcolor{gray!30}{1.14} & \cellcolor{gray!30}{1.24} \\
kdd\_cup\_2018/H/medium & \cellcolor{gray!30}{0.825} $\pm$ 0.024 & \cellcolor{gray!30}{0.700} & \cellcolor{gray!30}{0.857} & \textbf{1.03} & 1.14 & 1.14 & \cellcolor{gray!30}{0.954} & \cellcolor{gray!30}{1.05} & \underline{1.07} & \cellcolor{gray!30}{1.24} & \cellcolor{gray!30}{1.34} \\
kdd\_cup\_2018/H/short & \cellcolor{gray!30}{0.657} $\pm$ 0.006 & \cellcolor{gray!30}{0.601} & \cellcolor{gray!30}{0.667} & \textbf{0.941} & \underline{1.09} & 1.10 & \cellcolor{gray!30}{0.894} & \cellcolor{gray!30}{0.944} & \cellcolor{gray!30}{1.01} & \cellcolor{gray!30}{1.04} & \cellcolor{gray!30}{1.06} \\
loop\_seattle/5T/long & \textbf{1.02} $\pm$ 0.038 & 1.24 & 1.19 & 1.13 & 1.43 & \underline{1.05} & \cellcolor{gray!30}{0.556} & \cellcolor{gray!30}{0.591} & \cellcolor{gray!30}{1.11} & 1.31 & 1.38 \\
loop\_seattle/5T/medium & \textbf{0.941} $\pm$ 0.019 & 1.14 & 1.15 & 1.12 & 1.49 & \underline{0.972} & \cellcolor{gray!30}{0.450} & \cellcolor{gray!30}{0.523} & \cellcolor{gray!30}{1.07} & 1.61 & 1.61 \\
loop\_seattle/5T/short & \textbf{0.572} $\pm$ 0.005 & 0.628 & 0.631 & \underline{0.583} & 0.836 & 0.588 & \cellcolor{gray!30}{0.486} & \cellcolor{gray!30}{0.536} & \cellcolor{gray!30}{0.631} & 0.764 & 0.768 \\
loop\_seattle/D/short & \underline{0.878} $\pm$ 0.005 & 0.903 & 0.919 & \textbf{0.859} & 0.880 & 0.899 & 0.916 & 0.903 & \cellcolor{gray!30}{1.74} & 0.912 & 1.00 \\
loop\_seattle/H/long & \underline{0.917} $\pm$ 0.011 & 0.996 & 1.08 & \textbf{0.906} & 1.26 & 0.922 & 1.05 & 1.15 & \cellcolor{gray!30}{1.11} & 1.01 & 1.12 \\
loop\_seattle/H/medium & \underline{0.944} $\pm$ 0.012 & 1.02 & 1.10 & \textbf{0.934} & 1.32 & 0.948 & 1.00 & 1.14 & \cellcolor{gray!30}{1.19} & 1.05 & 1.14 \\
loop\_seattle/H/short & \underline{0.850} $\pm$ 0.005 & 0.900 & 0.915 & \textbf{0.832} & 1.15 & 0.912 & 0.945 & 1.06 & \cellcolor{gray!30}{1.08} & 0.926 & 0.967 \\
m4\_daily/D/short & \cellcolor{gray!30}{3.15} $\pm$ 0.063 & \cellcolor{gray!30}{3.20} & \cellcolor{gray!30}{3.19} & \cellcolor{gray!30}{3.09} & \cellcolor{gray!30}{3.27} & \textbf{4.31} & \cellcolor{gray!30}{4.18} & \cellcolor{gray!30}{5.37} & \underline{4.40} & \cellcolor{gray!30}{3.18} & \cellcolor{gray!30}{3.16} \\
m4\_hourly/H/short & \cellcolor{gray!30}{0.719} $\pm$ 0.026 & \cellcolor{gray!30}{0.837} & \cellcolor{gray!30}{0.866} & \cellcolor{gray!30}{0.596} & \cellcolor{gray!30}{0.768} & \textbf{0.780} & \cellcolor{gray!30}{0.886} & \cellcolor{gray!30}{0.971} & \underline{2.78} & \cellcolor{gray!30}{0.693} & \cellcolor{gray!30}{0.739} \\
m4\_monthly/M/short & \cellcolor{gray!30}{0.929} $\pm$ 0.004 & \cellcolor{gray!30}{0.949} & \cellcolor{gray!30}{0.954} & \cellcolor{gray!30}{0.600} & \cellcolor{gray!30}{0.957} & \textbf{0.895} & \cellcolor{gray!30}{0.977} & \cellcolor{gray!30}{0.953} & \underline{1.53} & \cellcolor{gray!30}{0.973} & \cellcolor{gray!30}{0.982} \\
m4\_quarterly/Q/short & \underline{1.18} $\pm$ 0.013 & 1.22 & 1.25 & \cellcolor{gray!30}{0.965} & \cellcolor{gray!30}{1.40} & \textbf{1.17} & \cellcolor{gray!30}{1.14} & \cellcolor{gray!30}{1.14} & 2.03 & 1.23 & 1.24 \\
m4\_weekly/W/short & \cellcolor{gray!30}{1.90} $\pm$ 0.029 & \cellcolor{gray!30}{2.08} & \cellcolor{gray!30}{2.11} & \cellcolor{gray!30}{2.22} & \cellcolor{gray!30}{2.42} & \textbf{2.07} & \cellcolor{gray!30}{2.58} & \cellcolor{gray!30}{2.81} & \underline{3.48} & \cellcolor{gray!30}{2.08} & \cellcolor{gray!30}{2.09} \\
m4\_yearly/A/short & 3.45 $\pm$ 0.075 & 3.51 & 3.69 & \textbf{2.54} & 3.35 & 3.16 & \underline{2.97} & 3.01 & 5.13 & 3.64 & 3.74 \\
m\_dense/D/short & 0.688 $\pm$ 0.016 & 0.716 & 0.742 & \underline{0.636} & 0.702 & \textbf{0.634} & 0.957 & 1.10 & 1.21 & 0.712 & 0.834 \\
m\_dense/H/long & \underline{0.730} $\pm$ 0.013 & 0.938 & 0.913 & 0.795 & 0.787 & 1.04 & \textbf{0.696} & 0.734 & 1.06 & 0.773 & 0.737 \\
m\_dense/H/medium & 0.736 $\pm$ 0.016 & 0.881 & 0.820 & 0.771 & 0.765 & 1.02 & \textbf{0.684} & 0.734 & 1.02 & 0.757 & \underline{0.712} \\
m\_dense/H/short & 0.788 $\pm$ 0.009 & \textbf{0.775} & 0.805 & 0.848 & 0.849 & 0.916 & \underline{0.777} & 0.837 & 1.09 & 0.800 & 0.803 \\
restaurant/D/short & \textbf{0.677} $\pm$ 0.002 & 0.700 & 0.700 & \underline{0.692} & 0.704 & 0.782 & \cellcolor{gray!30}{0.715} & \cellcolor{gray!30}{0.704} & 0.897 & 0.728 & 0.758 \\
saugeen/D/short & 3.12 $\pm$ 0.108 & \textbf{2.84} & \underline{2.96} & 3.34 & 3.30 & 3.30 & \cellcolor{gray!30}{3.29} & \cellcolor{gray!30}{2.91} & \cellcolor{gray!30}{4.03} & 3.29 & 2.98 \\
saugeen/M/short & 0.750 $\pm$ 0.020 & 0.739 & \underline{0.727} & 0.836 & 0.814 & \textbf{0.707} & \cellcolor{gray!30}{0.756} & \cellcolor{gray!30}{0.834} & 0.790 & 0.854 & 0.992 \\
saugeen/W/short & \textbf{1.18} $\pm$ 0.028 & \underline{1.22} & 1.24 & 1.95 & 1.28 & 1.25 & 1.38 & 1.41 & 1.87 & 1.35 & 1.37 \\
solar/10T/long & \textbf{0.828} $\pm$ 0.021 & 1.07 & 1.19 & 1.15 & 1.58 & \underline{0.915} & 1.95 & 2.02 & 1.10 & 1.66 & 1.98 \\
solar/10T/medium & \textbf{0.879} $\pm$ 0.037 & 1.03 & 1.06 & 1.17 & 1.38 & \underline{0.935} & 1.82 & 1.89 & 1.13 & 1.54 & 1.79 \\
solar/10T/short & 1.05 $\pm$ 0.032 & \underline{0.991} & \textbf{0.947} & 1.48 & 1.49 & 1.08 & 1.11 & 1.10 & 1.18 & 1.11 & 1.22 \\
solar/D/short & \textbf{0.971} $\pm$ 0.005 & \underline{0.982} & 0.995 & \textbf{0.971} & 0.990 & 0.985 & 0.987 & 1.02 & 1.07 & 1.01 & 1.08 \\
solar/H/long & \textbf{0.697} $\pm$ 0.024 & 1.03 & \underline{0.957} & 1.27 & 1.44 & 0.977 & 1.02 & 1.07 & 1.12 & 1.07 & 1.01 \\
solar/H/medium & \textbf{0.731} $\pm$ 0.026 & 0.931 & 0.926 & 0.959 & 1.04 & 0.921 & 0.917 & 0.892 & 1.05 & \underline{0.806} & 0.815 \\
solar/H/short & \textbf{0.699} $\pm$ 0.019 & \underline{0.813} & 0.852 & 1.02 & 1.04 & 0.937 & 0.875 & 0.893 & 0.980 & 0.827 & 0.855 \\
solar/W/short & 1.13 $\pm$ 0.075 & 0.980 & 0.991 & 1.28 & 1.15 & \underline{0.878} & 1.53 & 1.66 & 2.88 & 1.15 & \textbf{0.877} \\
sz\_taxi/15T/long & \textbf{0.509} $\pm$ 0.003 & 0.545 & \underline{0.538} & \cellcolor{gray!30}{0.514} & \cellcolor{gray!30}{0.535} & 0.560 & \cellcolor{gray!30}{0.554} & \cellcolor{gray!30}{0.537} & \cellcolor{gray!30}{0.531} & 0.567 & 0.584 \\
sz\_taxi/15T/medium & \textbf{0.536} $\pm$ 0.001 & \underline{0.559} & 0.562 & \cellcolor{gray!30}{0.546} & \cellcolor{gray!30}{0.558} & 0.585 & \cellcolor{gray!30}{0.569} & \cellcolor{gray!30}{0.558} & 0.566 & 0.597 & 0.618 \\
sz\_taxi/15T/short & \textbf{0.544} $\pm$ 0.001 & \underline{0.548} & 0.550 & \cellcolor{gray!30}{0.539} & \cellcolor{gray!30}{0.558} & 0.571 & \cellcolor{gray!30}{0.581} & \cellcolor{gray!30}{0.576} & \cellcolor{gray!30}{0.574} & 0.589 & 0.598 \\
sz\_taxi/H/short & 0.563 $\pm$ 0.002 & \underline{0.562} & 0.567 & \textbf{0.560} & 0.566 & 0.582 & 0.601 & 0.588 & \cellcolor{gray!30}{0.597} & 0.576 & 0.579 \\
temperature\_rain/D/short & \cellcolor{gray!30}{1.34} $\pm$ 0.003 & \cellcolor{gray!30}{1.30} & \cellcolor{gray!30}{1.32} & 1.43 & \underline{1.42} & \textbf{1.38} & \cellcolor{gray!30}{1.20} & \cellcolor{gray!30}{1.31} & 1.66 & \cellcolor{gray!30}{1.41} & \cellcolor{gray!30}{1.44} \\
us\_births/D/short & 0.404 $\pm$ 0.017 & 0.485 & 0.528 & \underline{0.370} & 0.552 & \textbf{0.320} & \cellcolor{gray!30}{0.503} & \cellcolor{gray!30}{0.509} & \cellcolor{gray!30}{1.63} & 0.420 & 0.436 \\
us\_births/M/short & 0.808 $\pm$ 0.066 & 0.924 & 0.756 & \textbf{0.497} & 0.622 & 0.588 & 0.771 & 0.723 & 1.32 & 0.778 & \underline{0.572} \\
us\_births/W/short & 1.08 $\pm$ 0.032 & 1.09 & 1.12 & 1.10 & 1.06 & \underline{0.929} & 1.47 & 1.44 & 1.78 & 0.932 & \textbf{0.921} \\
\bottomrule
\end{tabular}

}
\label{tab:gifteval-zs-zs-model-MASE-p2}
\end{table}

\setNextCaption{CRPS scores of different zero-shot models on the \underline{GiftEval} benchmark evaluation settings (Part 1/2). The models achieving the \textbf{best} and \underline{second-best} scores are highlighted. Results for datasets that are part of the training data for the respective models are shaded in grey, and these results are excluded from the calculation of the best score.
We trained \modelname with 6 different seeds and report the observed standard deviation in the plot.}

\begin{table}[h]
\centering
\caption{\nextcaption}
{\scriptsize
\begin{tabular}{llllllllllll}
\toprule
  & \rotatebox{90}{  \modelname } & \rotatebox{90}{ Chronos Bolt B } & \rotatebox{90}{ Chronos Bolt S } & \rotatebox{90}{ TimesFM 2.0 } & \rotatebox{90}{ TimesFM 1.0 } & \rotatebox{90}{ TabPFN-TS } & \rotatebox{90}{ Moirai L 1.1 } & \rotatebox{90}{ Moirai B 1.1 } & \rotatebox{90}{ TTM-r2 } & \rotatebox{90}{ Chronos B } & \rotatebox{90}{ Chronos S }\\
\midrule
bitbrains\_fast\_storage/5T/long & \textbf{0.655} $\pm$ 0.028 & 0.748 & 0.753 & 0.908 & 0.806 & 0.760 & 0.716 & 0.732 & 0.939 & 0.711 & \underline{0.709} \\
bitbrains\_fast\_storage/5T/medium & \textbf{0.605} $\pm$ 0.016 & 0.755 & 0.867 & 0.881 & 0.746 & 0.735 & \underline{0.636} & 0.662 & 0.906 & 0.804 & 0.803 \\
bitbrains\_fast\_storage/5T/short & \textbf{0.408} $\pm$ 0.005 & 0.454 & 0.435 & 0.447 & 0.476 & 0.456 & \underline{0.412} & 0.413 & 0.596 & 0.463 & 0.446 \\
bitbrains\_fast\_storage/H/short & 0.699 $\pm$ 0.013 & 0.774 & \textbf{0.589} & 0.688 & 0.699 & \underline{0.591} & 0.646 & 0.613 & 0.926 & 0.622 & 0.614 \\
bitbrains\_rnd/5T/long & \textbf{0.660} $\pm$ 0.065 & 0.756 & 0.756 & 0.706 & 0.806 & 0.730 & 0.678 & \underline{0.665} & 0.779 & 1.05 & 1.08 \\
bitbrains\_rnd/5T/medium & \textbf{0.594} $\pm$ 0.019 & \underline{0.605} & 0.792 & 0.727 & 0.734 & 0.710 & \textbf{0.594} & 0.616 & 0.835 & 0.644 & 0.615 \\
bitbrains\_rnd/5T/short & \textbf{0.403} $\pm$ 0.001 & 0.438 & 0.453 & 0.461 & 0.519 & 0.455 & \underline{0.418} & 0.446 & 0.605 & 0.507 & 0.493 \\
bitbrains\_rnd/H/short & 0.631 $\pm$ 0.013 & 0.624 & 0.623 & 0.649 & 0.654 & 0.637 & \textbf{0.566} & \underline{0.580} & 1.06 & 0.665 & 0.614 \\
bizitobs\_application/10S/long & \textbf{0.053} $\pm$ 0.004 & 0.109 & 0.092 & \underline{0.057} & 0.128 & 0.088 & 0.094 & 0.120 & 0.144 & 0.093 & 0.093 \\
bizitobs\_application/10S/medium & \underline{0.041} $\pm$ 0.003 & 0.104 & 0.085 & \textbf{0.033} & 0.136 & 0.070 & 0.084 & 0.104 & 0.126 & 0.117 & 0.078 \\
bizitobs\_application/10S/short & \textbf{0.013} $\pm$ 0.001 & 0.054 & 0.035 & \underline{0.014} & 0.056 & 0.031 & 0.038 & 0.033 & 0.058 & 0.031 & 0.031 \\
bizitobs\_l2c/5T/long & 0.581 $\pm$ 0.032 & 0.738 & 0.790 & 0.748 & 0.734 & \underline{0.511} & \textbf{0.508} & 0.554 & 0.785 & 0.722 & 0.743 \\
bizitobs\_l2c/5T/medium & \underline{0.366} $\pm$ 0.015 & 0.445 & 0.462 & 0.529 & 0.449 & \textbf{0.345} & 0.410 & 0.380 & 0.540 & 0.484 & 0.465 \\
bizitobs\_l2c/5T/short & 0.078 $\pm$ 0.002 & \underline{0.074} & \textbf{0.073} & 0.084 & 0.080 & 0.099 & 0.079 & 0.078 & 0.107 & 0.084 & 0.087 \\
bizitobs\_l2c/H/long & \textbf{0.276} $\pm$ 0.010 & \underline{0.278} & 0.295 & 0.728 & 0.724 & 0.440 & 0.600 & 0.495 & 0.751 & 0.738 & 0.780 \\
bizitobs\_l2c/H/medium & \textbf{0.253} $\pm$ 0.008 & \underline{0.254} & 0.285 & 0.640 & 0.856 & 0.380 & 0.619 & 0.688 & 0.782 & 0.793 & 0.780 \\
bizitobs\_l2c/H/short & 0.227 $\pm$ 0.011 & \textbf{0.189} & \underline{0.204} & 0.345 & 0.485 & 0.290 & 0.559 & 0.493 & 0.549 & 0.469 & 0.428 \\
bizitobs\_service/10S/long & \textbf{0.054} $\pm$ 0.002 & 0.113 & 0.096 & \underline{0.062} & 0.137 & 0.091 & 0.104 & 0.115 & 0.140 & 0.094 & 0.093 \\
bizitobs\_service/10S/medium & \textbf{0.034} $\pm$ 0.002 & 0.096 & 0.082 & \underline{0.038} & 0.109 & 0.067 & 0.069 & 0.090 & 0.117 & 0.073 & 0.068 \\
bizitobs\_service/10S/short & \textbf{0.013} $\pm$ 0.000 & 0.051 & 0.032 & \underline{0.015} & 0.051 & 0.031 & 0.032 & 0.042 & 0.053 & 0.027 & 0.027 \\
car\_parts/M/short & \underline{0.990} $\pm$ 0.010 & 0.995 & 1.01 & 1.05 & 1.02 & \textbf{0.955} & \cellcolor{gray!30}{1.18} & \cellcolor{gray!30}{0.999} & 2.29 & 1.07 & 1.03 \\
covid\_deaths/D/short & \textbf{0.037} $\pm$ 0.004 & 0.047 & 0.043 & 0.062 & 0.204 & \underline{0.040} & \cellcolor{gray!30}{0.046} & \cellcolor{gray!30}{0.044} & 0.123 & 0.045 & 0.061 \\
electricity/15T/long & \cellcolor{gray!30}{0.075} $\pm$ 0.001 & \cellcolor{gray!30}{0.084} & \cellcolor{gray!30}{0.086} & \textbf{0.083} & 0.137 & \underline{0.089} & 0.099 & 0.115 & 0.143 & \cellcolor{gray!30}{0.095} & \cellcolor{gray!30}{0.098} \\
electricity/15T/medium & \cellcolor{gray!30}{0.075} $\pm$ 0.001 & \cellcolor{gray!30}{0.083} & \cellcolor{gray!30}{0.087} & \textbf{0.080} & 0.138 & \underline{0.092} & 0.103 & 0.106 & 0.142 & \cellcolor{gray!30}{0.095} & \cellcolor{gray!30}{0.096} \\
electricity/15T/short & \cellcolor{gray!30}{0.082} $\pm$ 0.000 & \cellcolor{gray!30}{0.082} & \cellcolor{gray!30}{0.082} & \textbf{0.079} & 0.130 & \underline{0.104} & 0.128 & 0.120 & 0.152 & \cellcolor{gray!30}{0.092} & \cellcolor{gray!30}{0.099} \\
electricity/D/short & \underline{0.056} $\pm$ 0.001 & \textbf{0.055} & 0.058 & 0.060 & 0.077 & 0.060 & 0.069 & 0.061 & 0.093 & 0.061 & 0.071 \\
electricity/H/long & \cellcolor{gray!30}{0.092} $\pm$ 0.003 & \cellcolor{gray!30}{0.098} & \cellcolor{gray!30}{0.102} & \underline{0.089} & 0.101 & 0.112 & 0.103 & \textbf{0.086} & 0.128 & \cellcolor{gray!30}{0.105} & \cellcolor{gray!30}{0.107} \\
electricity/H/medium & \cellcolor{gray!30}{0.078} $\pm$ 0.002 & \cellcolor{gray!30}{0.081} & \cellcolor{gray!30}{0.084} & \textbf{0.073} & \underline{0.082} & 0.091 & 0.087 & \underline{0.082} & 0.109 & \cellcolor{gray!30}{0.087} & \cellcolor{gray!30}{0.088} \\
electricity/H/short & \cellcolor{gray!30}{0.061} $\pm$ 0.001 & \cellcolor{gray!30}{0.064} & \cellcolor{gray!30}{0.067} & \textbf{0.054} & \underline{0.064} & 0.072 & 0.077 & 0.075 & 0.097 & \cellcolor{gray!30}{0.064} & \cellcolor{gray!30}{0.070} \\
electricity/W/short & \cellcolor{gray!30}{0.046} $\pm$ 0.001 & \cellcolor{gray!30}{0.047} & \cellcolor{gray!30}{0.048} & \textbf{0.049} & 0.088 & \underline{0.051} & 0.062 & 0.077 & 0.159 & \cellcolor{gray!30}{0.049} & \cellcolor{gray!30}{0.052} \\
ett1/15T/long & \textbf{0.246} $\pm$ 0.003 & 0.298 & 0.296 & 0.283 & 0.358 & \underline{0.260} & 0.358 & 0.273 & 0.352 & 0.400 & 0.450 \\
ett1/15T/medium & \underline{0.251} $\pm$ 0.002 & 0.281 & 0.288 & 0.278 & 0.329 & \textbf{0.248} & 0.342 & 0.324 & 0.333 & 0.379 & 0.390 \\
ett1/15T/short & \underline{0.161} $\pm$ 0.003 & \textbf{0.158} & 0.169 & 0.168 & 0.193 & 0.183 & 0.226 & 0.193 & 0.235 & 0.198 & 0.217 \\
ett1/D/short & 0.282 $\pm$ 0.004 & 0.287 & 0.283 & \underline{0.281} & \textbf{0.280} & 0.292 & 0.286 & 0.301 & 0.416 & 0.387 & 0.360 \\
ett1/H/long & \textbf{0.263} $\pm$ 0.004 & 0.311 & 0.337 & 0.310 & 0.317 & 0.290 & 0.296 & \underline{0.287} & 0.342 & 0.350 & 0.360 \\
ett1/H/medium & \textbf{0.253} $\pm$ 0.002 & 0.303 & 0.295 & 0.282 & 0.304 & 0.276 & \underline{0.270} & 0.282 & 0.339 & 0.330 & 0.327 \\
ett1/H/short & \textbf{0.179} $\pm$ 0.002 & \underline{0.181} & 0.189 & 0.192 & 0.209 & 0.194 & 0.189 & 0.197 & 0.250 & 0.194 & 0.222 \\
ett1/W/short & 0.306 $\pm$ 0.009 & 0.296 & 0.293 & 0.272 & 0.307 & \textbf{0.256} & \underline{0.260} & 0.261 & 0.448 & 0.312 & 0.317 \\
ett2/15T/long & \textbf{0.097} $\pm$ 0.001 & 0.111 & 0.118 & 0.106 & 0.119 & \underline{0.101} & 0.115 & 0.137 & 0.126 & 0.134 & 0.129 \\
ett2/15T/medium & \textbf{0.093} $\pm$ 0.001 & 0.110 & 0.119 & 0.105 & 0.112 & \underline{0.098} & 0.105 & 0.109 & 0.128 & 0.122 & 0.117 \\
ett2/15T/short & \underline{0.066} $\pm$ 0.001 & 0.067 & 0.070 & \textbf{0.065} & 0.077 & 0.073 & 0.080 & 0.078 & 0.093 & 0.071 & 0.073 \\
ett2/D/short & \underline{0.092} $\pm$ 0.001 & 0.094 & \textbf{0.091} & 0.108 & 0.113 & 0.129 & 0.094 & 0.095 & 0.119 & \underline{0.092} & 0.097 \\
ett2/H/long & \underline{0.116} $\pm$ 0.004 & 0.117 & 0.121 & 0.125 & 0.125 & 0.136 & 0.125 & \textbf{0.110} & 0.144 & 0.136 & 0.122 \\
ett2/H/medium & \underline{0.106} $\pm$ 0.002 & 0.115 & 0.118 & 0.110 & 0.126 & 0.128 & 0.118 & \textbf{0.100} & 0.139 & 0.132 & 0.136 \\
ett2/H/short & \underline{0.065} $\pm$ 0.001 & \textbf{0.063} & \underline{0.065} & 0.066 & 0.074 & 0.070 & 0.069 & 0.072 & 0.088 & 0.071 & 0.072 \\
ett2/W/short & 0.088 $\pm$ 0.002 & 0.088 & 0.094 & 0.110 & 0.111 & 0.120 & 0.109 & \underline{0.087} & 0.200 & \textbf{0.077} & \textbf{0.077} \\
hierarchical\_sales/D/short & \textbf{0.572} $\pm$ 0.002 & 0.576 & 0.582 & 0.576 & \underline{0.573} & 0.593 & \cellcolor{gray!30}{0.580} & \cellcolor{gray!30}{0.575} & 0.792 & 0.600 & 0.619 \\
hierarchical\_sales/W/short & 0.349 $\pm$ 0.003 & 0.353 & 0.354 & \textbf{0.330} & 0.343 & \underline{0.342} & 0.359 & 0.357 & 0.725 & 0.367 & 0.367 \\
hospital/M/short & \underline{0.052} $\pm$ 0.000 & 0.057 & 0.058 & \textbf{0.050} & \underline{0.052} & \textbf{0.050} & \cellcolor{gray!30}{0.051} & \cellcolor{gray!30}{0.051} & 0.123 & 0.056 & 0.056 \\
\bottomrule
\end{tabular}

}
\label{tab:gifteval-zs-zs-model-WQL-p1}
\end{table}

\setNextCaption{CRPS scores of different zero-shot models on the \underline{GiftEval} benchmark evaluation settings (Part 2/2). The models achieving the \textbf{best} and \underline{second-best} scores are highlighted. Results for datasets that are part of the training data for the respective models are shaded in grey, and these results are excluded from the calculation of the best score. We trained \modelname with 6 different seeds and report the observed standard deviation in the plot.}

\begin{table}[h]
\centering
\caption{\nextcaption}
{\scriptsize
\begin{tabular}{llllllllllll}
\toprule
  & \rotatebox{90}{  \modelname } & \rotatebox{90}{ Chronos Bolt B } & \rotatebox{90}{ Chronos Bolt S } & \rotatebox{90}{ TimesFM 2.0 } & \rotatebox{90}{ TimesFM 1.0 } & \rotatebox{90}{ TabPFN-TS } & \rotatebox{90}{ Moirai L 1.1 } & \rotatebox{90}{ Moirai B 1.1 } & \rotatebox{90}{ TTM-r2 } & \rotatebox{90}{ Chronos B } & \rotatebox{90}{ Chronos S }\\
\midrule
jena\_weather/10T/long & \textbf{0.053} $\pm$ 0.001 & 0.064 & 0.063 & \cellcolor{gray!30}{0.035} & \cellcolor{gray!30}{0.069} & \underline{0.055} & 0.077 & 0.070 & 0.068 & 0.080 & 0.096 \\
jena\_weather/10T/medium & \textbf{0.051} $\pm$ 0.001 & \underline{0.057} & 0.060 & \cellcolor{gray!30}{0.031} & \cellcolor{gray!30}{0.067} & \underline{0.057} & 0.072 & 0.068 & 0.069 & 0.076 & 0.089 \\
jena\_weather/10T/short & \textbf{0.030} $\pm$ 0.001 & \underline{0.033} & 0.037 & \cellcolor{gray!30}{0.016} & \cellcolor{gray!30}{0.036} & 0.035 & 0.051 & 0.053 & 0.045 & 0.044 & 0.047 \\
jena\_weather/D/short & \underline{0.046} $\pm$ 0.001 & \textbf{0.045} & 0.047 & 0.058 & 0.059 & 0.047 & 0.051 & 0.050 & 0.124 & 0.049 & 0.051 \\
jena\_weather/H/long & \textbf{0.057} $\pm$ 0.002 & 0.062 & 0.068 & 0.068 & 0.089 & 0.066 & \underline{0.061} & 0.065 & 0.084 & 0.074 & 0.072 \\
jena\_weather/H/medium & \textbf{0.051} $\pm$ 0.002 & \underline{0.054} & 0.058 & 0.066 & 0.065 & 0.060 & 0.058 & 0.057 & 0.073 & 0.070 & 0.069 \\
jena\_weather/H/short & \textbf{0.041} $\pm$ 0.000 & \underline{0.042} & 0.043 & 0.045 & 0.048 & 0.043 & 0.045 & 0.044 & 0.060 & 0.046 & 0.047 \\
kdd\_cup\_2018/D/short & \cellcolor{gray!30}{0.381} $\pm$ 0.005 & \cellcolor{gray!30}{0.372} & \cellcolor{gray!30}{0.373} & \underline{0.378} & 0.380 & \textbf{0.359} & \cellcolor{gray!30}{0.381} & \cellcolor{gray!30}{0.376} & 0.452 & \cellcolor{gray!30}{0.503} & \cellcolor{gray!30}{0.512} \\
kdd\_cup\_2018/H/long & \cellcolor{gray!30}{0.341} $\pm$ 0.014 & \cellcolor{gray!30}{0.300} & \cellcolor{gray!30}{0.419} & \underline{0.518} & 0.537 & \textbf{0.462} & \cellcolor{gray!30}{0.378} & \cellcolor{gray!30}{0.418} & 0.542 & \cellcolor{gray!30}{0.624} & \cellcolor{gray!30}{0.712} \\
kdd\_cup\_2018/H/medium & \cellcolor{gray!30}{0.337} $\pm$ 0.011 & \cellcolor{gray!30}{0.301} & \cellcolor{gray!30}{0.364} & \underline{0.466} & 0.493 & \textbf{0.462} & \cellcolor{gray!30}{0.387} & \cellcolor{gray!30}{0.441} & 0.532 & \cellcolor{gray!30}{0.664} & \cellcolor{gray!30}{0.706} \\
kdd\_cup\_2018/H/short & \cellcolor{gray!30}{0.270} $\pm$ 0.004 & \cellcolor{gray!30}{0.246} & \cellcolor{gray!30}{0.267} & \textbf{0.376} & 0.446 & \underline{0.437} & \cellcolor{gray!30}{0.362} & \cellcolor{gray!30}{0.389} & \cellcolor{gray!30}{0.514} & \cellcolor{gray!30}{0.459} & \cellcolor{gray!30}{0.459} \\
loop\_seattle/5T/long & \textbf{0.090} $\pm$ 0.004 & 0.129 & 0.125 & 0.114 & 0.148 & \underline{0.094} & \cellcolor{gray!30}{0.049} & \cellcolor{gray!30}{0.052} & \cellcolor{gray!30}{0.125} & 0.143 & 0.150 \\
loop\_seattle/5T/medium & \textbf{0.083} $\pm$ 0.002 & 0.116 & 0.119 & 0.110 & 0.151 & \underline{0.087} & \cellcolor{gray!30}{0.038} & \cellcolor{gray!30}{0.045} & \cellcolor{gray!30}{0.121} & 0.176 & 0.175 \\
loop\_seattle/5T/short & \textbf{0.049} $\pm$ 0.000 & 0.055 & 0.055 & \underline{0.051} & 0.075 & 0.052 & \cellcolor{gray!30}{0.041} & \cellcolor{gray!30}{0.046} & \cellcolor{gray!30}{0.068} & 0.070 & 0.070 \\
loop\_seattle/D/short & \underline{0.042} $\pm$ 0.000 & 0.044 & 0.045 & \textbf{0.041} & 0.043 & 0.044 & 0.045 & 0.044 & \cellcolor{gray!30}{0.101} & 0.045 & 0.048 \\
loop\_seattle/H/long & \textbf{0.063} $\pm$ 0.001 & 0.076 & 0.082 & \underline{0.066} & 0.097 & \textbf{0.063} & 0.074 & 0.080 & \cellcolor{gray!30}{0.097} & 0.082 & 0.089 \\
loop\_seattle/H/medium & \textbf{0.065} $\pm$ 0.001 & 0.076 & 0.082 & \underline{0.067} & 0.100 & \textbf{0.065} & 0.070 & 0.080 & \cellcolor{gray!30}{0.104} & 0.084 & 0.088 \\
loop\_seattle/H/short & \textbf{0.059} $\pm$ 0.000 & 0.065 & 0.066 & \textbf{0.059} & 0.082 & \underline{0.064} & 0.066 & 0.074 & \cellcolor{gray!30}{0.095} & 0.066 & 0.069 \\
m4\_daily/D/short & \cellcolor{gray!30}{0.021} $\pm$ 0.000 & \cellcolor{gray!30}{0.021} & \cellcolor{gray!30}{0.021} & \cellcolor{gray!30}{0.021} & \cellcolor{gray!30}{0.021} & \textbf{0.024} & \cellcolor{gray!30}{0.030} & \cellcolor{gray!30}{0.040} & \underline{0.035} & \cellcolor{gray!30}{0.022} & \cellcolor{gray!30}{0.021} \\
m4\_hourly/H/short & \cellcolor{gray!30}{0.021} $\pm$ 0.000 & \cellcolor{gray!30}{0.025} & \cellcolor{gray!30}{0.020} & \cellcolor{gray!30}{0.011} & \cellcolor{gray!30}{0.021} & \textbf{0.028} & \cellcolor{gray!30}{0.020} & \cellcolor{gray!30}{0.022} & \underline{0.040} & \cellcolor{gray!30}{0.024} & \cellcolor{gray!30}{0.025} \\
m4\_monthly/M/short & \cellcolor{gray!30}{0.093} $\pm$ 0.000 & \cellcolor{gray!30}{0.094} & \cellcolor{gray!30}{0.094} & \cellcolor{gray!30}{0.067} & \cellcolor{gray!30}{0.097} & \textbf{0.088} & \cellcolor{gray!30}{0.095} & \cellcolor{gray!30}{0.094} & \underline{0.177} & \cellcolor{gray!30}{0.104} & \cellcolor{gray!30}{0.103} \\
m4\_quarterly/Q/short & \textbf{0.074} $\pm$ 0.000 & 0.077 & 0.078 & \cellcolor{gray!30}{0.062} & \cellcolor{gray!30}{0.085} & \underline{0.075} & \cellcolor{gray!30}{0.073} & \cellcolor{gray!30}{0.073} & 0.139 & 0.083 & 0.084 \\
m4\_weekly/W/short & \cellcolor{gray!30}{0.035} $\pm$ 0.001 & \cellcolor{gray!30}{0.038} & \cellcolor{gray!30}{0.038} & \cellcolor{gray!30}{0.042} & \cellcolor{gray!30}{0.041} & \textbf{0.036} & \cellcolor{gray!30}{0.046} & \cellcolor{gray!30}{0.048} & \underline{0.069} & \cellcolor{gray!30}{0.037} & \cellcolor{gray!30}{0.040} \\
m4\_yearly/A/short & 0.119 $\pm$ 0.002 & 0.121 & 0.128 & \textbf{0.091} & 0.117 & 0.113 & \underline{0.104} & 0.105 & 0.197 & 0.135 & 0.139 \\
m\_dense/D/short & 0.066 $\pm$ 0.002 & 0.069 & 0.072 & \underline{0.060} & 0.070 & \textbf{0.057} & 0.095 & 0.104 & 0.151 & 0.075 & 0.087 \\
m\_dense/H/long & \underline{0.122} $\pm$ 0.003 & 0.170 & 0.146 & 0.127 & 0.135 & 0.164 & \textbf{0.114} & \underline{0.122} & 0.222 & 0.135 & 0.133 \\
m\_dense/H/medium & \underline{0.121} $\pm$ 0.002 & 0.157 & 0.134 & 0.127 & 0.132 & 0.159 & \textbf{0.112} & 0.123 & 0.213 & 0.136 & 0.128 \\
m\_dense/H/short & 0.130 $\pm$ 0.001 & \textbf{0.125} & 0.133 & 0.139 & 0.140 & 0.154 & \underline{0.128} & 0.140 & 0.225 & 0.137 & 0.140 \\
restaurant/D/short & \textbf{0.254} $\pm$ 0.001 & 0.264 & 0.264 & \underline{0.261} & 0.265 & 0.297 & \cellcolor{gray!30}{0.270} & \cellcolor{gray!30}{0.266} & 0.438 & 0.279 & 0.292 \\
saugeen/D/short & 0.382 $\pm$ 0.014 & \textbf{0.338} & \underline{0.354} & 0.408 & 0.417 & 0.384 & \cellcolor{gray!30}{0.406} & \cellcolor{gray!30}{0.354} & \cellcolor{gray!30}{0.589} & 0.432 & 0.387 \\
saugeen/M/short & 0.303 $\pm$ 0.009 & 0.296 & \underline{0.293} & 0.342 & 0.328 & \textbf{0.278} & \cellcolor{gray!30}{0.324} & \cellcolor{gray!30}{0.348} & 0.405 & 0.408 & 0.464 \\
saugeen/W/short & \textbf{0.353} $\pm$ 0.009 & \underline{0.363} & 0.372 & 0.601 & 0.382 & 0.380 & 0.430 & 0.423 & 0.696 & 0.473 & 0.482 \\
solar/10T/long & \textbf{0.328} $\pm$ 0.008 & 0.443 & 0.497 & 0.498 & 0.703 & \underline{0.352} & 0.771 & 0.903 & 0.545 & 0.748 & 0.901 \\
solar/10T/medium & \textbf{0.354} $\pm$ 0.016 & 0.436 & 0.453 & 0.516 & 0.623 & \underline{0.359} & 0.747 & 0.832 & 0.573 & 0.686 & 0.796 \\
solar/10T/short & 0.542 $\pm$ 0.017 & \underline{0.511} & \textbf{0.498} & 0.804 & 0.871 & 0.545 & 0.596 & 0.614 & 0.785 & 0.579 & 0.635 \\
solar/D/short & 0.281 $\pm$ 0.002 & 0.287 & 0.286 & \underline{0.278} & 0.288 & \textbf{0.269} & 0.292 & 0.295 & 0.396 & 0.326 & 0.337 \\
solar/H/long & \textbf{0.243} $\pm$ 0.008 & 0.405 & 0.373 & 0.493 & 0.572 & \underline{0.324} & 0.347 & 0.360 & 0.512 & 0.464 & 0.441 \\
solar/H/medium & \textbf{0.260} $\pm$ 0.010 & 0.368 & 0.356 & 0.376 & 0.425 & \underline{0.324} & 0.346 & 0.331 & 0.493 & 0.356 & 0.361 \\
solar/H/short & \textbf{0.259} $\pm$ 0.009 & \underline{0.298} & 0.303 & 0.406 & 0.403 & 0.358 & 0.333 & 0.338 & 0.468 & 0.334 & 0.345 \\
solar/W/short & 0.154 $\pm$ 0.008 & \underline{0.133} & 0.136 & 0.171 & 0.157 & \textbf{0.124} & 0.213 & 0.235 & 0.531 & 0.161 & \textbf{0.124} \\
sz\_taxi/15T/long & \textbf{0.197} $\pm$ 0.001 & 0.248 & \underline{0.245} & \cellcolor{gray!30}{0.227} & \cellcolor{gray!30}{0.238} & 0.248 & \cellcolor{gray!30}{0.213} & \cellcolor{gray!30}{0.209} & \cellcolor{gray!30}{0.260} & 0.265 & 0.275 \\
sz\_taxi/15T/medium & \textbf{0.202} $\pm$ 0.001 & \underline{0.244} & 0.246 & \cellcolor{gray!30}{0.229} & \cellcolor{gray!30}{0.233} & 0.245 & \cellcolor{gray!30}{0.215} & \cellcolor{gray!30}{0.211} & 0.270 & 0.268 & 0.279 \\
sz\_taxi/15T/short & \textbf{0.200} $\pm$ 0.000 & \underline{0.202} & 0.203 & \cellcolor{gray!30}{0.199} & \cellcolor{gray!30}{0.206} & 0.215 & \cellcolor{gray!30}{0.215} & \cellcolor{gray!30}{0.213} & \cellcolor{gray!30}{0.268} & 0.236 & 0.241 \\
sz\_taxi/H/short & \underline{0.136} $\pm$ 0.000 & \underline{0.136} & 0.137 & \textbf{0.135} & 0.137 & 0.144 & 0.146 & 0.143 & \cellcolor{gray!30}{0.183} & 0.149 & 0.149 \\
temperature\_rain/D/short & \cellcolor{gray!30}{0.550} $\pm$ 0.002 & \cellcolor{gray!30}{0.538} & \cellcolor{gray!30}{0.544} & 0.586 & \underline{0.581} & \textbf{0.565} & \cellcolor{gray!30}{0.479} & \cellcolor{gray!30}{0.535} & 0.791 & \cellcolor{gray!30}{0.610} & \cellcolor{gray!30}{0.627} \\
us\_births/D/short & 0.021 $\pm$ 0.001 & 0.026 & 0.028 & \underline{0.019} & 0.029 & \textbf{0.018} & \cellcolor{gray!30}{0.027} & \cellcolor{gray!30}{0.027} & \cellcolor{gray!30}{0.104} & 0.022 & 0.023 \\
us\_births/M/short & 0.017 $\pm$ 0.002 & 0.019 & 0.016 & \textbf{0.011} & \underline{0.013} & \underline{0.013} & 0.016 & 0.015 & 0.036 & 0.018 & \underline{0.013} \\
us\_births/W/short & \underline{0.013} $\pm$ 0.000 & \underline{0.013} & \underline{0.013} & \underline{0.013} & \underline{0.013} & \textbf{0.011} & 0.018 & 0.017 & 0.027 & \textbf{0.011} & \textbf{0.011} \\
\bottomrule
\end{tabular}

}
\label{tab:gifteval-zs-zs-model-WQL-p2}
\end{table}

\setNextCaption{MASE scores of \modelname compared with various task-specific and local models on the \underline{GiftEval} benchmark evaluation settings (Part 1/2). Models achieving the \textbf{best} and \underline{second-best} scores are highlighted, while the grey shade indicates results on datasets \modelname trained on. We trained \modelname with 6 different seeds and report the observed standard deviation in the plot.}

\begin{table}[h]
\centering
\caption{\nextcaption}
{\scriptsize
\begin{tabular}{llllllllll}
\toprule
  & \rotatebox{90}{  \modelname } & \rotatebox{90}{ DeepAR } & \rotatebox{90}{ PatchTST } & \rotatebox{90}{ DLinear } & \rotatebox{90}{ TFT } & \rotatebox{90}{ N-Beats } & \rotatebox{90}{ Auto ARIMA } & \rotatebox{90}{ Auto Theta } & \rotatebox{90}{ Seas. Naive }\\
\midrule
bitbrains\_fast\_storage/5T/long & \textbf{0.916} $\pm$ 0.006 & 7.33 & \underline{1.14} & 3.47 & 1.21 & 1.40 & \underline{1.14} & 1.61 & \underline{1.14} \\
bitbrains\_fast\_storage/5T/medium & \textbf{1.00} $\pm$ 0.011 & 8.50 & \underline{1.20} & 3.33 & 1.38 & 1.60 & 1.22 & 1.42 & 1.22 \\
bitbrains\_fast\_storage/5T/short & \textbf{0.692} $\pm$ 0.007 & \underline{0.945} & 0.973 & 1.48 & 0.996 & 1.09 & 1.14 & 1.15 & 1.14 \\
bitbrains\_fast\_storage/H/short & \textbf{1.06} $\pm$ 0.013 & 6.06 & 1.34 & 2.65 & 1.73 & 1.37 & 1.43 & 1.35 & \underline{1.30} \\
bitbrains\_rnd/5T/long & \textbf{3.35} $\pm$ 0.013 & 4.44 & 3.72 & 6.35 & 3.71 & 3.95 & \underline{3.50} & 4.11 & \underline{3.50} \\
bitbrains\_rnd/5T/medium & \textbf{4.40} $\pm$ 0.007 & 4.89 & 4.65 & 7.08 & 4.81 & 4.79 & \underline{4.54} & 4.88 & \underline{4.54} \\
bitbrains\_rnd/5T/short & \textbf{1.66} $\pm$ 0.004 & 2.10 & 1.98 & 2.63 & 2.27 & 2.18 & \underline{1.97} & 2.07 & \underline{1.97} \\
bitbrains\_rnd/H/short & \underline{5.84} $\pm$ 0.011 & 6.06 & 6.11 & 8.50 & 6.19 & 6.16 & 6.08 & \textbf{5.75} & 6.04 \\
bizitobs\_application/10S/long & 3.67 $\pm$ 0.069 & 4.47 & \underline{3.19} & 4.25 & 14.8 & 3.83 & 36400 & \textbf{2.93} & 36400 \\
bizitobs\_application/10S/medium & 2.85 $\pm$ 0.057 & 3.22 & 2.77 & 3.88 & 13.8 & \underline{2.57} & 2.69 & \textbf{1.78} & 2.69 \\
bizitobs\_application/10S/short & \underline{1.40} $\pm$ 0.112 & 4.16 & 2.24 & 7.65 & 9.11 & 2.57 & 2.24 & \textbf{1.11} & 2.24 \\
bizitobs\_l2c/5T/long & 1.19 $\pm$ 0.040 & 1.32 & \textbf{0.686} & 1.12 & 1.06 & \underline{0.968} & 1.45 & 1.24 & 1.45 \\
bizitobs\_l2c/5T/medium & 0.849 $\pm$ 0.028 & 1.21 & \underline{0.787} & 0.894 & \textbf{0.786} & 0.935 & 1.24 & 0.868 & 1.24 \\
bizitobs\_l2c/5T/short & 0.290 $\pm$ 0.005 & 0.613 & \underline{0.266} & \textbf{0.243} & 0.278 & 0.297 & 0.986 & 0.292 & 0.986 \\
bizitobs\_l2c/H/long & \textbf{0.590} $\pm$ 0.022 & 0.727 & 0.617 & 0.744 & \underline{0.599} & 0.811 & 1.54 & 1.41 & 4.04 \\
bizitobs\_l2c/H/medium & \textbf{0.525} $\pm$ 0.020 & 0.737 & \underline{0.537} & 0.676 & 0.693 & 0.701 & 1.56 & 1.65 & 1.65 \\
bizitobs\_l2c/H/short & \underline{0.528} $\pm$ 0.021 & 1.50 & \textbf{0.495} & 0.640 & 0.862 & 0.536 & 1.25 & 1.19 & 1.21 \\
bizitobs\_service/10S/long & \underline{1.57} $\pm$ 0.057 & 3.96 & 1.69 & 2.29 & 1.75 & 2.62 & \textbf{1.37} & 1.62 & \textbf{1.37} \\
bizitobs\_service/10S/medium & \underline{1.32} $\pm$ 0.058 & 2.17 & 1.49 & 2.23 & 1.68 & 2.59 & \underline{1.32} & \textbf{1.06} & \underline{1.32} \\
bizitobs\_service/10S/short & \underline{0.884} $\pm$ 0.052 & 2.67 & 1.24 & 1.87 & 2.15 & 1.12 & 1.23 & \textbf{0.791} & 1.23 \\
car\_parts/M/short & 0.838 $\pm$ 0.004 & 0.835 & \textbf{0.797} & 0.997 & \underline{0.807} & 0.810 & 0.958 & 1.23 & 1.20 \\
covid\_deaths/D/short & 39.5 $\pm$ 0.803 & 50.7 & 37.7 & 33.2 & 32.9 & \underline{32.8} & \textbf{31.4} & 45.4 & 46.9 \\
electricity/15T/long & \cellcolor{gray!30}{0.891} $\pm$ 0.008 & 2.28 & \textbf{0.960} & 1.24 & \underline{1.03} & 1.15 & 1.16 & 1.50 & 1.16 \\
electricity/15T/medium & \cellcolor{gray!30}{0.841} $\pm$ 0.006 & 1.39 & \textbf{0.977} & 1.31 & \underline{1.11} & 1.62 & 1.15 & 1.43 & 1.15 \\
electricity/15T/short & \cellcolor{gray!30}{0.945} $\pm$ 0.008 & 1.67 & \underline{1.47} & 1.64 & 2.07 & 1.69 & 1.72 & \textbf{1.35} & 1.72 \\
electricity/D/short & \textbf{1.43} $\pm$ 0.010 & 1.89 & 1.85 & 3.56 & 1.86 & 1.85 & \underline{1.82} & 1.88 & 1.99 \\
electricity/H/long & \cellcolor{gray!30}{1.21} $\pm$ 0.018 & 2.67 & \textbf{1.39} & 2.21 & \underline{1.41} & 1.42 & 1.52 & 2.05 & 1.52 \\
electricity/H/medium & \cellcolor{gray!30}{1.08} $\pm$ 0.010 & 6.76 & \textbf{1.16} & 2.26 & \underline{1.31} & 1.35 & 1.39 & 1.78 & 1.39 \\
electricity/H/short & \cellcolor{gray!30}{0.869} $\pm$ 0.009 & \underline{1.23} & \textbf{1.08} & 1.31 & 1.29 & 1.44 & 1.36 & 1.74 & 1.36 \\
electricity/W/short & \cellcolor{gray!30}{1.46} $\pm$ 0.010 & 2.25 & \underline{1.96} & \textbf{1.84} & 2.10 & 2.10 & 2.09 & 2.14 & 2.09 \\
ett1/15T/long & \textbf{1.05} $\pm$ 0.009 & 9.34 & \underline{1.10} & 1.19 & 1.34 & 1.42 & 1.19 & 1.76 & 1.19 \\
ett1/15T/medium & \textbf{1.04} $\pm$ 0.009 & 1.35 & \underline{1.08} & 1.20 & \underline{1.08} & 1.41 & 1.19 & 1.25 & 1.19 \\
ett1/15T/short & \textbf{0.706} $\pm$ 0.007 & 1.44 & 0.835 & \underline{0.804} & 1.05 & 0.870 & 0.934 & 0.863 & 0.934 \\
ett1/D/short & 1.71 $\pm$ 0.016 & \underline{1.69} & \textbf{1.68} & 1.98 & 1.86 & 2.04 & 1.85 & 1.75 & 1.78 \\
ett1/H/long & \textbf{1.34} $\pm$ 0.030 & 2.68 & 1.47 & \underline{1.46} & 1.55 & 1.96 & 1.65 & 2.51 & 1.48 \\
ett1/H/medium & \textbf{1.25} $\pm$ 0.017 & 3.12 & \underline{1.39} & 1.66 & 1.58 & 1.67 & 1.57 & 1.84 & 1.57 \\
ett1/H/short & \textbf{0.827} $\pm$ 0.007 & 1.06 & \underline{0.893} & 0.945 & 0.947 & 0.930 & 0.995 & 1.28 & 0.977 \\
ett1/W/short & 1.72 $\pm$ 0.044 & 4.16 & 1.89 & 2.16 & \textbf{1.61} & \underline{1.63} & 1.99 & 1.89 & 1.77 \\
ett2/15T/long & \textbf{0.932} $\pm$ 0.012 & 3.70 & \underline{0.961} & 1.10 & 1.15 & 0.980 & 1.01 & 1.10 & 1.01 \\
ett2/15T/medium & \textbf{0.910} $\pm$ 0.010 & 3.27 & \underline{0.933} & 1.21 & 1.10 & 1.09 & 1.05 & 1.04 & 1.05 \\
ett2/15T/short & \textbf{0.749} $\pm$ 0.010 & 4.11 & 0.879 & 0.937 & 1.06 & 1.01 & 1.07 & \underline{0.832} & 1.07 \\
ett2/D/short & \textbf{1.28} $\pm$ 0.019 & 3.64 & 2.17 & 3.25 & \underline{1.31} & 1.54 & 1.45 & 1.85 & 1.39 \\
ett2/H/long & \underline{1.16} $\pm$ 0.037 & 2.49 & 1.43 & 1.58 & 1.45 & 1.26 & 1.28 & 1.46 & \textbf{1.13} \\
ett2/H/medium & \textbf{1.05} $\pm$ 0.021 & 2.52 & 1.27 & 1.36 & 1.32 & \textbf{1.05} & 1.46 & 1.30 & \underline{1.24} \\
ett2/H/short & \textbf{0.742} $\pm$ 0.006 & 1.48 & 0.858 & \underline{0.817} & 0.956 & 0.819 & 0.952 & 1.02 & 0.923 \\
ett2/W/short & \underline{0.797} $\pm$ 0.040 & 7.17 & 1.49 & 1.93 & 1.60 & 2.69 & 1.13 & 1.41 & \textbf{0.779} \\
hierarchical\_sales/D/short & \textbf{0.744} $\pm$ 0.002 & 0.757 & \underline{0.756} & 0.860 & 0.771 & 0.773 & 0.813 & 0.932 & 1.13 \\
hierarchical\_sales/W/short & \textbf{0.721} $\pm$ 0.001 & 0.781 & \underline{0.771} & 0.993 & 0.793 & 0.778 & 0.850 & 0.849 & 1.03 \\
hospital/M/short & \underline{0.767} $\pm$ 0.003 & 0.834 & 0.820 & 0.811 & 0.833 & 0.771 & 0.826 & \textbf{0.761} & 0.921 \\
\bottomrule
\end{tabular}

}
\label{tab:gifteval-zs-taskspec-MASE-p1}
\end{table}

\setNextCaption{MASE scores of \modelname compared with various task-specific and local models on the \underline{GiftEval} benchmark evaluation settings (Part 2/2). Models achieving the \textbf{best} and \underline{second-best} scores are highlighted, while the grey shade indicates results on datasets \modelname trained on. We trained \modelname with 6 different seeds and report the observed standard deviation in the plot.}

\begin{table}[h]
\centering
\caption{\nextcaption}
{\scriptsize
\begin{tabular}{llllllllll}
\toprule
  & \rotatebox{90}{  \modelname } & \rotatebox{90}{ DeepAR } & \rotatebox{90}{ PatchTST } & \rotatebox{90}{ DLinear } & \rotatebox{90}{ TFT } & \rotatebox{90}{ N-Beats } & \rotatebox{90}{ Auto ARIMA } & \rotatebox{90}{ Auto Theta } & \rotatebox{90}{ Seas. Naive }\\
\midrule
jena\_weather/10T/long & \textbf{0.641} $\pm$ 0.015 & 3.15 & 1.07 & 0.912 & \underline{0.741} & 0.855 & 0.761 & 0.990 & 0.761 \\
jena\_weather/10T/medium & \textbf{0.610} $\pm$ 0.005 & 1.20 & 0.943 & 1.17 & 0.737 & 0.749 & \underline{0.716} & 0.806 & \underline{0.716} \\
jena\_weather/10T/short & \textbf{0.297} $\pm$ 0.006 & 0.574 & 0.552 & 1.96 & 0.450 & 0.527 & 0.743 & \underline{0.368} & 0.743 \\
jena\_weather/D/short & \textbf{1.02} $\pm$ 0.014 & \underline{1.30} & 1.39 & 1.60 & 1.80 & 1.83 & 1.45 & 1.60 & 1.57 \\
jena\_weather/H/long & \textbf{0.987} $\pm$ 0.031 & 6.89 & 1.31 & 1.90 & 1.15 & \underline{1.13} & 1.98 & 2.64 & 1.27 \\
jena\_weather/H/medium & \textbf{0.828} $\pm$ 0.023 & 1.30 & 1.09 & 0.997 & 0.939 & 0.902 & 1.45 & 1.36 & \underline{0.889} \\
jena\_weather/H/short & \textbf{0.516} $\pm$ 0.003 & 18.8 & 0.641 & 0.982 & \underline{0.634} & 0.763 & 1.08 & 0.878 & 0.723 \\
kdd\_cup\_2018/D/short & \cellcolor{gray!30}{1.21} $\pm$ 0.009 & 1.23 & 1.22 & 1.23 & \underline{1.21} & 1.35 & \textbf{1.18} & 1.38 & 1.50 \\
kdd\_cup\_2018/H/long & \cellcolor{gray!30}{0.759} $\pm$ 0.027 & 3.34 & \textbf{1.02} & \underline{1.09} & 1.11 & 1.18 & 1.18 & 1.37 & 1.34 \\
kdd\_cup\_2018/H/medium & \cellcolor{gray!30}{0.825} $\pm$ 0.024 & 1.17 & \textbf{1.05} & \underline{1.12} & 1.16 & 1.29 & 1.42 & 1.33 & 1.43 \\
kdd\_cup\_2018/H/short & \cellcolor{gray!30}{0.657} $\pm$ 0.006 & 1.28 & \textbf{1.12} & \underline{1.13} & 1.15 & 1.30 & 1.34 & 1.27 & 1.34 \\
loop\_seattle/5T/long & \underline{1.02} $\pm$ 0.038 & 1.96 & 1.06 & 1.17 & \textbf{0.977} & 1.05 & 1.25 & 1.44 & 1.25 \\
loop\_seattle/5T/medium & \textbf{0.941} $\pm$ 0.019 & 1.27 & 1.05 & 1.12 & \underline{1.01} & 1.05 & 1.15 & 2.06 & 1.15 \\
loop\_seattle/5T/short & \textbf{0.572} $\pm$ 0.005 & 0.803 & 0.744 & 0.895 & \underline{0.731} & 0.735 & 0.762 & 0.780 & 0.762 \\
loop\_seattle/D/short & \textbf{0.878} $\pm$ 0.005 & 1.08 & 0.934 & 0.900 & 0.973 & \underline{0.898} & 1.49 & 1.39 & 1.73 \\
loop\_seattle/H/long & \textbf{0.917} $\pm$ 0.011 & 0.985 & 0.979 & 1.03 & 0.972 & \underline{0.932} & 2.59 & 2.02 & 1.55 \\
loop\_seattle/H/medium & \textbf{0.944} $\pm$ 0.012 & 1.03 & 1.03 & 1.20 & \underline{0.971} & 1.03 & 2.00 & 1.61 & 1.48 \\
loop\_seattle/H/short & \textbf{0.850} $\pm$ 0.005 & \underline{0.941} & 1.07 & 1.13 & 1.04 & 1.03 & 1.29 & 1.40 & 1.29 \\
m4\_daily/D/short & \cellcolor{gray!30}{3.15} $\pm$ 0.063 & 4.58 & \textbf{3.22} & 3.42 & 3.29 & 3.35 & \underline{3.26} & 3.34 & 3.28 \\
m4\_hourly/H/short & \cellcolor{gray!30}{0.719} $\pm$ 0.026 & 3.53 & 1.40 & 1.69 & 2.47 & 1.34 & \textbf{1.03} & 2.46 & \underline{1.19} \\
m4\_monthly/M/short & \cellcolor{gray!30}{0.929} $\pm$ 0.004 & 3.18 & 1.06 & 1.13 & 1.21 & 1.05 & \underline{0.976} & \textbf{0.966} & 1.26 \\
m4\_quarterly/Q/short & \textbf{1.18} $\pm$ 0.013 & 1.44 & 1.32 & 1.46 & 1.30 & 1.21 & 1.28 & \underline{1.19} & 1.60 \\
m4\_weekly/W/short & \cellcolor{gray!30}{1.90} $\pm$ 0.029 & 4.62 & \underline{2.34} & 4.64 & 2.68 & \textbf{1.97} & 2.36 & 2.66 & 2.78 \\
m4\_yearly/A/short & 3.45 $\pm$ 0.075 & 3.40 & 3.29 & 4.16 & \textbf{3.09} & 3.15 & 3.71 & \underline{3.11} & 3.97 \\
m\_dense/D/short & \textbf{0.688} $\pm$ 0.016 & 0.793 & 0.732 & 1.01 & 0.799 & \underline{0.706} & 1.34 & 1.22 & 1.67 \\
m\_dense/H/long & \underline{0.730} $\pm$ 0.013 & 0.805 & 0.738 & 1.24 & \textbf{0.723} & 1.18 & 1.21 & 2.29 & 1.48 \\
m\_dense/H/medium & \underline{0.736} $\pm$ 0.016 & 0.738 & 0.757 & 0.930 & \textbf{0.732} & 0.890 & 1.27 & 1.74 & 1.57 \\
m\_dense/H/short & \textbf{0.788} $\pm$ 0.009 & \underline{0.795} & 1.03 & 1.04 & 0.878 & 0.915 & 1.49 & 1.69 & 1.49 \\
restaurant/D/short & \textbf{0.677} $\pm$ 0.002 & 0.713 & \underline{0.690} & 0.706 & 0.750 & 0.712 & 0.929 & 0.843 & 1.01 \\
saugeen/D/short & \textbf{3.12} $\pm$ 0.108 & 4.31 & 3.28 & 4.20 & \underline{3.22} & 3.28 & 3.74 & 3.60 & 3.41 \\
saugeen/M/short & \underline{0.750} $\pm$ 0.020 & 1.63 & 0.893 & 0.955 & 0.865 & 0.758 & \textbf{0.725} & 0.912 & 0.976 \\
saugeen/W/short & \textbf{1.18} $\pm$ 0.028 & \underline{1.31} & 1.55 & 1.81 & 1.55 & 1.54 & 1.55 & 2.12 & 1.99 \\
solar/10T/long & \textbf{0.828} $\pm$ 0.021 & 1.28 & 0.912 & 1.18 & 1.00 & 2.03 & \underline{0.871} & 4.53 & \underline{0.871} \\
solar/10T/medium & \textbf{0.879} $\pm$ 0.037 & 1.21 & \underline{0.913} & 1.08 & 0.931 & 1.98 & 0.927 & 2.69 & 0.927 \\
solar/10T/short & \underline{1.05} $\pm$ 0.032 & 1.47 & 2.20 & 1.24 & 1.11 & \textbf{0.848} & 1.11 & 1.80 & 1.11 \\
solar/D/short & \underline{0.971} $\pm$ 0.005 & 2.49 & \textbf{0.962} & 1.03 & 0.999 & 1.21 & 1.01 & 1.05 & 1.16 \\
solar/H/long & \textbf{0.697} $\pm$ 0.024 & \underline{0.972} & 0.978 & 1.35 & 1.12 & 2.21 & 0.995 & 5.24 & 1.07 \\
solar/H/medium & \textbf{0.731} $\pm$ 0.026 & 0.992 & 0.965 & 1.17 & 0.884 & 2.12 & \underline{0.848} & 2.87 & 0.935 \\
solar/H/short & \textbf{0.699} $\pm$ 0.019 & 1.02 & 0.954 & 1.06 & 0.960 & 1.04 & \underline{0.952} & 2.05 & \underline{0.952} \\
solar/W/short & 1.13 $\pm$ 0.075 & 1.69 & \underline{1.10} & 1.13 & \textbf{0.691} & 2.33 & 1.12 & 1.15 & 1.47 \\
sz\_taxi/15T/long & \textbf{0.509} $\pm$ 0.003 & 0.733 & 0.761 & 0.841 & \underline{0.535} & 0.666 & 0.598 & 0.759 & 0.691 \\
sz\_taxi/15T/medium & \textbf{0.536} $\pm$ 0.001 & 0.558 & 0.588 & 0.629 & \underline{0.548} & 0.662 & 0.632 & 0.716 & 0.713 \\
sz\_taxi/15T/short & \textbf{0.544} $\pm$ 0.001 & 0.602 & \underline{0.560} & 0.582 & 0.603 & 0.604 & 0.764 & 0.649 & 0.764 \\
sz\_taxi/H/short & \textbf{0.563} $\pm$ 0.002 & \underline{0.576} & 0.591 & 0.667 & 0.595 & 0.624 & 0.624 & 0.691 & 0.738 \\
temperature\_rain/D/short & \cellcolor{gray!30}{1.34} $\pm$ 0.003 & 1.72 & \underline{1.51} & 1.83 & \textbf{1.44} & 1.90 & 1.71 & 1.93 & 2.01 \\
us\_births/D/short & \underline{0.404} $\pm$ 0.017 & 0.535 & 0.487 & 0.645 & \textbf{0.315} & 0.456 & 1.58 & 1.63 & 1.86 \\
us\_births/M/short & 0.808 $\pm$ 0.066 & \underline{0.760} & 0.782 & 1.17 & 0.871 & 0.928 & \textbf{0.466} & 0.883 & 0.761 \\
us\_births/W/short & \textbf{1.08} $\pm$ 0.032 & 1.45 & \underline{1.23} & 1.46 & 1.59 & 1.40 & 1.48 & 1.49 & 1.56 \\
\bottomrule
\end{tabular}

}
\label{tab:gifteval-zs-taskspec-MASE-p2}
\end{table}

\setNextCaption{CRPS scores of \modelname compared with various task-specific and local models on the \underline{GiftEval} benchmark evaluation settings (Part 1/2). Models achieving the \textbf{best} and \underline{second-best} scores are highlighted, while the grey shade indicates results on datasets \modelname trained on. We trained \modelname with 6 different seeds and report the observed standard deviation in the plot.}

\begin{table}[h]
\centering
\caption{\nextcaption}
{\scriptsize
\begin{tabular}{llllllllll}
\toprule
  & \rotatebox{90}{  \modelname } & \rotatebox{90}{ DeepAR } & \rotatebox{90}{ PatchTST } & \rotatebox{90}{ DLinear } & \rotatebox{90}{ TFT } & \rotatebox{90}{ N-Beats } & \rotatebox{90}{ Auto ARIMA } & \rotatebox{90}{ Auto Theta } & \rotatebox{90}{ Seas. Naive }\\
\midrule
bitbrains\_fast\_storage/5T/long & \textbf{0.655} $\pm$ 0.028 & 1.01 & \underline{0.669} & 1.33 & 0.734 & 0.791 & 1.29 & 1.36 & 1.29 \\
bitbrains\_fast\_storage/5T/medium & \textbf{0.605} $\pm$ 0.016 & 0.990 & 0.642 & 0.967 & \underline{0.610} & 0.841 & 1.27 & 1.45 & 1.27 \\
bitbrains\_fast\_storage/5T/short & \textbf{0.408} $\pm$ 0.005 & 0.493 & 0.471 & 0.577 & \underline{0.451} & 0.622 & 1.21 & 0.731 & 1.21 \\
bitbrains\_fast\_storage/H/short & 0.699 $\pm$ 0.013 & 0.778 & \textbf{0.549} & 0.803 & \underline{0.595} & 0.812 & 0.844 & 1.15 & 1.08 \\
bitbrains\_rnd/5T/long & \underline{0.660} $\pm$ 0.065 & 0.672 & 0.664 & 1.30 & \textbf{0.624} & 1.06 & 1.29 & 1.60 & 1.29 \\
bitbrains\_rnd/5T/medium & \textbf{0.594} $\pm$ 0.019 & 0.647 & \underline{0.620} & 1.01 & 0.628 & 0.699 & 1.26 & 1.47 & 1.26 \\
bitbrains\_rnd/5T/short & \textbf{0.403} $\pm$ 0.001 & 0.557 & \underline{0.474} & 0.571 & 0.486 & 0.656 & 1.10 & 0.741 & 1.10 \\
bitbrains\_rnd/H/short & 0.631 $\pm$ 0.013 & \textbf{0.585} & \underline{0.603} & 1.07 & 0.650 & 0.715 & 0.874 & 1.38 & 1.30 \\
bizitobs\_application/10S/long & \underline{0.053} $\pm$ 0.004 & 0.083 & 0.054 & 0.070 & 0.056 & 0.062 & 0.973 & \textbf{0.035} & 0.973 \\
bizitobs\_application/10S/medium & \underline{0.041} $\pm$ 0.003 & 0.053 & 0.047 & 0.056 & 0.047 & 0.047 & 0.042 & \textbf{0.024} & 0.042 \\
bizitobs\_application/10S/short & \underline{0.013} $\pm$ 0.001 & 0.064 & 0.022 & 0.079 & 0.090 & 0.043 & 0.035 & \textbf{0.010} & 0.035 \\
bizitobs\_l2c/5T/long & 0.581 $\pm$ 0.032 & 0.719 & \textbf{0.324} & 0.653 & \underline{0.472} & 0.546 & 0.674 & 0.632 & 0.674 \\
bizitobs\_l2c/5T/medium & 0.366 $\pm$ 0.015 & 0.589 & \textbf{0.332} & 0.490 & \underline{0.346} & 0.505 & 0.530 & 0.415 & 0.530 \\
bizitobs\_l2c/5T/short & 0.078 $\pm$ 0.002 & 0.179 & \textbf{0.074} & 0.080 & \underline{0.077} & 0.100 & 0.262 & 0.080 & 0.262 \\
bizitobs\_l2c/H/long & \textbf{0.276} $\pm$ 0.010 & 0.338 & 0.291 & 0.422 & \underline{0.286} & 0.479 & 0.787 & 0.819 & 1.82 \\
bizitobs\_l2c/H/medium & \textbf{0.253} $\pm$ 0.008 & 0.345 & \underline{0.263} & 0.398 & 0.345 & 0.420 & 0.813 & 0.892 & 1.42 \\
bizitobs\_l2c/H/short & \underline{0.227} $\pm$ 0.011 & 0.789 & \textbf{0.217} & 0.336 & 0.401 & 0.288 & 0.547 & 0.507 & 0.536 \\
bizitobs\_service/10S/long & \underline{0.054} $\pm$ 0.002 & 0.070 & 0.057 & 0.067 & 0.056 & 0.061 & 0.056 & \textbf{0.052} & 0.056 \\
bizitobs\_service/10S/medium & \underline{0.034} $\pm$ 0.002 & 0.044 & 0.045 & 0.053 & 0.044 & 0.043 & 0.049 & \textbf{0.027} & 0.049 \\
bizitobs\_service/10S/short & \textbf{0.013} $\pm$ 0.000 & 0.032 & 0.025 & 0.032 & 0.025 & \underline{0.021} & 0.040 & \textbf{0.013} & 0.040 \\
car\_parts/M/short & 0.990 $\pm$ 0.010 & \underline{0.953} & 1.00 & 1.26 & \textbf{0.890} & 1.02 & 1.29 & 1.34 & 1.72 \\
covid\_deaths/D/short & \underline{0.037} $\pm$ 0.004 & 0.177 & 0.067 & 0.063 & \underline{0.037} & 0.071 & \textbf{0.030} & 0.095 & 0.125 \\
electricity/15T/long & \cellcolor{gray!30}{0.075} $\pm$ 0.001 & 0.155 & \textbf{0.081} & 0.129 & \underline{0.084} & 0.123 & 0.129 & 0.401 & 0.129 \\
electricity/15T/medium & \cellcolor{gray!30}{0.075} $\pm$ 0.001 & 0.119 & \textbf{0.086} & 0.142 & \underline{0.094} & 0.176 & 0.124 & 0.328 & 0.124 \\
electricity/15T/short & \cellcolor{gray!30}{0.082} $\pm$ 0.000 & 0.152 & \textbf{0.134} & 0.177 & 0.184 & 0.180 & 0.165 & \underline{0.140} & 0.165 \\
electricity/D/short & \textbf{0.056} $\pm$ 0.001 & \underline{0.078} & 0.083 & 0.169 & 0.084 & 0.110 & 0.083 & 0.088 & 0.122 \\
electricity/H/long & \cellcolor{gray!30}{0.092} $\pm$ 0.003 & 0.176 & \underline{0.104} & 0.203 & \textbf{0.094} & 0.126 & 0.190 & 0.300 & 0.190 \\
electricity/H/medium & \cellcolor{gray!30}{0.078} $\pm$ 0.002 & 0.454 & \textbf{0.081} & 0.206 & \underline{0.091} & 0.115 & 0.156 & 0.254 & 0.156 \\
electricity/H/short & \cellcolor{gray!30}{0.061} $\pm$ 0.001 & 0.094 & \textbf{0.079} & 0.112 & \underline{0.089} & 0.123 & 0.109 & 0.177 & 0.109 \\
electricity/W/short & \cellcolor{gray!30}{0.046} $\pm$ 0.001 & \textbf{0.092} & \underline{0.095} & 0.111 & 0.107 & 0.123 & 0.100 & 0.101 & 0.099 \\
ett1/15T/long & \textbf{0.246} $\pm$ 0.003 & 2.22 & \underline{0.247} & 0.343 & 0.280 & 0.431 & 0.396 & 1.39 & 0.396 \\
ett1/15T/medium & 0.251 $\pm$ 0.002 & 0.315 & \underline{0.250} & 0.347 & \textbf{0.247} & 0.430 & 0.352 & 1.13 & 0.352 \\
ett1/15T/short & \textbf{0.161} $\pm$ 0.003 & 0.320 & \underline{0.191} & 0.233 & 0.245 & 0.254 & 0.241 & 0.410 & 0.241 \\
ett1/D/short & \underline{0.282} $\pm$ 0.004 & 0.293 & 0.304 & 0.376 & 0.330 & 0.387 & \textbf{0.279} & 0.341 & 0.515 \\
ett1/H/long & \textbf{0.263} $\pm$ 0.004 & 0.469 & \underline{0.297} & 0.363 & 0.313 & 0.567 & 0.430 & 1.94 & 0.616 \\
ett1/H/medium & \textbf{0.253} $\pm$ 0.002 & 0.535 & \underline{0.273} & 0.455 & 0.316 & 0.450 & 0.384 & 1.65 & 0.540 \\
ett1/H/short & \textbf{0.179} $\pm$ 0.002 & 0.233 & \underline{0.190} & 0.256 & 0.199 & 0.249 & 0.223 & 0.668 & 0.250 \\
ett1/W/short & \underline{0.306} $\pm$ 0.009 & 0.686 & 0.323 & 0.447 & 0.406 & 0.372 & \textbf{0.305} & 0.319 & 0.338 \\
ett2/15T/long & \textbf{0.097} $\pm$ 0.001 & 0.304 & \underline{0.098} & 0.141 & 0.109 & 0.126 & 0.165 & 0.169 & 0.165 \\
ett2/15T/medium & \textbf{0.093} $\pm$ 0.001 & 0.258 & \underline{0.094} & 0.151 & 0.104 & 0.138 & 0.143 & 0.150 & 0.143 \\
ett2/15T/short & \textbf{0.066} $\pm$ 0.001 & 0.378 & \underline{0.076} & 0.102 & 0.081 & 0.109 & 0.096 & 0.077 & 0.096 \\
ett2/D/short & \textbf{0.092} $\pm$ 0.001 & 0.207 & 0.131 & 0.218 & \underline{0.096} & 0.140 & 0.125 & 0.164 & 0.205 \\
ett2/H/long & \textbf{0.116} $\pm$ 0.004 & 0.196 & \underline{0.130} & 0.165 & 0.138 & 0.156 & 0.272 & 0.336 & 0.287 \\
ett2/H/medium & \textbf{0.106} $\pm$ 0.002 & 0.281 & 0.125 & 0.166 & \underline{0.122} & 0.130 & 0.245 & 0.284 & 0.241 \\
ett2/H/short & \textbf{0.065} $\pm$ 0.001 & 0.122 & \underline{0.074} & 0.088 & 0.078 & 0.091 & 0.089 & 0.102 & 0.094 \\
ett2/W/short & \textbf{0.088} $\pm$ 0.002 & 0.728 & 0.142 & 0.194 & 0.160 & 0.294 & \underline{0.136} & 0.160 & 0.169 \\
hierarchical\_sales/D/short & \textbf{0.572} $\pm$ 0.002 & 0.600 & \underline{0.590} & 0.817 & 0.600 & 0.728 & 0.735 & 0.967 & 2.36 \\
hierarchical\_sales/W/short & \textbf{0.349} $\pm$ 0.003 & 0.379 & \underline{0.358} & 0.582 & 0.382 & 0.439 & 0.485 & 0.474 & 1.03 \\
hospital/M/short & \textbf{0.052} $\pm$ 0.000 & 0.062 & 0.064 & 0.076 & 0.058 & 0.068 & 0.060 & \underline{0.055} & 0.062 \\
\bottomrule
\end{tabular}

}
\label{tab:gifteval-zs-taskspec-WQL-p1}
\end{table}

\setNextCaption{CRPS scores of \modelname compared with various task-specific and local models on the \underline{GiftEval} benchmark evaluation settings (Part 2/2). Models achieving the \textbf{best} and \underline{second-best} scores are highlighted, while the grey shade indicates results on datasets \modelname trained on. We trained \modelname with 6 different seeds and report the observed standard deviation in the plot.}

\begin{table}[h]
\centering
\caption{\nextcaption}
{\scriptsize
\begin{tabular}{llllllllll}
\toprule
  & \rotatebox{90}{  \modelname } & \rotatebox{90}{ DeepAR } & \rotatebox{90}{ PatchTST } & \rotatebox{90}{ DLinear } & \rotatebox{90}{ TFT } & \rotatebox{90}{ N-Beats } & \rotatebox{90}{ Auto ARIMA } & \rotatebox{90}{ Auto Theta } & \rotatebox{90}{ Seas. Naive }\\
\midrule
jena\_weather/10T/long & \underline{0.053} $\pm$ 0.001 & 0.143 & 0.066 & 0.093 & \textbf{0.052} & 0.134 & 0.304 & 0.424 & 0.304 \\
jena\_weather/10T/medium & \textbf{0.051} $\pm$ 0.001 & 0.073 & 0.065 & 0.098 & \underline{0.052} & 0.089 & 0.277 & 0.350 & 0.277 \\
jena\_weather/10T/short & \textbf{0.030} $\pm$ 0.001 & \underline{0.063} & 0.064 & 0.129 & 0.069 & 0.104 & 0.155 & 0.130 & 0.155 \\
jena\_weather/D/short & \textbf{0.046} $\pm$ 0.001 & 0.062 & \underline{0.053} & 0.073 & 0.069 & 0.193 & 0.080 & 0.082 & 0.297 \\
jena\_weather/H/long & \textbf{0.057} $\pm$ 0.002 & 0.197 & \underline{0.076} & 0.139 & 0.090 & 0.131 & 0.230 & 1.29 & 0.598 \\
jena\_weather/H/medium & \textbf{0.051} $\pm$ 0.002 & 0.078 & \underline{0.069} & 0.093 & 0.073 & 0.097 & 0.211 & 0.832 & 0.486 \\
jena\_weather/H/short & \textbf{0.041} $\pm$ 0.000 & 0.699 & 0.050 & 0.086 & \underline{0.048} & 0.098 & 0.143 & 0.296 & 0.173 \\
kdd\_cup\_2018/D/short & \cellcolor{gray!30}{0.381} $\pm$ 0.005 & \underline{0.383} & 0.401 & 0.482 & \textbf{0.380} & 0.529 & 0.393 & 0.459 & 0.888 \\
kdd\_cup\_2018/H/long & \cellcolor{gray!30}{0.341} $\pm$ 0.014 & 1.09 & \textbf{0.477} & 0.583 & \underline{0.503} & 0.660 & 1.05 & 0.970 & 1.25 \\
kdd\_cup\_2018/H/medium & \cellcolor{gray!30}{0.337} $\pm$ 0.011 & \textbf{0.442} & \textbf{0.442} & 0.548 & \underline{0.472} & 0.660 & 0.851 & 0.791 & 0.949 \\
kdd\_cup\_2018/H/short & \cellcolor{gray!30}{0.270} $\pm$ 0.004 & 0.517 & \textbf{0.457} & 0.581 & \underline{0.467} & 0.683 & 0.559 & 0.531 & 0.559 \\
loop\_seattle/5T/long & \underline{0.090} $\pm$ 0.004 & 0.184 & 0.095 & 0.131 & \textbf{0.088} & 0.117 & 0.137 & 0.231 & 0.137 \\
loop\_seattle/5T/medium & \textbf{0.083} $\pm$ 0.002 & 0.118 & 0.095 & 0.126 & \underline{0.092} & 0.118 & 0.123 & 0.240 & 0.123 \\
loop\_seattle/5T/short & \textbf{0.049} $\pm$ 0.000 & 0.072 & 0.066 & 0.099 & \underline{0.065} & 0.080 & 0.081 & 0.082 & 0.081 \\
loop\_seattle/D/short & \textbf{0.042} $\pm$ 0.000 & 0.052 & \underline{0.046} & 0.053 & 0.048 & 0.053 & 0.078 & 0.072 & 0.131 \\
loop\_seattle/H/long & \textbf{0.063} $\pm$ 0.001 & \underline{0.068} & 0.069 & 0.088 & \underline{0.068} & 0.080 & 0.193 & 0.468 & 0.245 \\
loop\_seattle/H/medium & \textbf{0.065} $\pm$ 0.001 & 0.072 & 0.071 & 0.104 & \underline{0.069} & 0.088 & 0.154 & 0.390 & 0.206 \\
loop\_seattle/H/short & \textbf{0.059} $\pm$ 0.000 & \underline{0.066} & 0.076 & 0.099 & 0.073 & 0.090 & 0.108 & 0.165 & 0.108 \\
m4\_daily/D/short & \cellcolor{gray!30}{0.021} $\pm$ 0.000 & 0.030 & \textbf{0.023} & 0.029 & \textbf{0.023} & 0.029 & \textbf{0.023} & \underline{0.024} & 0.026 \\
m4\_hourly/H/short & \cellcolor{gray!30}{0.021} $\pm$ 0.000 & 0.133 & \underline{0.039} & 0.055 & 0.040 & 0.050 & \textbf{0.034} & 0.041 & 0.040 \\
m4\_monthly/M/short & \cellcolor{gray!30}{0.093} $\pm$ 0.000 & 0.184 & \underline{0.102} & 0.129 & 0.113 & 0.122 & \textbf{0.098} & \textbf{0.098} & 0.126 \\
m4\_quarterly/Q/short & \textbf{0.074} $\pm$ 0.000 & 0.083 & 0.083 & 0.110 & 0.083 & 0.096 & 0.082 & \underline{0.079} & 0.099 \\
m4\_weekly/W/short & \cellcolor{gray!30}{0.035} $\pm$ 0.001 & 0.062 & \textbf{0.040} & 0.070 & 0.049 & \underline{0.047} & 0.050 & 0.053 & 0.073 \\
m4\_yearly/A/short & 0.119 $\pm$ 0.002 & \underline{0.113} & 0.117 & 0.168 & \textbf{0.110} & 0.134 & 0.130 & 0.115 & 0.138 \\
m\_dense/D/short & \textbf{0.066} $\pm$ 0.002 & 0.076 & \underline{0.070} & 0.123 & 0.077 & 0.087 & 0.135 & 0.126 & 0.294 \\
m\_dense/H/long & 0.122 $\pm$ 0.003 & 0.130 & \underline{0.120} & 0.259 & \textbf{0.115} & 0.243 & 0.270 & 1.43 & 0.552 \\
m\_dense/H/medium & 0.121 $\pm$ 0.002 & \underline{0.118} & 0.127 & 0.191 & \textbf{0.114} & 0.184 & 0.255 & 1.21 & 0.479 \\
m\_dense/H/short & \underline{0.130} $\pm$ 0.001 & \textbf{0.128} & 0.173 & 0.214 & 0.139 & 0.190 & 0.281 & 0.549 & 0.281 \\
restaurant/D/short & \textbf{0.254} $\pm$ 0.001 & 0.270 & \underline{0.262} & 0.340 & 0.284 & 0.342 & 0.362 & 0.329 & 0.907 \\
saugeen/D/short & \textbf{0.382} $\pm$ 0.014 & 0.572 & \underline{0.408} & 0.613 & 0.419 & 0.478 & 0.564 & 0.669 & 0.754 \\
saugeen/M/short & \textbf{0.303} $\pm$ 0.009 & 0.689 & 0.372 & 0.490 & 0.340 & 0.388 & \underline{0.326} & 0.373 & 0.445 \\
saugeen/W/short & \textbf{0.353} $\pm$ 0.009 & \underline{0.397} & 0.484 & 0.673 & 0.491 & 0.574 & 0.549 & 0.734 & 0.855 \\
solar/10T/long & \textbf{0.328} $\pm$ 0.008 & 0.549 & \underline{0.339} & 0.585 & 0.379 & 1.01 & 0.786 & 6.64 & 0.786 \\
solar/10T/medium & \textbf{0.354} $\pm$ 0.016 & 0.485 & \underline{0.356} & 0.552 & 0.362 & 1.00 & 0.771 & 5.67 & 0.771 \\
solar/10T/short & \textbf{0.542} $\pm$ 0.017 & 0.933 & 1.37 & 0.824 & 0.618 & \underline{0.576} & 0.860 & 2.36 & 0.860 \\
solar/D/short & \underline{0.281} $\pm$ 0.002 & 0.682 & 0.287 & 0.383 & \textbf{0.277} & 0.450 & 0.282 & 0.286 & 0.757 \\
solar/H/long & \textbf{0.243} $\pm$ 0.008 & 0.381 & \underline{0.353} & 0.612 & 0.401 & 1.00 & 0.607 & 7.32 & 1.47 \\
solar/H/medium & \textbf{0.260} $\pm$ 0.010 & 0.352 & 0.344 & 0.552 & \underline{0.330} & 1.00 & 0.557 & 6.13 & 1.27 \\
solar/H/short & \textbf{0.259} $\pm$ 0.009 & 0.389 & \underline{0.340} & 0.507 & 0.367 & 0.498 & 0.628 & 2.33 & 0.628 \\
solar/W/short & 0.154 $\pm$ 0.008 & 0.242 & 0.162 & 0.210 & \textbf{0.114} & 0.432 & \underline{0.152} & 0.155 & 0.236 \\
sz\_taxi/15T/long & \textbf{0.197} $\pm$ 0.001 & 0.286 & 0.281 & 0.396 & \underline{0.241} & 0.323 & 0.398 & 0.629 & 0.554 \\
sz\_taxi/15T/medium & \textbf{0.202} $\pm$ 0.001 & 0.210 & 0.220 & 0.296 & \underline{0.206} & 0.314 & 0.351 & 0.529 & 0.454 \\
sz\_taxi/15T/short & \textbf{0.200} $\pm$ 0.000 & 0.219 & \underline{0.207} & 0.271 & 0.222 & 0.281 & 0.309 & 0.288 & 0.309 \\
sz\_taxi/H/short & \textbf{0.136} $\pm$ 0.000 & \underline{0.139} & 0.144 & 0.201 & 0.144 & 0.190 & 0.170 & 0.232 & 0.229 \\
temperature\_rain/D/short & \cellcolor{gray!30}{0.550} $\pm$ 0.002 & 0.682 & \underline{0.644} & 0.830 & \textbf{0.592} & 0.840 & 0.694 & 0.761 & 1.63 \\
us\_births/D/short & \underline{0.021} $\pm$ 0.001 & 0.028 & 0.025 & 0.041 & \textbf{0.016} & 0.029 & 0.074 & 0.075 & 0.144 \\
us\_births/M/short & 0.017 $\pm$ 0.002 & \underline{0.016} & 0.017 & 0.032 & 0.021 & 0.025 & \textbf{0.010} & 0.019 & 0.017 \\
us\_births/W/short & \textbf{0.013} $\pm$ 0.000 & 0.017 & \underline{0.015} & 0.022 & 0.019 & 0.021 & 0.018 & 0.018 & 0.022 \\
\bottomrule
\end{tabular}

}
\label{tab:gifteval-zs-taskspec-WQL-p2}
\end{table}

\setNextCaption{MASE scores of different zero-shot models on the \underline{Chronos-ZS} benchmark datasets. The models achieving the \textbf{best} and \underline{second-best} scores are highlighted. Results for datasets that are part of the training data for the respective models are shaded in grey, and these results are excluded from the calculation of the best score. We trained \modelname with 6 different seeds and report the observed standard deviation in the plot.}

\begin{table}[h]
\centering
\caption{\nextcaption}
{\scriptsize
\begin{tabular}{llllllllllll}
\toprule
  & \rotatebox{90}{  \modelname } & \rotatebox{90}{ Chronos Bolt B } & \rotatebox{90}{ Chronos Bolt S } & \rotatebox{90}{ TimesFM 2.0 } & \rotatebox{90}{ TimesFM 1.0 } & \rotatebox{90}{ TabPFN-TS } & \rotatebox{90}{ Moirai L 1.1 } & \rotatebox{90}{ Moirai B 1.1 } & \rotatebox{90}{ TTM-r2 } & \rotatebox{90}{ Chronos B } & \rotatebox{90}{ Chronos S }\\
\midrule
ETTh & 0.793 $\pm$ 0.019 & \textbf{0.748} & 0.793 & 0.854 & 0.890 & 0.903 & 0.825 & 0.902 & 0.973 & \underline{0.774} & 0.827 \\
ETTm & 0.641 $\pm$ 0.006 & 0.633 & \underline{0.621} & \textbf{0.620} & 1.04 & 0.806 & 0.755 & 0.826 & 0.781 & 0.803 & 0.718 \\
australian electricity & 1.06 $\pm$ 0.096 & \textbf{0.740} & \underline{0.813} & 1.32 & 1.63 & 0.997 & \cellcolor{gray!30}{0.930} & \cellcolor{gray!30}{1.25} & \cellcolor{gray!30}{1.22} & 1.11 & 1.28 \\
car parts & \textbf{0.838} $\pm$ 0.004 & \underline{0.855} & 0.858 & 0.986 & 0.901 & \textbf{0.838} & \cellcolor{gray!30}{0.846} & \cellcolor{gray!30}{0.863} & 1.16 & 0.889 & 0.875 \\
cif 2016 & \underline{0.923} $\pm$ 0.005 & 0.999 & 1.02 & 0.949 & 1.04 & \textbf{0.894} & 1.03 & 1.06 & 1.61 & 0.985 & 0.996 \\
covid deaths & 39.5 $\pm$ 0.803 & 38.9 & \textbf{36.5} & 42.2 & 55.6 & \underline{37.8} & \cellcolor{gray!30}{32.1} & \cellcolor{gray!30}{34.3} & 49.5 & 43.6 & 42.6 \\
dominick & \textbf{0.797} $\pm$ 0.004 & 0.856 & 0.871 & 0.935 & 1.13 & 0.864 & 0.830 & 0.828 & 1.31 & 0.824 & \underline{0.812} \\
ercot & 0.630 $\pm$ 0.040 & 0.693 & 0.756 & 0.688 & 0.589 & 0.603 & \underline{0.568} & 0.569 & 1.03 & \textbf{0.463} & 0.571 \\
exchange rate & 2.20 $\pm$ 0.157 & \underline{1.71} & 1.77 & 1.98 & 3.34 & \textbf{1.44} & 1.74 & 1.81 & 2.01 & 2.18 & 1.78 \\
fred md & \underline{0.451} $\pm$ 0.023 & 0.646 & 0.612 & 0.498 & 0.650 & 0.530 & \cellcolor{gray!30}{0.564} & \cellcolor{gray!30}{0.572} & 0.698 & \textbf{0.445} & \textbf{0.445} \\
hospital & 0.767 $\pm$ 0.003 & 0.791 & 0.801 & \underline{0.755} & 0.783 & \textbf{0.686} & \cellcolor{gray!30}{0.830} & \cellcolor{gray!30}{0.820} & 0.843 & 0.812 & 0.815 \\
m1 monthly & \underline{1.05} $\pm$ 0.008 & 1.10 & 1.13 & \underline{1.05} & 1.07 & \textbf{1.04} & \cellcolor{gray!30}{1.14} & \cellcolor{gray!30}{1.14} & 1.47 & 1.12 & 1.16 \\
m1 quarterly & 1.69 $\pm$ 0.011 & 1.77 & 1.84 & 1.71 & \underline{1.67} & \textbf{1.66} & \cellcolor{gray!30}{1.79} & \cellcolor{gray!30}{1.64} & 2.12 & 1.76 & 1.81 \\
m1 yearly & 3.67 $\pm$ 0.095 & 4.40 & 5.10 & \underline{3.60} & 4.00 & \textbf{3.59} & \cellcolor{gray!30}{3.52} & \cellcolor{gray!30}{3.61} & 5.78 & 4.48 & 4.90 \\
m3 monthly & 0.847 $\pm$ 0.009 & 0.851 & 0.869 & \textbf{0.832} & 0.935 & \underline{0.845} & \cellcolor{gray!30}{0.902} & \cellcolor{gray!30}{0.897} & 1.23 & 0.863 & 0.893 \\
m3 quarterly & 1.17 $\pm$ 0.011 & 1.29 & 1.33 & 1.20 & \underline{1.15} & \textbf{1.10} & \cellcolor{gray!30}{1.12} & \cellcolor{gray!30}{1.15} & 1.72 & 1.18 & 1.29 \\
m3 yearly & 2.76 $\pm$ 0.070 & 2.91 & 3.41 & 2.82 & \textbf{2.70} & \underline{2.73} & \cellcolor{gray!30}{2.75} & \cellcolor{gray!30}{2.73} & 3.87 & 3.17 & 3.42 \\
m4 quarterly & \underline{1.18} $\pm$ 0.013 & 1.22 & 1.25 & \cellcolor{gray!30}{0.965} & \cellcolor{gray!30}{1.40} & \textbf{1.17} & \cellcolor{gray!30}{1.17} & \cellcolor{gray!30}{1.17} & 1.81 & 1.24 & 1.25 \\
m4 yearly & \underline{3.46} $\pm$ 0.076 & 3.51 & 3.69 & \cellcolor{gray!30}{2.55} & \cellcolor{gray!30}{3.37} & \textbf{3.18} & \cellcolor{gray!30}{3.04} & \cellcolor{gray!30}{3.07} & 4.89 & 3.69 & 3.80 \\
m5 & \textbf{0.904} $\pm$ 0.001 & \underline{0.915} & 0.920 & 0.919 & 0.919 & 0.927 & \cellcolor{gray!30}{0.927} & \cellcolor{gray!30}{0.924} & 1.10 & 0.934 & 0.931 \\
nn5 & \textbf{0.563} $\pm$ 0.007 & \underline{0.576} & 0.583 & 0.608 & 0.629 & 0.665 & \cellcolor{gray!30}{0.586} & \cellcolor{gray!30}{0.638} & \cellcolor{gray!30}{0.998} & 0.594 & 0.618 \\
nn5 weekly & 0.913 $\pm$ 0.010 & 0.916 & 0.927 & \textbf{0.865} & 0.949 & \underline{0.882} & \cellcolor{gray!30}{0.979} & \cellcolor{gray!30}{0.939} & 0.994 & 0.941 & 0.946 \\
tourism monthly & \underline{1.47} $\pm$ 0.014 & 1.53 & 1.61 & 1.62 & 1.92 & \textbf{1.46} & \cellcolor{gray!30}{1.65} & \cellcolor{gray!30}{1.75} & 3.21 & 1.84 & 1.93 \\
tourism quarterly & \underline{1.64} $\pm$ 0.026 & 1.76 & 1.84 & 1.80 & 2.06 & \textbf{1.59} & \cellcolor{gray!30}{1.82} & \cellcolor{gray!30}{1.92} & 3.77 & 1.81 & 1.77 \\
tourism yearly & 3.50 $\pm$ 0.100 & 3.69 & 3.89 & 3.49 & \underline{3.23} & \textbf{3.02} & \cellcolor{gray!30}{3.21} & \cellcolor{gray!30}{3.15} & 3.56 & 3.84 & 3.98 \\
traffic & 0.799 $\pm$ 0.013 & \textbf{0.784} & 0.860 & \cellcolor{gray!30}{0.726} & \cellcolor{gray!30}{0.641} & \underline{0.791} & \cellcolor{gray!30}{0.758} & \cellcolor{gray!30}{0.768} & 0.835 & 0.812 & 0.831 \\
weather & \textbf{0.790} $\pm$ 0.008 & 0.812 & \underline{0.802} & 0.822 & 0.913 & 0.811 & \cellcolor{gray!30}{0.810} & \cellcolor{gray!30}{0.823} & \cellcolor{gray!30}{0.896} & 0.809 & 0.850 \\
\bottomrule
\end{tabular}

}
\label{tab:chronos-zs-zs-model-MASE}
\end{table}

\setNextCaption{WQL scores of different zero-shot models on the \underline{Chronos-ZS} benchmark datasets. The models achieving the \textbf{best} and \underline{second-best} scores are highlighted. Results for datasets that are part of the training data for the respective models are shaded in grey, and these results are excluded from the calculation of the best score. We trained \modelname with 6 different seeds and report the observed standard deviation in the plot.}

\begin{table}[h]
\centering
\caption{\nextcaption}
{\scriptsize
\begin{tabular}{llllllllllll}
\toprule
  & \rotatebox{90}{  \modelname } & \rotatebox{90}{ Chronos Bolt B } & \rotatebox{90}{ Chronos Bolt S } & \rotatebox{90}{ TimesFM 2.0 } & \rotatebox{90}{ TimesFM 1.0 } & \rotatebox{90}{ TabPFN-TS } & \rotatebox{90}{ Moirai L 1.1 } & \rotatebox{90}{ Moirai B 1.1 } & \rotatebox{90}{ TTM-r2 } & \rotatebox{90}{ Chronos B } & \rotatebox{90}{ Chronos S }\\
\midrule
ETTh & 0.079 $\pm$ 0.003 & \textbf{0.071} & \underline{0.076} & 0.085 & 0.092 & 0.098 & 0.082 & 0.091 & 0.118 & 0.080 & 0.083 \\
ETTm & 0.054 $\pm$ 0.001 & \underline{0.052} & \textbf{0.051} & \underline{0.052} & 0.084 & 0.077 & 0.070 & 0.076 & 0.076 & 0.070 & 0.063 \\
australian electricity & 0.059 $\pm$ 0.005 & \textbf{0.036} & \underline{0.042} & 0.067 & 0.089 & 0.045 & \cellcolor{gray!30}{0.038} & \cellcolor{gray!30}{0.054} & \cellcolor{gray!30}{0.074} & 0.067 & 0.072 \\
car parts & \underline{0.990} $\pm$ 0.010 & 0.995 & 1.01 & 1.65 & 1.04 & \textbf{0.949} & \cellcolor{gray!30}{0.990} & \cellcolor{gray!30}{1.02} & 1.38 & 1.07 & 1.03 \\
cif 2016 & \underline{0.012} $\pm$ 0.002 & 0.016 & 0.016 & 0.053 & 0.020 & \textbf{0.009} & 0.015 & 0.016 & 0.038 & \underline{0.012} & \underline{0.012} \\
covid deaths & \textbf{0.037} $\pm$ 0.004 & 0.047 & 0.043 & 0.215 & 0.204 & \underline{0.040} & \cellcolor{gray!30}{0.036} & \cellcolor{gray!30}{0.045} & 0.108 & 0.045 & 0.063 \\
dominick & \textbf{0.321} $\pm$ 0.001 & 0.345 & 0.348 & 0.371 & 0.412 & 0.335 & 0.344 & 0.343 & 0.552 & \underline{0.331} & 0.336 \\
ercot & 0.019 $\pm$ 0.002 & 0.021 & 0.026 & 0.021 & 0.021 & 0.020 & 0.017 & 0.018 & 0.044 & \textbf{0.013} & \underline{0.015} \\
exchange rate & 0.013 $\pm$ 0.002 & 0.012 & \underline{0.011} & 0.015 & 0.013 & \textbf{0.010} & \underline{0.011} & \underline{0.011} & 0.377 & 0.012 & 0.012 \\
fred md & 0.023 $\pm$ 0.006 & 0.042 & 0.037 & 0.027 & 0.036 & 0.055 & \cellcolor{gray!30}{0.042} & \cellcolor{gray!30}{0.050} & 0.050 & \underline{0.020} & \textbf{0.015} \\
hospital & \underline{0.052} $\pm$ 0.000 & 0.057 & 0.058 & \textbf{0.050} & \underline{0.052} & 0.063 & \cellcolor{gray!30}{0.059} & \cellcolor{gray!30}{0.058} & 0.079 & 0.056 & 0.057 \\
m1 monthly & 0.136 $\pm$ 0.004 & 0.139 & 0.134 & \underline{0.130} & \textbf{0.123} & 0.150 & \cellcolor{gray!30}{0.177} & \cellcolor{gray!30}{0.169} & 0.215 & 0.131 & 0.138 \\
m1 quarterly & 0.099 $\pm$ 0.004 & 0.101 & 0.094 & 0.113 & \textbf{0.087} & \underline{0.090} & \cellcolor{gray!30}{0.093} & \cellcolor{gray!30}{0.076} & 0.156 & 0.102 & 0.106 \\
m1 yearly & \underline{0.135} $\pm$ 0.008 & 0.151 & 0.157 & 0.145 & 0.163 & \textbf{0.118} & \cellcolor{gray!30}{0.127} & \cellcolor{gray!30}{0.123} & 0.246 & 0.198 & 0.183 \\
m3 monthly & 0.091 $\pm$ 0.000 & 0.093 & 0.094 & \textbf{0.089} & 0.098 & \underline{0.090} & \cellcolor{gray!30}{0.099} & \cellcolor{gray!30}{0.101} & 0.153 & 0.096 & 0.100 \\
m3 quarterly & \underline{0.070} $\pm$ 0.000 & 0.076 & 0.077 & 0.075 & 0.072 & \textbf{0.067} & \cellcolor{gray!30}{0.070} & \cellcolor{gray!30}{0.070} & 0.115 & 0.074 & 0.080 \\
m3 yearly & 0.131 $\pm$ 0.004 & \underline{0.129} & 0.155 & 0.144 & \textbf{0.123} & 0.130 & \cellcolor{gray!30}{0.130} & \cellcolor{gray!30}{0.131} & 0.191 & 0.149 & 0.157 \\
m4 quarterly & \textbf{0.074} $\pm$ 0.000 & 0.077 & 0.078 & \cellcolor{gray!30}{0.062} & \cellcolor{gray!30}{0.085} & \underline{0.075} & \cellcolor{gray!30}{0.076} & \cellcolor{gray!30}{0.076} & 0.125 & 0.083 & 0.083 \\
m4 yearly & \underline{0.119} $\pm$ 0.002 & 0.121 & 0.128 & \cellcolor{gray!30}{0.091} & \cellcolor{gray!30}{0.117} & \textbf{0.114} & \cellcolor{gray!30}{0.108} & \cellcolor{gray!30}{0.109} & 0.192 & 0.135 & 0.138 \\
m5 & \textbf{0.551} $\pm$ 0.001 & 0.562 & 0.567 & \underline{0.557} & 0.561 & 0.565 & \cellcolor{gray!30}{0.594} & \cellcolor{gray!30}{0.583} & 0.668 & 0.585 & 0.586 \\
nn5 & \textbf{0.146} $\pm$ 0.002 & \underline{0.150} & 0.151 & 0.155 & 0.160 & 0.173 & \cellcolor{gray!30}{0.152} & \cellcolor{gray!30}{0.165} & \cellcolor{gray!30}{0.311} & 0.162 & 0.169 \\
nn5 weekly & 0.083 $\pm$ 0.001 & 0.084 & 0.085 & \textbf{0.079} & 0.086 & \underline{0.081} & \cellcolor{gray!30}{0.091} & \cellcolor{gray!30}{0.089} & 0.113 & 0.089 & 0.089 \\
tourism monthly & \textbf{0.077} $\pm$ 0.002 & 0.090 & 0.094 & 0.085 & 0.101 & \underline{0.084} & \cellcolor{gray!30}{0.098} & \cellcolor{gray!30}{0.099} & 0.283 & 0.101 & 0.108 \\
tourism quarterly & \textbf{0.061} $\pm$ 0.002 & \underline{0.065} & 0.067 & 0.070 & 0.085 & 0.093 & \cellcolor{gray!30}{0.067} & \cellcolor{gray!30}{0.070} & 0.200 & 0.077 & 0.070 \\
tourism yearly & \textbf{0.148} $\pm$ 0.006 & 0.166 & 0.168 & \underline{0.163} & \textbf{0.148} & \textbf{0.148} & \cellcolor{gray!30}{0.137} & \cellcolor{gray!30}{0.138} & 0.212 & 0.199 & 0.201 \\
traffic & 0.235 $\pm$ 0.004 & \underline{0.231} & 0.252 & \cellcolor{gray!30}{0.212} & \cellcolor{gray!30}{0.185} & \textbf{0.229} & \cellcolor{gray!30}{0.236} & \cellcolor{gray!30}{0.238} & 5.54 & 0.255 & 0.258 \\
weather & \textbf{0.127} $\pm$ 0.000 & 0.134 & 0.133 & 0.133 & 0.150 & \underline{0.132} & \cellcolor{gray!30}{0.133} & \cellcolor{gray!30}{0.135} & \cellcolor{gray!30}{0.159} & 0.137 & 0.147 \\
\bottomrule
\end{tabular}

}
\label{tab:chronos-zs-zs-model-WQL}
\end{table}

\clearpage
\newpage

\end{document}